\documentclass[letterpaper,twoside]{article}
\PassOptionsToPackage{round,authoryear}{natbib}
\usepackage[preprint]{aistats2027}
\usepackage[T1]{fontenc}
\usepackage{amsmath,amssymb,amsthm,mathtools,bm}
\usepackage{graphicx,booktabs,longtable,microtype,enumitem,array}
\newcolumntype{P}[1]{>{\raggedright\arraybackslash}p{#1}}
\usepackage{natbib}

\usepackage[hidelinks]{hyperref}
\usepackage{url}
\newtheorem{theorem}{Theorem}
\newtheorem{proposition}[theorem]{Proposition}
\newtheorem{corollary}[theorem]{Corollary}
\newtheorem{lemma}[theorem]{Lemma}
\theoremstyle{definition}\newtheorem{assumption}{Assumption}
\newcommand{\E}{\mathbb E}\newcommand{\Pp}{\mathbb P}
\newcommand{\R}{\mathbb R}\newcommand{\N}{\mathcal N}
\newcommand{\Wass}{\mathsf W}\newcommand{\distK}{d_{\mathrm K}}
\newcommand{\iid}{i.i.d.\ }
\newcommand{\eps}{\varepsilon}\newcommand{\onevec}{\mathbf 1}
\DeclareMathOperator{\Var}{Var}\DeclareMathOperator{\Cov}{Cov}
\newcommand{\norm}[1]{\lVert#1\rVert}

\newcommand{\calJ}{\mathcal J}\newcommand{\calR}{\mathcal R}
\hypersetup{pdftitle={Beyond Gaussian Calibration: Local Inference in Decentralized SGD},pdfauthor={Xiaoli Li, Qianqian Lei, Wei Biao Wu}}
\begin{document}
\raggedbottom
\runningtitle{Beyond Gaussian Calibration: Local Inference in Decentralized SGD}
\runningauthor{Xiaoli Li, Qianqian Lei, Wei Biao Wu}
\twocolumn[
\aistatstitle{Beyond Gaussian Calibration:\\Local Inference in Decentralized SGD}
\aistatsauthor{Xiaoli Li \And Qianqian Lei \And Wei Biao Wu}
\aistatsaddress{University of Chicago\\
\href{mailto:xiaolil@uchicago.edu}{\texttt{xiaolil@uchicago.edu}}
\And University of Chicago\\
\href{mailto:qianqian@uchicago.edu}{\texttt{qianqian@uchicago.edu}}
\And University of Chicago\\
\href{mailto:wbwu@uchicago.edu}{\texttt{wbwu@uchicago.edu}}}
]
\fancyhead[CE]{\small\bfseries Beyond Gaussian Calibration}
\fancyhead[CO]{\small\bfseries Xiaoli Li, Qianqian Lei, Wei Biao Wu}
\thispagestyle{plain}
\begin{abstract}
Decentralized SGD minimizes a common objective by local gradient steps and neighbor averaging. Its statistical theory mostly concerns network averages, yet clients act on their own iterates, and one-sided decisions such as alarms use one tail of the local error. We show that a client's constant-step endpoint can have nearly nominal Gaussian coverage while both tails are badly miscalibrated. An exact noise filter explains why: a quantity computable from the graph, stepsize and curvature decides whether the endpoint is asymptotically Gaussian, and when the number of clients outgrows the inverse stepsize on networks with summable return probabilities, such as lazy tori of dimension at least three and expanders, a few recent noise terms at nearby clients keep a fixed share of local variance. On these tori the limit is a non-Gaussian local field plus an independent Gaussian consensus term, and no reference built from finitely many noise moments, including Edgeworth and Cornish--Fisher corrections, is calibrated at all levels uniformly over noise laws. We propose a hybrid reference that resamples the influential noise terms from their empirical law and replaces the rest by a Gaussian; its non-asymptotic coverage bound allows noise laws and curvature estimated from fresh gradients or the client's trajectory, and extends to nonlinear vector objectives. In full-network simulations it restores tail balance at nearly unchanged length, also under rare large shocks that break Cornish--Fisher intervals given enough residuals; gains vanish, as predicted, when consensus dominates.
\end{abstract}
\section{Introduction}\label{sec:intro}
\begin{figure*}[t]
\centering\includegraphics[width=\textwidth]{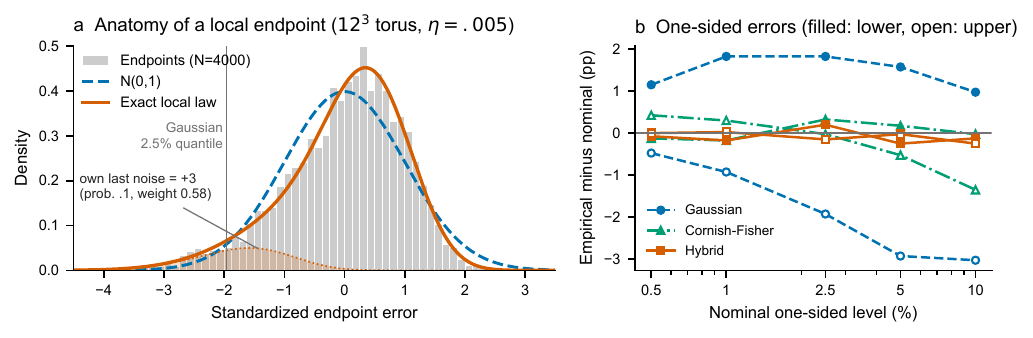}
\caption{A local endpoint under skew noise on the lazy $12^3$ torus with $\eta=.005$ (4,000 full-network trajectories). (a) Standardized endpoint errors with the exact local law (characteristic-function inversion) and $\N(0,1)$. The shaded component is the 10\% event that the client's own last innovation equals $+3$; its weight $.58$ places it near the Gaussian 2.5\% quantile. (b) Empirical minus nominal one-sided error probabilities of the fitted references (filled: lower, open: upper tails); Cornish--Fisher is CF$_3$ of Section~\ref{sec:experiments}.}\label{fig:anatomy}
\end{figure*}
Decentralized stochastic gradient descent (DSGD) alternates local gradient steps with averaging among neighbors \citep{tsitsiklis1986,nedic2009,lian2017}. It underlies estimation over sensor networks and serverless federated learning \citep{rabbat2004,kar2012distributed,kairouz2021,mcmahan2017}, and its theory explains how the spectral gap and client heterogeneity slow optimization \citep{duchi2012dual,shi2015extra,yuan2016convergence,koloskova2020}. Its statistical theory mostly describes network-level behavior: with decreasing stepsizes, distributed stochastic approximation satisfies a central limit theorem whose covariance, under doubly stochastic weights, matches that of centralized SGD \citep{bianchi2013,morral2017}, and the mean-square error of DSGD becomes asymptotically independent of the network \citep{pu2020asymptotic,pu2022transient}. Inference likewise targets time-averaged, client-averaged or aggregated statistics and one-step corrections \citep{li2022localsgd,gu2024,bonnerjee2025,levin2026}. Yet in many deployments each device acts on its own current iterate, for example a sensor that raises an alarm when its estimate crosses a threshold. Such one-sided decisions depend on one tail of the local error distribution, and a central interval with nearly nominal coverage can place most of its errors in one tail.

We study the constant-step endpoint $\widehat\theta_i$ of one client $i$ and construct confidence intervals for $\theta^\star$ from it. An exact representation writes its error as a linear filter of past gradient innovations, with weights set by the communication matrix. On a fixed network the consensus component dominates as the stepsize decreases, and the endpoint becomes Gaussian. On growing networks with $m\eta\to\infty$, the consensus variance, of order $\eta/m$, becomes negligible against the local variance, of order $\eta^2$. If the random walk of the network also has summable return probabilities, as on lazy tori of dimension at least three and on expanders, a few recent innovations at nearby clients keep a fixed share of the local variance; on the three-dimensional lattice the client's own most recent innovation carries 29\% of it. Figure~\ref{fig:anatomy}(a) shows the consequence for skewed noise on a $12^3$ torus: a 10\% chance of a large shock in the client's own last gradient creates a component of the endpoint law near the Gaussian 2.5\% quantile, and the Gaussian interval misses 4.3\% below and 0.6\% above. Network independence, which holds on fixed networks, fails here.

Two simple remedies do not resolve this. A one-step estimate from $M$ fresh gradients discards information: in the same example the endpoint is as precise as about 27{,}000 gradients and eight times more accurate than the one-step estimate with $M=400$. Edgeworth and Cornish--Fisher corrections \citep{cornish1938,hall1992} expand around a small parameter that is absent here, and Figure~\ref{fig:anatomy}(b) shows that a Cornish--Fisher interval repairs the 2.5\% level but not other levels. We therefore calibrate the endpoint with a reference that keeps the influential innovations exactly, resampled from their empirical law, and replaces the remaining sum by a Gaussian.

\paragraph{Contributions.}
\begin{itemize}[leftmargin=1.1em,itemsep=1pt,topsep=2pt]
\item \emph{Local Gaussian boundary and limit law.} The endpoint is asymptotically Gaussian exactly when its largest standardized filter weight $\Lambda_i$ vanishes, and $\Lambda_i$ is computable before data are observed. On lazy tori with $d\ge3$ and $m\eta\to\varrho$, the limit is a harness-type local field plus an independent Gaussian consensus component (Theorem~\ref{thm:transientlaw}).
\item \emph{A moment-based impossibility result.} In this local regime, no reference that depends on the noise law only through finitely many moments is calibrated at all levels uniformly over noise laws (Proposition~\ref{prop:moments}); a moment correction can repair one level, but calibration across levels needs the law itself.
\item \emph{Feasible hybrid calibration.} We bound the coverage, one-sided and probability-integral-transform errors of the hybrid reference with an estimated noise law and curvature, including estimates computed from the endpoint trajectory (Theorem~\ref{thm:hybrid}).
\item \emph{Nonlinear vector objectives.} Under samplewise contraction, fixed contrasts of nonlinear vector objectives inherit the linear analysis on the local scale, without commutation of the Hessian and noise covariance (Theorem~\ref{thm:vectortransfer}).
\end{itemize}
In experiments the hybrid reference reduces the $12^3$ tail errors from $4.33\%/0.58\%$ to $2.70\%/2.35\%$ at the same length, repairs rare large shocks that break Cornish--Fisher intervals given 1,600 trajectory residuals, and matches the Gaussian reference when the Gaussian is adequate.

\paragraph{Scope and related work.} We assume homogeneous objectives, independent inputs and a known communication matrix. Constant-step analyses of single-agent SGD give stationary bias, small-stepsize limits, high-probability bounds and Gaussian approximation \citep{dieuleveut2020,chenstationary2021,mandt2017,yu2021constant,durmus2021,wei2025constant}; adaptive-network studies give each agent's steady-state mean-square error \citep{chen2015learning,sayed2014adaptation}, and \citet{versini2026} expand local bias and covariance for heterogeneous clients. We add the distributional shape of the local endpoint. Inference for SGD uses averaged iterates, plug-in or random-scaling covariances, online bootstrap and Gaussian approximation \citep{polyak1992,toulis2017,cheninference2016,fang2018,zhu2023cov,lee2022random,anastasiou2019,mou2020,su2023higrad,sheshukova2026}, and distributed inference aggregates estimates or bootstrap replicates across machines \citep{kleiner2014,jordan2019communication}. Multiplicative noise \citep{gurbuzbalaban2021heavy,hodgkinson2021multiplicative,gurbuzbalaban2025heavy} and flat minima \citep{zhang2026flat} cause other non-Gaussian behavior; ours arises with positive curvature and bounded noise. The reference combines residual bootstrap \citep{efron1979,buhlmann1997,kreiss2011} with large-component retention \citep{asmussen2001}; Appendix~\ref{app:related} compares in detail.

\section{Problem setting and local innovation filters}\label{sec:filter}
We consider a model consisting of $m$ clients that communicate over an undirected network without a central server. The general decentralized optimization problem is
\[
\theta^\star\in\arg\min_{x\in\R^p} f(x),\qquad f(x)=\frac1m\sum_{j=1}^m f_j(x),
\]
where $f_j(x)=\E[\ell_j(x;Z_j)]$ is client $j$'s population objective and $\ell_j$ is its sample loss. Client $j$ maintains an iterate $\theta_{j,t}\in\R^p$ and observes a stochastic gradient $g_j(\theta_{j,t},Z_{j,t+1})=\nabla_\theta\ell_j(\theta_{j,t},Z_{j,t+1})$ from its local data. In the adapt-then-combine (ATC) form of DSGD, each round consists of a local gradient step followed by neighbor averaging:
\[
\begin{aligned}
\widetilde\theta_{j,t+1}&=\theta_{j,t}-\eta g_j(\theta_{j,t},Z_{j,t+1}),\\
\theta_{i,t+1}&=\sum_{j=1}^m W_{ij}\widetilde\theta_{j,t+1},
\end{aligned}
\]
with constant stepsize $\eta>0$. The fixed matrix $W$ specifies the mixing weights: $W_{ij}=0$ for distinct clients that cannot communicate directly, and $W_{ii}$ is the weight on the client's own update. The analysis starts from the scalar common objective $f_j(x)=h(x-\theta^\star)^2/2$, $h>0$, with gradients $h(\theta_{j,t}-\theta^\star)+\eps_{j,t+1}$. This model isolates how gradient noise propagates, separately from heterogeneity and nonlinear drift; Section~\ref{sec:nonlinear} returns to nonlinear objectives.
\begin{assumption}[Network and innovations]\label{ass:filter}
(i) The communication matrix $W\in\R^{m\times m}$ is known, symmetric, entrywise nonnegative and doubly stochastic: $W\onevec=\onevec$ and $\onevec^\top W=\onevec^\top$. (ii) The innovations $\eps_{j,t}$ are \iid over clients $j$ and times $t$, with law $F_\eps$, mean zero, variance $\sigma^2>0$ and $\beta_3=\E|\eps_{j,t}|^3/\sigma^3<\infty$.
\end{assumption}
With the stacked errors $e_t=\theta_t-\theta^\star\onevec\in\R^m$ and $W\onevec=\onevec$, the update becomes
\begin{equation}\label{eq:recursion}
 e_{t+1}=W\{(1-\eta h)e_t-\eta\eps_{t+1}\},\qquad e_t=\theta_t-\theta^\star\onevec.
\end{equation}
Assume $0<\eta h\le1$ and write $a=1-\eta h\in[0,1)$, with $a^0=1$ also when $a=0$. After $T$ ATC rounds we allow $q\ge0$ communication-only rounds, so the endpoint is $\widehat\theta_i=(W^q\theta_T)_i$, and $q=0$ gives the usual endpoint $\theta_{i,T}$. The innovation $\eps_{j,T-k}$, generated at client $j$ in round $T-k$ ($k=0$ is the latest), enters with factor $-\eta$, is mixed once in its own round, and then passes through $k$ rounds of $aW$ and $q$ gossip rounds. Its contribution to $\widehat\theta_i-\theta^\star$ is therefore $-\eta c_{kj}\eps_{j,T-k}$, with propagation weights and power sums
\begin{equation}\label{eq:coeff}
 c_{kj}=a^k(W^{k+q+1})_{ij},\quad 0\le k<T,\quad S_r=\sum_{k,j}|c_{kj}|^r.
\end{equation}
Summing all contributions gives the exact filter representation
\begin{equation}\label{eq:filter}
 \frac{(W^qe_T)_i}{\eta}=\frac{a^T(W^{T+q}e_0)_i}{\eta}-\sum_{k,j}c_{kj}\eps_{j,T-k}.
\end{equation}
The sums run over $0\le k<T$ and $1\le j\le m$; all weights and their sums refer to the chosen client $i$. We write $T=\infty$ for the stationary filter, whose sum extends over all past innovations without an initialization term. Independence of the innovations gives variance $\eta^2\sigma^2S_2$ for their contribution to the endpoint error. We write $s_i=\eta\sigma\sqrt{S_2}$ for this local standard deviation and $b_{kj}=c_{kj}/\sqrt{S_2}$, so that $\sum_{k,j}b_{kj}^2=1$ and $b_{kj}^2$ is the share of local variance due to $\eps_{j,T-k}$.

A one-step estimate $x-\bar g/h$ from $n$ independent gradients at a point $x$ has variance $\sigma^2/(h^2n)$, so the endpoint is as precise as $n_{\rm eff}=1/(h^2\eta^2S_2)$ gradients; in the local regimes below $S_2$ is bounded and $n_{\rm eff}\asymp(h\eta)^{-2}$ far exceeds the calibration samples of Section~\ref{sec:hybrid}. Calibration uses local observations but global inputs: the known $W,\eta,T,q$ and client $i$'s rows of graph powers, which require $O(Tm)$ storage shared across replications.

\section{When local fluctuations are Gaussian}\label{sec:boundary}
We now ask when the local fluctuations described by the filter representation \eqref{eq:filter} admit a Gaussian approximation. Separating the initialization contribution, we consider the standardized innovation sum
\[
Y=-\frac1\sigma\sum_{k,j}b_{kj}\eps_{j,T-k},\qquad \E Y=0,\quad\Var(Y)=1.
\]
To quantify whether any single innovation contributes a substantial share of this variance, define the standardized largest influence
\begin{equation}\label{eq:lambda}
\Lambda_i=\max_{k,j}|b_{kj}|,
\end{equation}
Thus, $\Lambda_i^2$ is the largest variance share contributed by an individual innovation. We measure the discrepancy between two distributions using the Kolmogorov distance $\distK(F,G)=\sup_x|F(x)-G(x)|$. The following lemma relates this approximation error to the innovation weights.
\begin{lemma}[Gaussian boundary]\label{thm:boundary}
Under Assumption~\ref{ass:filter},
\begin{equation}\label{eq:be}
 \distK\{\mathcal L(Y),\N(0,1)\}
 \le C\beta_3\sum_{k,j}|b_{kj}|^3\le C\beta_3\Lambda_i.
\end{equation}
Along any sequence of filters with a fixed non-Gaussian innovation law of finite variance, with no third moment needed, $Y\Rightarrow\N(0,1)$ if and only if $\Lambda_i\to0$.
\end{lemma}
The lemma combines the Berry--Esseen and Lindeberg theorems with Cram\'er's decomposition theorem \citep{petrov1995,cramer1936}; Appendix~\ref{app:tools} gives the proof. With $C\le.56$ \citep{shevtsova2010}, \eqref{eq:be} gives a certificate computable before any data are seen: for a curvature bracket $[h_-,h_+]\ni h$, a bound $\bar\beta_3\ge\beta_3$ and weights $b_{kj}(u)$ at curvature $u$, the quantity $\mathcal B_i=.56\,\bar\beta_3\sup_{h_-\le u\le h_+}\sum_{k,j}|b_{kj}(u)|^3$ bounds $\distK\{\mathcal L(Y),\N(0,1)\}$, so a small $\mathcal B_i$ certifies the Gaussian approximation.

We compare fixed and growing networks through the stationary filter ($T=\infty$, $q=0$), with $m$ clients and fixed curvature $h>0$. Symmetry of $W$ gives
\[
S_2=\frac1{m(1-a^2)}+\sum_{k\ge0}a^{2k}\left\{(W^{2k+2})_{ii}-\frac1m\right\}.
\]
Multiplying the two terms by $\eta^2\sigma^2$ gives the consensus and disagreement contributions to the local variance. For a fixed connected, aperiodic $W$, the disagreement term stays bounded as $\eta\downarrow0$. Hence
\[
S_2\sim(2m\eta h)^{-1},\qquad \Lambda_i\le S_2^{-1/2}\to0,
\]
and Lemma~\ref{thm:boundary} gives a Gaussian limit.

For growing networks, suppose $W_{ii}\ge w_0>0$ and
\begin{equation}\label{eq:return}
\sum_{k\ge0}\left\{(W^{2k+2})_{ii}-\frac1m\right\}\le C_0
\end{equation}
uniformly in the network size. If $m\eta\to\infty$, then $S_2\le C_0+(m\eta h)^{-1}$ stays bounded, while $c_{0i}\ge w_0$. Thus $\Lambda_i$ stays bounded away from zero, precluding a Gaussian limit for any fixed non-Gaussian innovation law. Under these conditions, the competing consensus and disagreement variance scales are $\eta/m$ and $\eta^2$.

Condition~\eqref{eq:return} bounds the sum of return probabilities at even times, centered at $1/m$, for the random walk with transition matrix $W$ \citep{levin2017}. Lazy bounded-degree expanders \citep{hoory2006} and lazy complete graphs satisfy~\eqref{eq:return} (Corollary~\ref{cor:gap}). On lazy tori of side $L$ with $L\sqrt\eta\to\infty$, $S_2$ has order $\eta^{-1/2}$, $\log(1/\eta)$ and one for $d=1,2$ and $d\ge3$ \citep{lawler2010}, mirroring the recurrence and transience of random walks \citep{spitzer1964}. Table~\ref{tab:influence} in Appendix~\ref{app:influence} lists $\Lambda_i$ for common graphs; with $m=4096$ and $\eta=.01$ it ranges from $.24$ on a lazy ring to $.85$ on a lazy complete graph.
\begin{assumption}[Growing lazy tori]\label{ass:tori}
The $n$th network is the torus $(\mathbb Z/L_n\mathbb Z)^d$ with $m_n=L_n^d$ clients, fixed $d\ge3$, and lazy nearest-neighbor matrix $W_n=\frac12I+\frac1{4d}\sum_{r=1}^d(P_r+P_r^{-1})$, where $P_r$ shifts coordinate $r$. As $n\to\infty$, $L_n\to\infty$, $\eta_n\to0$ and $m_n\eta_n\to\varrho\in[0,\infty]$; $h>0$ is fixed, $q=0$, the filter is stationary, and $\eps/\sigma$ has a fixed law $\mathcal L(\xi)$.
\end{assumption}
To describe the local contribution to the limiting distribution, consider the lazy random walk $(X_k)_{k\ge0}$ on $\mathbb Z^d$ with the transition rule in Assumption~\ref{ass:tori}. Write $p_k(x)=\Pp(X_k=x\mid X_0=0)$ for its transition probability from the origin to $x$ in $k$ steps. Let $\xi_{k,x}$ be independent copies of the standardized innovation $\xi=\eps/\sigma$, and define
\begin{equation}\label{eq:transientlaw}
 \mathcal H_d=-\sum_{k\ge0}\sum_{x\in\mathbb Z^d}p_{k+1}(x)\xi_{k,x},\qquad
 R_d=\sum_{k\ge0}p_{2k+2}(0).
\end{equation}
The index $k+1$ includes the mixing step in the innovation's own ATC round. By symmetry, $\sum_xp_{k+1}(x)^2=p_{2k+2}(0)$. For $d\ge3$, these return probabilities are summable, so the series defining $\mathcal H_d$ converges in $L^2$, with $\E\mathcal H_d=0$ and $\Var(\mathcal H_d)=R_d$.
\begin{theorem}[Consensus--local transition]\label{thm:transientlaw}
Under Assumption~\ref{ass:tori}, let $Y_n=(\widehat\theta_{i,n}-\theta^\star)/(\eta_n\sigma\sqrt{S_{2,n}})$. If $\varrho=0$, then $Y_n\Rightarrow\N(0,1)$. If $0<\varrho\le\infty$, then
\[
 Y_n\Rightarrow\frac{\mathcal H_d+\sqrt{\tau_\varrho}Z}{\sqrt{R_d+\tau_\varrho}},
 \qquad \tau_\varrho=(2h\varrho)^{-1},\quad\tau_\infty=0,
\]
where $Z\sim\N(0,1)$ is independent of $\mathcal H_d$. The limit is non-Gaussian whenever $\xi$ is.
\end{theorem}
The local field is related to stationary harness processes \citep{hammersley1967,ferrari2006,seppalainen2015}. In the pure local regime $\varrho=\infty$, the client's own latest innovation has the largest coefficient, $p_1(0)=1/2$, and carries $1/(4R_d)$ of the variance; for $d=3$, $R_3\approx.856$, about 29\%. The same limits hold for finite endpoints when $\eta_nT_n\to\infty$ and the initial term is negligible (Appendix~\ref{app:newlimits}). The filter also fixes the direction of skewness: with $\gamma_\eps=\E(\eps/\sigma)^3$,
\begin{equation}\label{eq:skewness}
 \E Y^3=-\gamma_\eps\frac{\sum_{k,j}c_{kj}^3}{S_2^{3/2}},
\end{equation}
and since $c_{kj}\ge0$, positive innovation skewness gives negative skewness of $Y$. Because the limit depends on the whole innovation law, moments cannot pin it down:
\begin{proposition}[Moment-based references]\label{prop:moments}
Under Assumption~\ref{ass:tori} with $0<\varrho\le\infty$, let $\xi$ and $\xi'$ be distinct compactly supported standardized innovation laws with the same first $r\ge2$ moments, and let $Y_n,Y_n'$ be the corresponding standardized endpoints. Then $\distK\{\mathcal L(Y_n),\mathcal L(Y_n')\}\to\delta_r>0$, and any references $G_n$ that depend on the innovation law only through its first $r$ moments satisfy
\[
 \max\{\distK(\mathcal L(Y_n),G_n),\distK(\mathcal L(Y_n'),G_n)\}\ge\delta_r/2-o(1).
\]
\end{proposition}
This covers Gaussian references ($r=2$) and Edgeworth and Cornish--Fisher references of every fixed order \citep{bhattacharya1976,hall1992}. Since $\distK$ is a supremum over levels, a moment correction can still be accurate at one level, as CF$_3$ is at 2.5\% in Section~\ref{sec:experiments}; the obstruction concerns calibration across levels. The hybrid reference of Section~\ref{sec:hybrid} avoids it by resampling influential innovations from empirical law.

\section{A reference that preserves influential noise}\label{sec:hybrid}
We keep the innovation law in the influential terms and approximate the rest by a Gaussian. Choose a deterministic set $\calJ$ of time--client indices from the graph and a prespecified nominal curvature $h_0$, with $D=|\calJ|$ and complement $\calR$. Split $Y=-\sum_{\calJ}b_{kj}\xi_{kj}-\sum_{\calR}b_{kj}\xi_{kj}$ with $\xi_{kj}$ \iid as $\eps/\sigma$. Replacing the second sum by a Gaussian of the same variance $v=\sum_{\calR}b_{kj}^2>0$ gives the oracle reference
\begin{equation}\label{eq:oracle}
G_0=\mathcal L\left(-\sum_{\calJ}b_{kj}\xi_{kj}+\sqrt vZ\right),
\end{equation}
with $Z\sim\N(0,1)$ independent of the retained innovations. Its replacement error is controlled by the Berry--Esseen ratio
\begin{equation}\label{eq:tailquantities}
\mathcal E_{\calR}=\beta_3\frac{\sum_{\calR}|b_{kj}|^3}{v^{3/2}}.
\end{equation}
\paragraph{Fitting and evaluating the reference.}
At a fixed pilot point $x_p$, typically $x_p=\widehat\theta_i$, collect $M$ fresh gradients $r_\ell=h(x_p-\theta^\star)+\eps_\ell$. Centering removes the unknown mean, $r_\ell-\bar r=\eps_\ell-\bar\eps$, so we estimate the innovation law and variance by $\widehat F=M^{-1}\sum_\ell\delta_{r_\ell-\bar r}$ and $\widehat\sigma^2=M^{-1}\sum_\ell(r_\ell-\bar r)^2$; residuals stored along the trajectory are an alternative (Appendix~\ref{app:curvature}). Project a curvature estimate onto a known bracket $[h_-,h_+]\ni h$ with $h_->0$ and $\eta h_+\le1$ to obtain $\widehat h$, and set $\widehat c=c(\widehat h)$. On the scale of~\eqref{eq:filter}, the feasible reference is
\begin{equation}\label{eq:feasible}
 \widehat S=-\sum_{\calJ}\widehat c_{kj}\eps^*_{kj}
 +\widehat\sigma\|\widehat c_{\calR}\|_2Z,
 \qquad \eps^*_{kj}\stackrel{\mathrm{iid}}\sim\widehat F.
\end{equation}
with $Z\sim\N(0,1)$ independent of the resampled innovations given the data. Draw $\Xi_r^*=-\sum_{\calJ}\widehat c_{kj}\eps_{kj,r}^*$ independently for $r=1,\ldots,B$ and integrate the Gaussian remainder analytically, a Rao--Blackwellization (RB) of the simulated CDF \citep{casella1996}. With $\widehat s=\widehat\sigma\|\widehat c_{\calR}\|_2>0$,
\begin{equation}\label{eq:rb}
 \widetilde F_B(x)=B^{-1}\sum_{r=1}^B\Phi\{(x-\Xi_r^*)/\widehat s\},
\end{equation}
whose quantiles $\widehat q_p$ give the equal-tail confidence interval for $\theta^\star$
\begin{equation}\label{eq:ci}
 [\widehat\theta_i-\eta\widehat q_{1-\alpha/2},\;
  \widehat\theta_i-\eta\widehat q_{\alpha/2}].
\end{equation}
Integration reduces the conditional CDF variance at every point, and the Dvoretzky--Kiefer--Wolfowitz (DKW) inequality \citep{massart1990} bounds the conditional CDF error by $t_B=\sqrt{\log(2/\zeta_B)/(2B)}$ outside probability $\zeta_B$.

\paragraph{Calibration error.}
Four sources contribute: the Gaussian replacement, estimation of the innovation law and curvature, memory of the initialization, and Monte Carlo error. Let $\Wass_2$ denote the minimum root-mean-squared discrepancy over couplings.
\begin{assumption}[Retention and nuisance accuracy]\label{ass:nuisance}
(i) The retained set $\calJ$ is deterministic and $v>0$. (ii) For deterministic $\omega_F,\delta_c\in[0,1]$, with probability at least $1-\zeta$ the fitted residual variance is positive and
\begin{equation}\label{eq:nuisance}
 \Wass_2(\widehat F,F_\eps)/\sigma\le \omega_F,\qquad
 \|\widehat c-c\|_2/\sqrt{S_2}\le\delta_c.
\end{equation}
(iii) The computed CDF \eqref{eq:rb} is within $t_B$ of its conditional expectation outside conditional probability $\zeta_B$. (iv) The initial term satisfies $|a^T(W^{T+q}e_0)_i|/(\eta\sigma\sqrt{S_2})\le u_0$, with $u_0=0$ for stationary filters. The estimates in (ii) may depend on the endpoint trajectory.
\end{assumption}
\begin{theorem}[Feasible coverage]\label{thm:hybrid}
Under Assumptions~\ref{ass:filter} and \ref{ass:nuisance}, let $X=(\widehat\theta_i-\theta^\star)/\eta$. The coverage error of \eqref{eq:ci} and the one-sided errors
\begin{equation}\label{eq:onesided}
 |\Pp\{X<\widehat q_{\alpha/2}\}-\alpha/2|,\qquad
 |\Pp\{X>\widehat q_{1-\alpha/2}\}-\alpha/2|
\end{equation}
are each at most $C_\alpha$ times
\begin{equation}\label{eq:coverbound}
 \mathcal E_{\calR}+\frac{\omega_F+\delta_c+u_0}{\sqrt v}+t_B+\zeta+\zeta_B.
\end{equation}
\end{theorem}
The first term is the Berry--Esseen error of the omitted sum; centering makes coupled estimation errors accumulate through squared coefficients, and Gaussian smoothing turns them into the second term. Because every fitted reference is compared with the same fixed oracle $G_0$, the fitted quantiles may depend arbitrarily on the endpoint; the guarantee holds marginally over repeated sampling (Appendix~\ref{app:hybrid}).

\begin{corollary}[Distributional calibration]\label{cor:calibrationcurve}
Under the assumptions of Theorem~\ref{thm:hybrid}, the probability integral transform $U=\widetilde F_B(X)$ satisfies $\sup_{0\le p\le1}|\Pp\{U\le p\}-p|\le C$ times \eqref{eq:coverbound}.
\end{corollary}
The probability integral transform (PIT) compares empirical with nominal probabilities at every level, as in the evaluation of probabilistic forecasts \citep{dawid1984,diebold1998,gneiting2007}. If the two tail deviations at level $\alpha/2$ are $\Delta_L$ and $\Delta_U$, the central coverage error is $-(\Delta_L+\Delta_U)$, so opposite errors cancel (Appendix~\ref{app:calibrationcurve}); this is why coverage alone cannot diagnose the problem.

\paragraph{Retention, curvature and rates.}
Top-$D$ selection at nominal curvature keeps the largest filter weights, the most recent innovations at the client and its neighbors, and minimizes the omitted cubic energy at a fixed count; complete lags $k<K$ are a simpler alternative. On lazy tori with $d\ge3$, nominal top-$D$ retention with $\eta D^{2/(d+2)}$ bounded leaves cubic energy $O(D^{-2(d-1)/(d+2)}+m^{-2}\min(T,\eta^{-1}))$ (Proposition~\ref{prop:budget}), and once the filter row is prepared the reference costs $O(BD)$ (Appendix~\ref{app:resources}). Curvature error $r_h=|\widehat h-h|$ gives $\delta_c\le Cr_h/h_-$, and stored residuals admit a pathwise contamination bound (Appendix~\ref{app:curvature}). For bounded innovations a centered second-order distance can replace the Wasserstein term (Corollary~\ref{cor:ordertwocalibration}), sharpening the sufficient rate on three-dimensional tori from $(\log M/M)^{5/24}$ to $(\log M/M)^{5/14}$.

\section{Nonlinear and vector objectives}\label{sec:nonlinear}
For $\theta^\star\in\R^p$ ($p$ is the parameter dimension, $d$ the graph dimension), consider a fixed unit contrast $u^\top\theta^\star$. With a common Hessian $H$, the innovation vector at lag $k$ has coefficient $(W^{k+q+1})_{ij}(I-\eta H)^ku$. The hybrid reference draws a complete innovation vector for each retained index, preserving dependence between coordinates, and uses its empirical covariance for the Gaussian remainder; Appendix~\ref{app:vector} gives directional calibration and Hessian estimation without commutation of $H$, $\widehat H$ and the noise covariance.
\begin{assumption}[Samplewise contraction]\label{ass:contraction}
Inputs $\mathsf Z_{j,t}$ are independent over clients and times. The stochastic gradient satisfies $\E g(x,\mathsf Z)=\nabla f(x)$ and $\nabla f(\theta^\star)=0$; its sample Jacobian is symmetric with spectrum in $[\mu,L_g]$, $0<\mu\le L_g$; $\nabla^2f$ is $M_3$-Lipschitz; and $\|g(\theta^\star,\mathsf Z)\|\le B_*$ almost surely.
\end{assumption}
Bounded-feature ridge logistic regression satisfies Assumption~\ref{ass:contraction}, and samplewise contraction makes DSGD an iterated random contraction with a unique stationary law \citep{diaconis1999}. Write $H=\nabla^2f(\theta^\star)$, $A=I-\eta H$, $\eps_{j,t}=g(\theta^\star,\mathsf Z_{j,t})$ and $\Sigma=\Cov(\eps_{j,t})\succ0$. The nonlinear and linear errors obey
\begin{equation}\label{eq:nonlinearrecursions}
 \begin{aligned}
 e_{i,t+1}&=\textstyle\sum_jW_{ij}\{e_{j,t}-\eta g(\theta^\star+e_{j,t},\mathsf Z_{j,t+1})\},\\
 z_{i,t+1}&=\textstyle\sum_jW_{ij}\{Az_{j,t}-\eta\eps_{j,t+1}\}.
 \end{aligned}
\end{equation}
Let $\mathcal S_i=\sum_{k\ge0}(1-\eta\mu)^{2k}(W^{2k+2})_{ii}$ be the contracted local energy, $\chi_i=\max_j\mathcal S_j/\mathcal S_i$ the local energy ratio, $s_u^2=\Var(u^\top z_i)/\eta^2$, and $\|X\|_r=(\E|X|^r)^{1/r}$.
\begin{theorem}[Nonlinear fixed-contrast transfer]\label{thm:vectortransfer}
Under Assumptions~\ref{ass:filter}(i) and \ref{ass:contraction} with fixed $p$, $q=0$ and $\eta L_g\le1/2$, couple the stationary errors in \eqref{eq:nonlinearrecursions} through the same inputs. For every fixed unit $u$ and $r\ge2$,
\begin{equation}\label{eq:vectortransfermain}
 \frac{\|u^\top(e_i-z_i)\|_r}{\eta s_u}
 \le Cr\chi_i\eta\sqrt{\mathcal S_i}.
\end{equation}
If a fixed reference has density at most $f_0$ on the scale $\eta s_u$, the Kolmogorov distance from the nonlinear endpoint law to that reference exceeds the linear one by at most $C\psi(f_0\chi_i\eta\sqrt{\mathcal S_i})$, where $\psi(x)=\min\{1,x\log(e+1/x)\}$. The constant $C$ depends only on $\mu,L_g,M_3,B_*,\lambda_{\min}(\Sigma)$ and $p$.
\end{theorem}
The proof combines contraction of the population update with a Taylor expansion and martingale moment bounds \citep{pinelis1994} (Appendix~\ref{app:vectornonlinear}). Weighted vertex transitivity gives $\chi_i=1$, and $\chi_i$ stays bounded when $I-W$ is comparable to the Laplacian of a vertex-transitive lazy matrix, covering unequal link weights and missing edges, where local errors on tori with $d\ge3$ and $m\eta$ bounded below remain non-Gaussian (Corollary~\ref{cor:graphpersistence}). Appendix~\ref{app:vectorfinite} covers finite horizons, final gossip and fitted references.

\begin{figure*}[t]
\centering\includegraphics[width=\textwidth]{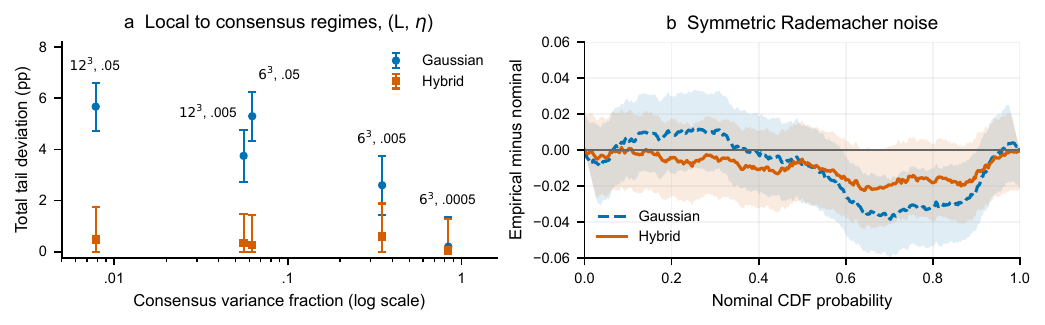}
\caption{Regime and symmetry checks, 4,000 trajectories per setting. Left: total tail deviation $\Delta_{\rm tail}=|p_L-.025|+|p_U-.025|$ against the consensus variance fraction $\rho(h)=\{1-a^{2T}\}/\{m(1-a^2)S_2(h)\}$ at $h=1.4$, labeled by torus size and stepsize, with conservative simultaneous 95\% intervals. Right: Rademacher noise on the $6^3$ torus with $\eta=.05$, with 95\% DKW bands.}\label{fig:empiricalregimes}
\end{figure*}

\section{Experiments}\label{sec:experiments}
\paragraph{Design.}
All experiments simulate the full network on lazy tori. Scalar quadratics use $h=1.4$ and unit-variance innovations: a skew law with probabilities $.1,.9$ at $3,-1/3$, a rare-shock law with probabilities $.02,.98$ at $7,-1/7$, a Gaussian-smoothed skew law, Gaussian noise, a standardized Beta$(4,7)$ law and Rademacher noise. Scalar ridge logistic regression has discrete root innovations. The main local cell is the $12^3$ torus with $\eta=.005$, where $m\eta=8.6$, consensus supplies $5.6\%$ of the variance and $\Lambda_i=.58$; the $6^3$ torus carries the five-law quadratic comparison with $\eta=.05$, $T=410$, and the scalar logistic comparison with $\eta=.08$, $T=2004$, and a regime grid varies $(L,\eta)$. All references share the endpoint, $M=400$ fresh gradients at it and a curvature estimate. \emph{Gaussian} uses the fitted variance $\eta^2\widehat\sigma^2S_2(\widehat h)$ with normal quantiles. \emph{CF$_3$} shifts both normal quantiles by the one-term Cornish--Fisher correction with the estimated filter skewness $-\widehat\gamma_\eps\sum\widehat c_{kj}^3/S_2(\widehat h)^{3/2}$ from \eqref{eq:skewness}. \emph{Hybrid} retains the $D=128$ nominal top innovations and uses $B=1499$ draws. Paired comparisons use exact McNemar tests \citep{mcnemar1947} with Holm adjustment \citep{holm1979}; Appendix~\ref{app:allstudies} gives all settings and interval methods \citep{wilson1927,clopper1934}.

\paragraph{Coverage conceals tail errors.}
In the $12^3$ cell the Gaussian interval has $95.10\%$ coverage but places $4.33\%$ of errors below and $0.58\%$ above (Table~\ref{tab:maincalibration}). Hybrid gives $2.70\%/2.35\%$ at the same length; in paired comparisons its lower-tail error falls by $1.63$ points and its upper-tail error rises by $1.78$ points (exact McNemar $p<10^{-6}$ for both), while coverage changes by $-.15$ points ($p=.67$). Its PIT discrepancy is $.012$, versus $.057$ for Gaussian. Here $n_{\rm eff}\approx27,000$, and the endpoint root mean squared error (RMSE) is $.0043$, against $.0356$ for the one-step estimate $x-\bar g/\widehat h$ from the same 400 gradients, so calibrating the endpoint is far more informative than replacing it. On the $6^3$ torus with $\eta=.05$, where $n_{\rm eff}\approx370$, the two RMSEs are similar for the skew model (endpoint $.0366$, one-step $.0362$), while the one-step RMSE is $35\%$ larger in the logistic model with $\eta=.08$. On the same $6^3$ torus, Hybrid lowers the skew-model lower-tail error by $2.95$ points ($[2.43,3.47]$) and changes coverage by $.35$ points ($[-.38,1.08]$).
\begin{table}[t]
\caption{Scalar 95\% intervals, $M=400$ unless stated. Low/Up: \% of estimation errors below/above the fitted reference quantiles (equivalently, misses above/below the parameter interval); Cov.: coverage (\%); Len.: mean length; PIT: PIT discrepancy; MAE: mean absolute one-sided error (points) at levels $.5$--$10\%$ in both tails; reuse: 1,600 residuals from the client's own trajectory. Tail MCSE $.25$ points.}\label{tab:maincalibration}
\begin{center}\small\setlength{\tabcolsep}{3.2pt}
\begin{tabular}{lrrrrrr}
\toprule
Reference&Low&Up&Cov.&Len.&PIT&MAE\\
\midrule
\multicolumn{7}{l}{\emph{Skew; $12^3$ torus, $\eta=.005$; $N=4000$}}\\
Gaussian&4.33&0.58&95.10&.0168&.057&1.66\\
CF$_3$&2.83&2.48&94.70&.0168&.017&.35\\
Hybrid&2.70&2.35&94.95&.0170&.012&.13\\
\multicolumn{7}{l}{\emph{Skew; $6^3$ torus, $\eta=.05$; $N=4000$}}\\
Gaussian&5.43&0.13&94.45&.1447&.089&2.42\\
CF$_3$&2.63&2.53&94.85&.1447&.044&.76\\
Hybrid&2.48&2.73&94.80&.1453&.008&.12\\
\multicolumn{7}{l}{\emph{Rare shock; $12^3$ torus, $\eta=.005$; $N=3999$}}\\
Gaussian&4.25&0.60&95.15&.0167&.100&2.00\\
CF$_3$&2.58&11.20&86.22&.0167&.107&3.15\\
Hybrid&3.00&4.88&92.12&.0174&.025&1.08\\
Hybrid, reuse&2.43&2.73&94.85&.0177&.013&.15\\
\bottomrule
\end{tabular}
\end{center}
\end{table}

\begin{figure*}[t]
\centering\includegraphics[width=\textwidth]{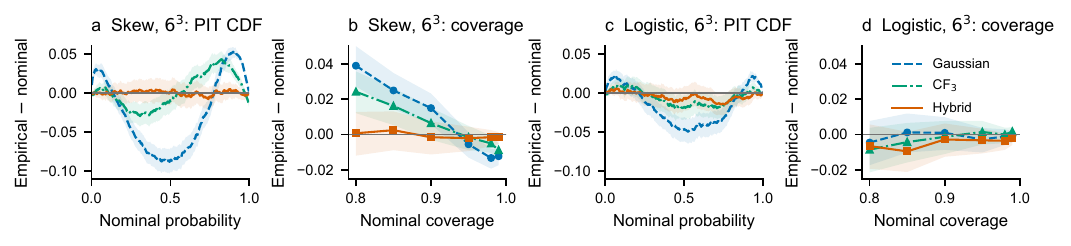}
\caption{Distributional calibration on the $6^3$ torus, 4,000 paired trajectories per model: empirical minus nominal PIT distribution function (a, c) and central coverage at levels 80--99\% (b, d), with pointwise 95\% Wilson intervals. Gaussian coverage is nearly nominal at 95\% only because its tail errors offset.}\label{fig:calibrationrow}
\end{figure*}

\paragraph{Moment corrections repair one level.}
CF$_3$ also balances the 2.5\% tails in the skew cells, where its paired tail differences from Hybrid are not significant after Holm adjustment, but its errors reappear at other levels (Figures~\ref{fig:anatomy}(b) and \ref{fig:calibrationrow}). In the $12^3$ cell, exact binomial tests at ten one-sided levels from $.5\%$ to $10\%$ reject calibration of CF$_3$ at the $.5\%$ and $10\%$ upper levels ($.93\%$ and $8.65\%$, Holm-adjusted $p=.007$ and $.04$) and of Hybrid at none. The mean absolute one-sided error of CF$_3$ exceeds that of Hybrid by $.22$ points on $12^3$ (paired bootstrap interval $[.01,.31]$) and by $.63$ points on $6^3$ ($[.31,.71]$), where the PIT discrepancies are $.044$ and $.008$. With exact cumulants, Cornish--Fisher references still have PIT discrepancy $.043$--$.044$ there, so the gap reflects the missing small parameter of Proposition~\ref{prop:moments} rather than estimation error. For the Gaussian-smoothed law, with half the filter skewness, CF$_3$ and Hybrid perform similarly (all cells, CF$_4$ and Edgeworth: Table~\ref{tab:momentall} in Appendix~\ref{app:momentexperiments}).

\paragraph{Rare large shocks.}
A shock of $+7$ with probability $.02$ (skewness $6.9$) on the same $12^3$ cell separates the references sharply (Table~\ref{tab:maincalibration}). The estimated Cornish--Fisher map is not monotone below the $97.5\%$ quantile, and CF$_3$ misses $11.2\%$ above; CF$_4$ and Edgeworth miss $14.1\%$ and $7.4\%$, and CF$_3$ still misses $7.7\%$ with $M=6400$. Hybrid has the smallest PIT discrepancy, $.025$ against $.100$ for Gaussian. With $M=400$ it covers $92.1\%$, because about eight shocks among 400 gradients make the law-estimation term $\omega_F$ of Theorem~\ref{thm:hybrid} large. The client's own trajectory supplies more law-estimation records without additional gradient evaluations, while retaining the 200 fresh gradient calls used to fit curvature: with its last 1,600 residuals, Hybrid's tails are $2.43\%/2.73\%$, as with 1,600 fresh gradients, while Gaussian stays at $4.13\%/.08\%$ and CF$_3$ still misses $8.8\%$ above (Appendix~\ref{app:rareshock}).

\paragraph{Regimes and controls.}
Tail distortion follows the predicted regime (Figure~\ref{fig:empiricalregimes}). On the $12^3$ torus with $\eta=.05$, Gaussian tails are $5.73\%/.05\%$ and Hybrid tails $2.55\%/2.93\%$. On the $6^3$ torus with $\eta=.0005$, where consensus supplies $.839$ of the variance, both references are nearly balanced (paired change in $\Delta_{\rm tail}$ $-.15$ points, $[-.35,.30]$). Under Gaussian, bounded beta and symmetric Rademacher noise, Hybrid matches or improves on the Gaussian reference. In the $6^3$ skew cell, $D=8$ to $512$ retained innovations give total tail deviations of $.25$--$.40$ points against $5.30$ for Gaussian. Complete lag-3 retention performs similarly; top-$D$ is five to thirteen times more accurate at population level (Table~\ref{tab:momentfloors}), but with $M=400$ both are limited by law estimation (budgets, reuse and final gossip: Appendices~\ref{app:revisionexperiments}--\ref{app:v3experiments}).

\paragraph{Nonlinear vector contrasts.}
A three-parameter ridge logistic model on the $6^3$ torus, with noncommuting Hessian and noise covariance, uses 4,000 trajectories at each of $\eta=.08,.04$ and two prespecified directions, $e_1$ and $u_2=(1,-1,1)/\sqrt3$ (Appendix~\ref{app:vectorexperiments}). For $e_1$ at $\eta=.04$, Hybrid reduces the total tail deviation by $1.18$ points at nearly equal length. Along $u_2$ the exact law is within $.01$ of Gaussian, leaving little shape to correct and no consistent gain. At $\eta=.08$ along $e_1$ Hybrid's upper error is $3.45\%$, against $2.93\%$ for the exact linear law on the same endpoints, so fitting the reference adds $.53$ points (paired exact test, $p<10^{-4}$) and the remainder is within Monte Carlo error.

\section{Discussion}\label{sec:discussion}
Communication determines which innovations remain visible in a client's uncertainty. The largest standardized weight $\Lambda_i$ separates two regimes: when it vanishes, Gaussian and moment-based references suffice; when it does not, calibration requires the innovation law, which the hybrid reference supplies with finite-sample control. In practice, $\Lambda_i$ and the certificate $\mathcal B_i$ of Section~\ref{sec:boundary} can be computed from $W$, $\eta$, a curvature bracket and a bound on $\beta_3$ before data are collected. When $\mathcal B_i$ is not small, about a hundred retained innovations restored tail balance in all our scalar local settings at a cost of milliseconds per endpoint, with a few hundred calibration gradients under moderate skewness and, under rare large shocks, 1,600 residuals from the client's own trajectory. The diagnostic also exposes a design trade-off: at a fixed self-weight $W_{ii}$, weights optimized for fast averaging \citep{xiao2004fast,boyd2006} reduce the local variance but not the weight of the client's own latest innovation, so $\Lambda_i$ grows.

The analysis assumes homogeneous objectives, independent inputs and a known, fixed communication matrix. Homogeneity is substantive: client offsets create a stationary bias of the order of the local fluctuations (Appendix~\ref{app:vectormodels}). Natural next steps are bias-corrected methods such as EXTRA, $D^2$ and gradient tracking \citep{shi2015extra,tang2018d2,pu2021distributed}, time-varying or directed communication \citep{nedic2015distributed,assran2019stochastic} and compressed gossip \citep{koloskova2019decentralized}, which change the filter, as well as data-driven retention and simultaneous inference across clients.
\label{mainend}
\section*{Use of generative AI}
We used generative AI tools for literature search, drafting and checking mathematical arguments, manuscript editing, experimental design, code development, and running and checking simulations.

We have reviewed all AI-assisted work. Experimental code and numerical results were executed and verified by the authors.

We take responsibility for the final content of this work, including all text, claims, proofs, code, experiments, and other artifacts produced with the aid of generative AI.
\bibliographystyle{plainnat}
\bibliography{references}
\clearpage
\appendix
\onecolumn
\numberwithin{equation}{section}
\setcounter{theorem}{0}\renewcommand{\thetheorem}{\Alph{section}.\arabic{theorem}}
\makeatletter\@addtoreset{theorem}{section}\makeatother
\renewcommand{\theHtheorem}{appendix.\Alph{section}.\arabic{theorem}}

\section{Roadmap, notation and algorithm}\label{app:roadmap}
Table~\ref{tab:roadmap} lists where each main-text result is proved and its main ingredients. Table~\ref{tab:notation} collects recurring symbols. Constants $C$ may change between displays.
\begin{table}[ht]
\caption{Main results, the assumptions they use, their proofs and key ingredients. A1--A4 denote Assumptions~\ref{ass:filter}--\ref{ass:contraction}.}\label{tab:roadmap}
\begin{center}\small
\begin{tabular}{@{}P{.27\textwidth}P{.13\textwidth}P{.14\textwidth}P{.38\textwidth}@{}}
\toprule Result & Assumptions & Proof & Key ingredients\\\midrule
Lemma~\ref{thm:boundary} (Gaussian boundary) & A1 & Appendix~\ref{app:tools} & Berry--Esseen; Lindeberg; Cram\'er's decomposition theorem\\
Corollary~\ref{cor:gap}, graph regimes, Table~\ref{tab:influence} & A1, spectral gap & Appendix~\ref{app:graphs} & Consensus split \eqref{eq:consensus}; heat-kernel bounds\\
Theorem~\ref{thm:transientlaw} (transition law) & A2 & Appendix~\ref{app:newlimits} & Proposition~\ref{prop:finitephase}; disjoint time windows; Lemma~\ref{lem:toruskernel}\\
Proposition~\ref{prop:moments} (moment-based references) & A2, bounded laws & Appendix~\ref{app:momentfloors} & Nonnegative weights; cumulant identity; moment determinacy\\
Theorem~\ref{thm:hybrid}, Corollary~\ref{cor:calibrationcurve} & A1, A3 & Appendices~\ref{app:hybrid}, \ref{app:calibrationcurve} & Oracle bound \eqref{eq:oracleBE}; centered coupling; Lemmas~\ref{lem:gaussianperturb} and \ref{lem:quantile}\\
Retention budget, Proposition~\ref{prop:budget} & Lazy tori, $d\ge3$ & Appendix~\ref{app:newlimits} & Heat-kernel level sets; nominal-to-true coefficient comparison\\
Second-order bounds, Corollaries~\ref{cor:ordertwocalibration}--\ref{cor:ordertworates} & A1, A3(i,iii,iv) & Appendix~\ref{app:ordertwo} & Lemma~\ref{lem:ordertwo}; Proposition~\ref{prop:ordertworef}\\
Curvature and residual reuse & A1 & Appendix~\ref{app:curvature} & Derivative bound \eqref{eq:curvature}; pathwise coupling \eqref{eq:reuse}\\
Theorem~\ref{thm:vectortransfer} (nonlinear transfer) & A1(i), A4 & Appendix~\ref{app:vectornonlinear} & Proposition~\ref{prop:vectormoments}: bounded differences, mean contraction, $L^r$ martingale bound\\
Corollary~\ref{cor:graphpersistence} (graph perturbations) & A1(i), A4, Laplacian comparison & Appendix~\ref{app:energycomparison} & Proposition~\ref{prop:energycomparison}\\
\bottomrule
\end{tabular}
\end{center}
\end{table}
\begin{table}[ht]
\caption{Recurring notation.}\label{tab:notation}
\begin{center}\small
\begin{tabular}{p{.3\textwidth}p{.62\textwidth}}
\toprule Symbol & Meaning\\\midrule
$m$, $W$, $d$, $L$ & number of clients, communication matrix, torus dimension, torus side length\\
$\eta$, $h$, $a=1-\eta h$ & stepsize, curvature at the target, scalar contraction factor\\
$T$, $q$ & gradient updates, communication-only rounds after the last update\\
$c_{kj}$, $S_r$, $b_{kj}=c_{kj}/\sqrt{S_2}$ & filter coefficients \eqref{eq:coeff}, their power sums, standardized weights\\
$s_i=\eta\sigma\sqrt{S_2}$, $\Lambda_i$ & local standard deviation, largest standardized weight \eqref{eq:lambda}\\
$\mathcal B_i$, $\bar\beta_3$, $[h_-,h_+]$ & curvature-uniform Gaussian certificate of Section~\ref{sec:boundary}, bound on $\beta_3$, curvature bracket\\
$\mathcal H_d$, $R_d$, $\varrho$, $\tau_\varrho$ & lattice local field, return energy, limit of $m\eta$, limiting consensus variance\\
$\calJ$, $\calR$, $D$, $v$ & retained and residual index sets, retained count, residual variance share\\
$\mathcal E_{\calR}$ & Berry--Esseen ratio of the residual sum \eqref{eq:tailquantities}\\
$G_0$, $\widehat S$, $\widetilde F_B$, $\Xi_r^*$ & oracle reference, fitted reference, computed reference CDF, retained draws\\
$M$, $B$, $h_0$, $\widehat h$, $x_p$ & calibration observations, reference draws, nominal and fitted curvature, pilot\\
$\omega_F$, $\delta_c$, $u_0$ & law, coefficient and initialization errors in Theorem~\ref{thm:hybrid}\\
$t_B$, $\zeta$, $\zeta_B$ & reference-simulation radius and failure probabilities\\
$Q$, $\widehat Q$, $\mathfrak d_2$, $z_{2,M}$ & true and fitted standardized laws, centered second-order distance, its bound\\
$\Delta_{\rm tail}$ & total tail deviation $|p_L-.025|+|p_U-.025|$\\
$p$, $u$, $H$, $A$, $\Sigma$ & parameter dimension, contrast, Hessian, $I-\eta H$, innovation covariance\\
$\mathsf Z_{j,t}$, $\mu$, $L_g$, $M_3$, $B_*$ & data inputs and the constants of Assumption~\ref{ass:contraction}\\
$\mathcal S_i$, $\chi_i$, $f_0$ & contracted local energy, local energy ratio, density bound of a fixed reference\\
\bottomrule
\end{tabular}
\end{center}
\end{table}

\par\medskip\noindent\begin{minipage}{\textwidth}
\hrule\vspace{4pt}
\textbf{Algorithm 1: Local hybrid calibration}
\begin{enumerate}[leftmargin=1.4em,itemsep=1pt,topsep=3pt]
\item Fix the client, graph, horizon, nominal curvature $h_0$, retained set $\calJ$ and budgets $M,D,B$.
\item Collect and center $M$ gradients at the fixed pilot, or use stored residuals, and fit $\widehat h\in[h_-,h_+]$.
\item Evaluate the coefficients; the reference is undefined if the residual variance is nonpositive.
\item Simulate $B$ retained sums, integrate the remainder by \eqref{eq:rb} and invert at $\alpha/2$ and $1-\alpha/2$ to obtain \eqref{eq:ci}.
\end{enumerate}
\vspace{2pt}\hrule
\end{minipage}\par\medskip

\section{Notation, independent-sum tools, and exact filters}\label{app:tools}
For distribution functions $F,G$, $\distK(F,G)=\sup_x|F(x)-G(x)|$. For distributions with finite $p$th moments, $\Wass_p(F,G)$ is the infimum of $(\E|X-Y|^p)^{1/p}$ over couplings. Constants denoted $C$ may change between displays; universal constants are distinguished from constants depending on the specified primitive parameter bracket. For a countable array $c$, $\norm c_r^r=\sum|c|^r$. Since $\norm c_2<\infty$, centered independent innovation sums are defined by their $L^2$ limits. All limits with changing graphs refer to a sequence of deterministic graph and algorithm parameters.

\subsection{Finite filters and initialization}
The finite filter energy has the spectral representation
\begin{equation}\label{eq:spectral}
 S_2=\sum_{\ell=1}^m u_\ell(i)^2\lambda_\ell^{2(q+1)}
 \frac{1-(a^2\lambda_\ell^2)^T}{1-a^2\lambda_\ell^2}
\end{equation}
The corresponding endpoint variance is $\eta^2\sigma^2S_2$. Here $\lambda_\ell,u_\ell$ are orthonormal eigenpairs of $W$; the numerator is one when $T=\infty$.

Iterating \eqref{eq:recursion}, commuting $aI$ with $W$, and applying $W^q$ gives
\[
 W^qe_T=a^TW^{T+q}e_0-\eta\sum_{k=0}^{T-1}a^kW^{k+q+1}\eps_{T-k}.
\]
Taking coordinate $i$ proves \eqref{eq:filter}. Orthogonality of independent centered innovations yields variance $\eta^2\sigma^2S_2$. Symmetry gives $\sum_j(W^{k+q+1})_{ij}^2=(W^{2k+2q+2})_{ii}$; expansion in the eigenbasis proves \eqref{eq:spectral}. Negative eigenvalues cause no problem because the variance expression contains their even powers. Entrywise nonnegativity implies the deterministic bound
\begin{equation}\label{eq:initbound}
 \frac{|a^T(W^{T+q}e_0)_i|}{\eta\sigma\sqrt{S_2}}
 \le\frac{a^T\norm{e_0}_\infty}{\eta\sigma\sqrt{S_2}}.
\end{equation}
If initialization is random, the same bound on a high-probability event contributes its failure probability to coverage. For fixed $\eta>0$, the stationary sum converges because $S_2\le\sum_{k\ge0}a^{2k}<\infty$. It is the unique invariant law by contraction. The case $a=0$ is a one-step filter and follows directly; derivative arguments below use $a>0$ and extend by continuity where appropriate.

\subsection{Proof of Lemma~\ref{thm:boundary}}
For a finite array, apply the independent, non-identically distributed Berry--Esseen inequality \citep{petrov1995} to the centered variables $-b_{kj}\eps_{j,T-k}/\sigma$, whose variances sum to one. The sum of their third absolute moments is $\beta_3\sum|b_{kj}|^3$. Since $\sum b_{kj}^2=1$, this is at most $\beta_3\Lambda_i$. For a countable array, apply the finite result to a truncation normalized by its own variance, and pass to the limit. The omitted sum converges in $L^2$, the variance tends to one, and the normal CDF is continuous. This proves \eqref{eq:be}. With $C\le.56$, the certificate $\mathcal B_i$ of Section~\ref{sec:boundary} dominates the right side of \eqref{eq:be}, because the true curvature lies in $[h_-,h_+]$ and $\bar\beta_3\ge\beta_3$; hence $\mathcal B_i$ bounds $\distK\{\mathcal L(Y),\N(0,1)\}$.

For the qualitative sufficiency statement, assume only $\E\xi^2=1$ and $\Lambda_i\to0$. For every $\epsilon>0$,
\[
 \sum_{k,j}\E\{b_{kj}^2\xi_{kj}^2\onevec (|b_{kj}\xi_{kj}|>\epsilon)\}
 \le\E\{\xi^2\onevec (|\xi|>\epsilon/\Lambda_i)\}\longrightarrow0.
\]
This is Lindeberg's condition. For countable rows, first truncate each row so its omitted variance tends to zero, apply the finite-row central limit theorem, and use the $L^2$ bound on the remainder.

For necessity, suppose a subsequence has $\Lambda_i\ge\delta>0$ and its normalized sum converges to a standard normal. On a further subsequence, choose an index with coefficient $b_n$ satisfying $|b_n|\ge\delta/2$ and $b_n\to b\ne0$. Write the sum as $-b_n\xi_n+R_n$, where $\xi_n$ has the same fixed innovation law and is independent of $R_n$. Since $\E R_n^2\le1$, the sequence $R_n$ is tight. Pass to a weakly convergent subsequence $R_n\Rightarrow R$. Product characteristic functions show that the sum converges to $-b\xi+R$, with $\xi$ independent of $R$. By assumption this distribution is standard normal. The Gaussian decomposition theorem of \citet{cramer1936} then forces $-b\xi$ to be Gaussian, contradicting the fixed non-Gaussian innovation law. This proves necessity without a fourth-moment assumption.

For Rademacher noise, an additional explicit diagnostic is
\begin{equation}\label{eq:kurtosis}
 \E Y^4=3-2\sum b_{kj}^4.
\end{equation}
Its nonvanishing excess kurtosis gives a direct special-case obstruction. The decomposition argument above is stronger: it also covers distributions with Gaussian fourth moment and non-Gaussian shape.

\subsection{Smoothing and quantile lemmas}
\begin{lemma}[Moment-to-CDF transfer]\label{lem:smooth}
Suppose $V$ has density bounded by $D$ and $(U,V)$ is a coupling with $\E|U-V|^p\le r^p$, $p\ge1$. Then
\[
 \distK\{\mathcal L(U),\mathcal L(V)\}\le
 \inf_{u>0}\{Du+r^p/u^p\}\le C_p(Dr)^{p/(p+1)}.
\]
More generally, if $\distK\{\mathcal L(V),H\}\le\beta$ and $H$ has density bounded by $D$, the upper bound is $\beta+C_p(Dr)^{p/(p+1)}$.
\end{lemma}
\begin{proof}
The inclusion $\{U\le x\}\subseteq\{V\le x+u\}\cup\{|U-V|>u\}$ gives $F_U(x)\le F_V(x+u)+r^p/u^p$ by Markov's inequality. The reverse bound uses $x-u$. Insert the density bound, or first compare $F_V$ to $H$, and optimize. No independence between $U$ and $V$ is required.
\end{proof}

\begin{lemma}[Random-reference quantiles]\label{lem:quantile}
Let $H$ be a fixed continuous strictly increasing CDF and $X$ a statistic with $\distK(\mathcal L(X),H)\le\beta$. Let $\widehat H$ be a random CDF such that $\norm{\widehat H-H}_\infty\le\delta$ on an event of probability at least $1-\zeta$. For $0<p_1<p_2<1$ and $\delta<\min(p_1,1-p_2)$, generalized quantiles satisfy
\[
 \left|\Pp\{\widehat H^{-1}(p_1)\le X\le\widehat H^{-1}(p_2)\}-(p_2-p_1)\right|
 \le C(\delta+\beta)+\zeta.
\]
For any $p\in(\delta,1-\delta)$, the same assumptions imply
\[
 |\Pp\{X<\widehat H^{-1}(p)\}-p|\le C(\delta+\beta)+\zeta,\qquad
 |\Pp\{X>\widehat H^{-1}(p)\}-(1-p)|\le C(\delta+\beta)+\zeta.
\]
The constant accounts for possible atoms of $X$; it can be taken universal. Neither $\widehat H$ nor its good event must be independent of $X$.
\end{lemma}
\begin{proof}
On the good event $H^{-1}(p-\delta)\le\widehat H^{-1}(p)\le H^{-1}(p+\delta)$, up to an arbitrarily small displacement for endpoint conventions. The random interval therefore lies between two deterministic intervals, whose probabilities under $\mathcal L(X)$ differ from their $H$-probabilities by at most $2\beta$; adding the bad event gives the first bound. Since $H$ is continuous, every atom of $X$ has mass at most $2\beta$, which handles closed endpoints and the strict inequalities of the one-sided statements. If $\delta\ge\min(p_1,1-p_2)$, the bound holds trivially with a larger constant.
\end{proof}

\section{Graph regimes and local return energy}\label{app:graphs}
For $q=0$ and a connected aperiodic symmetric stochastic matrix, separate the consensus eigenvalue $\lambda_1=1$, $u_1=m^{-1/2}\onevec$. Then
\begin{equation}\label{eq:consensus}
 S_2=\frac{1}{m(1-a^2)}+
 \sum_{k\ge0}a^{2k}\{(W^{2k+2})_{ii}-1/m\}.
\end{equation}
For a fixed graph, all remaining eigenvalues have magnitude less than one, so the second term stays bounded as $\eta\to0$ while the first is asymptotic to $(2m\eta h)^{-1}$. Since $|c_{kj}|\le1$, $\Lambda_i\to0$ follows. This is the consensus-dominated regime.

If \eqref{eq:return} holds uniformly, the second term in \eqref{eq:consensus} is at most $C_0$. When $m\eta\to\infty$, the first term tends to zero. The term $c_{0i}^2=W_{ii}^2\ge w_0^2$ bounds $S_2$ below, hence $\Lambda_i$ stays bounded away from zero. Lemma~\ref{thm:boundary} gives the non-Gaussian conclusion for every fixed non-Gaussian innovation law of finite variance. Spatial averaging cancels the non-consensus component, so its limiting behavior differs from that of a local coordinate.

\begin{corollary}[Uniform-gap networks]\label{cor:gap}
Let $T=\infty$, $q=0$, $h>0$ be fixed and $\eta h\le1$. Suppose $W_{ii}\ge w_0>0$ and $\max_{\ell\ge2}|\lambda_\ell(W)|\le1-g$ for fixed $g>0$. If $m\eta\to\infty$, then $\liminf\Lambda_i\ge w_0\sqrt{1-(1-g)^2}>0$, and every fixed non-Gaussian innovation law retains non-Gaussian local fluctuations.
\end{corollary}

\paragraph{Proof of Corollary~\ref{cor:gap}.}
Write $W=\sum_{\ell=1}^m\lambda_\ell u_\ell u_\ell^\top$ with $\lambda_1=1$ and $u_1=m^{-1/2}\onevec$. For every $k\ge0$,
\[
 0\le (W^{2k+2})_{ii}-\frac1m=\sum_{\ell\ge2}\lambda_\ell^{2k+2}u_\ell(i)^2
 \le\lambda_*^{2k+2}\sum_{\ell\ge2}u_\ell(i)^2\le\lambda_*^{2k+2},
\]
since the even powers are nonnegative and $\sum_\ell u_\ell(i)^2=1$. Summing over $k$ gives \eqref{eq:return} with $C_0=\lambda_*^2/(1-\lambda_*^2)\le1/\{1-(1-g)^2\}$. In \eqref{eq:consensus}, $a^{2k}\le1$ and the centered return terms are nonnegative, so $S_2\le\{m(1-a^2)\}^{-1}+C_0$. Because $\eta h\le1$, $1-a^2=\eta h(2-\eta h)\ge\eta h$, and the first term is at most $(m\eta h)^{-1}\to0$. Hence $\Lambda_i\ge b_{0i}=W_{ii}/\sqrt{S_2}\ge w_0/\sqrt{C_0+(m\eta h)^{-1}}$, whose limit inferior is at least $w_0\{1-(1-g)^2\}^{1/2}$. Lemma~\ref{thm:boundary} gives the non-Gaussian conclusion. For the lazy complete graph $W=I/2+\onevec\onevec^\top/(2m)$, every nonconsensus eigenvalue equals $1/2$.

\subsection{Lazy nearest-neighbor tori}
Let the graph be $(\mathbb Z/L\mathbb Z)^d$ and
\begin{equation}\label{eq:torus}
 W=\frac12I+\frac1{4d}\sum_{r=1}^d(P_r+P_r^{-1}),
\end{equation}
where $P_r$ shifts one lattice coordinate. Its eigenvalues, indexed by $u\in\{0,\ldots,L-1\}^d$, are
\[
 \lambda(u)=\frac12+\frac1{2d}\sum_{r=1}^d\cos(2\pi u_r/L),
\]
and each eigenvector has squared modulus $1/L^d$ at any vertex. Consequently
\[
 S_2=\frac1{L^d}\sum_u\frac{\lambda(u)^2}{1-a^2\lambda(u)^2}.
\]
For frequencies near zero, $1-\lambda(u)\asymp\norm{u/L}_{\mathrm{per}}^2$, and $1-a^2\lambda(u)^2\asymp\eta+\norm{u/L}_{\mathrm{per}}^2$ on a fixed positive curvature bracket. Frequencies bounded away from zero contribute a bounded amount. If $\eta\to0$ and $L\sqrt\eta\to\infty$, the grid resolves the region of radius $\sqrt\eta$, and annular summation compares the nonzero-frequency contribution to
\[
 \int_0^1\frac{r^{d-1}}{\eta+r^2}\,dr.
\]
For $d=1$ this has order $\eta^{-1/2}$, and the zero mode $1/(L\eta)$ is smaller by a factor $1/(L\sqrt\eta)$. For $d=2$ the integral has order $\log(1/\eta)$ and the zero mode tends to zero. For $d\ge3$ the integral is bounded above and below, and $1/(L^d\eta)\to0$. This proves the variance orders in Section~\ref{sec:boundary}. Since $W_{ii}=1/2$, the bounded-variance regime retains a visible current local innovation. In dimensions one and two, the diverging $S_2$ and bounded coefficients give Gaussian convergence.

For later use, Fourier summation also gives a uniform centered return bound
\begin{equation}\label{eq:hk}
 0\le (W^{2k+2})_{ii}-L^{-d}\le C_d(k+1)^{-d/2},\qquad k\ge0.
\end{equation}
Indeed $\lambda(u)^{2k+2}\le\exp[-c_d(k+1)\norm{u/L}_{\mathrm{per}}^2]$ for nonzero modes. Summing the Gaussian lattice bound gives \eqref{eq:hk} for $k\le L^2$; for $k>L^2$, factor out $\exp(-c_d k/L^2)$ and use the bound at order $L^2$. The inequality $\exp(-cx)\le C_dx^{-d/2}$ completes the uniform estimate. These are standard heat-kernel calculations \citep{lawler2010}. The invariant-field connection is related to the harness-process representation of \citet{seppalainen2015}; the present application uses the stepsize-discounted field to calibrate a local stochastic-optimization estimate.

\subsection{Influence diagnostics for common graphs}\label{app:influence}
Table~\ref{tab:influence} evaluates stationary filters with $h=1$ and $q=0$ at one client of five lazy graphs: a ring, two- and three-dimensional tori of side $m^{1/d}$, a random $4$-regular graph with neighbor weights $1/8$, and $W=I/2+\onevec\onevec^\top/(2m)$. Entries are computed exactly from graph powers. The first four graphs have $W_{ii}=1/2$; the lazy complete graph has $W_{ii}=1/2+1/(2m)$. In each case $W_{ii}$ is the largest coefficient, so $\Lambda_i=W_{ii}/\sqrt{S_2}$. Lemma~\ref{thm:boundary} turns small entries into a certificate: for the skew law of Section~\ref{sec:experiments} ($\beta_3=2.73$), the bound $.56\beta_3\sum|b_{kj}|^3$ is $.015$--$.025$ on all five graphs with $m=64$ and $\eta=.001$, but exceeds $.2$ on every graph except the ring with $m=4096$ and $\eta=.01$.
\begin{table}[ht]
\caption{Largest standardized influence $\Lambda_i$ and, in parentheses, the consensus share of the local variance, for stationary filters with $h=1$.}\label{tab:influence}
\begin{center}\small\setlength{\tabcolsep}{4pt}
\begin{tabular}{rrrccccc}
\toprule $m$&$\eta$&$m\eta$&Ring&2D torus&3D torus&4-regular&Complete\\\midrule
64 & .1 & 6.4 & .46 (.07) & .62 (.13) & .69 (.16) & .64 (.13) & .81 (.21)\\
64 & .01 & 0.64 & .24 (.18) & .39 (.49) & .44 (.60) & .41 (.54) & .48 (.71)\\
64 & .001 & 0.064 & .13 (.50) & .17 (.90) & .17 (.94) & .17 (.92) & .18 (.96)\\
729 & .1 & 72.9 & .46 (.01) & .62 (.01) & .70 (.01) & .65 (.01) & .88 (.02)\\
729 & .01 & 7.29 & .24 (.02) & .44 (.05) & .58 (.09) & .54 (.08) & .79 (.17)\\
729 & .001 & 0.729 & .13 (.05) & .33 (.31) & .42 (.49) & .41 (.45) & .50 (.67)\\
4096 & .1 & 409.6 & .46 (.00) & .62 (.00) & .70 (.00) & .65 (.00) & .89 (.00)\\
4096 & .01 & 40.96 & .24 (.00) & .44 (.01) & .59 (.02) & .55 (.02) & .85 (.04)\\
4096 & .001 & 4.096 & .13 (.01) & .35 (.06) & .53 (.14) & .51 (.13) & .74 (.27)\\
\bottomrule
\end{tabular}
\end{center}
\end{table}

\paragraph{The lattice constant.}
For the lazy walk on $\mathbb Z^3$, with step characteristic function $\phi(\theta)=1/2+\sum_{r=1}^3\cos\theta_r/6$, write $G(z)=(2\pi)^{-3}\int_{[-\pi,\pi]^3}\{1-z\phi(\theta)\}^{-1}d\theta$. Then $R_3=\sum_{k\ge0}p_{2k+2}(0)=\{G(1)+G(-1)\}/2-1$. Since $1-\phi$ is half the corresponding quantity for the simple random walk, $G(1)=2\times1.516386$ by Watson's integral \citep{watson1939}, and quadrature of the smooth integrand gives $G(-1)=.679623$. Hence $R_3=.85620$, in agreement with Riemann sums over tori of side 64 to 256. The client's own most recent innovation therefore has limiting variance share $1/(4R_3)=.292$, and $\Lambda_\infty=.540$.

\section{Joint graph--stepsize limits and retained-count guarantees}\label{app:newlimits}
\subsection{Finite-horizon consensus and local decomposition}
Let $P_n$ be the lazy nearest-neighbor transition matrix on a $d$-dimensional torus of side $L_n$, $d\ge3$, and let $m_n=L_n^d$.
Assume $L_n\to\infty$, $\eta_n\to0$, fixed $h>0$, $a_n=1-\eta_nh$, and $T_n\to\infty$, allowing $T_n=\infty$.
Innovations $\xi_{k,j}$ have a fixed centered unit-variance law and are independent.
Define the centered endpoint filter at error-over-$\eta\sigma$ scale by
\[
 X_n=-\sum_{0\le k<T_n,j}a_n^kP_n^{k+1}(i,j)\xi_{k,j}=C_n+D_n,
\]
\[
 C_n=-\sum_{k<T_n,j}\frac{a_n^k}{m_n}\xi_{k,j},\qquad
 D_n=-\sum_{k<T_n,j}a_n^k\{P_n^{k+1}(i,j)-m_n^{-1}\}\xi_{k,j}.
\]
The centered rows sum to zero, so $\operatorname{Cov}(C_n,D_n)=0$ and
\begin{equation}\label{eq:exactgamma}
 \tau_n:=\operatorname{Var}(C_n)=\frac{1-a_n^{2T_n}}{m_n(1-a_n^2)},\qquad
 V_n:=\operatorname{Var}(D_n)=\sum_{k<T_n}a_n^{2k}\{P_n^{2k+2}(i,i)-m_n^{-1}\},\quad
 S_{2,n}=\tau_n+V_n.
\end{equation}
For an infinite horizon set $a_n^{2T_n}=0$.
Zero covariance here is an energy identity; independence is proved below only in the limit.

Let $p_k(x)$ be the corresponding lattice transition kernel, and put
\[
 \mathcal H_d=-\sum_{k\ge0,x\in\mathbb Z^d}p_{k+1}(x)\xi_{k,x},\qquad
 R_d=\sum_{k\ge0}p_{2k+2}(0).
\]
The series converges in $L^2$ because $p_{2k+2}(0)\le C_d(k+1)^{-d/2}$ and $d>2$.
Its variance is $R_d>0$.

\begin{proposition}[Finite-horizon transition]\label{prop:finitephase}
Under the preceding assumptions, $V_n\to R_d$. If $\tau_n\to\gamma\in[0,\infty)$, then
\[
 (C_n,D_n)\Rightarrow(\sqrt\gamma Z,\mathcal H_d),\qquad Z\perp \mathcal H_d,\qquad
 X_n/\sqrt{S_{2,n}}\Rightarrow
 (\mathcal H_d+\sqrt\gamma Z)/\sqrt{R_d+\gamma}.
\]
If $\tau_n\to\infty$, then $S_{2,n}\sim\tau_n$ and $X_n/\sqrt{S_{2,n}}\Rightarrow \N(0,1)$.
These assertions require only finite innovation variance.
\end{proposition}
\begin{proof}
The torus Fourier estimate \eqref{eq:hk} is uniform in $L$:
\[
0\le P_L^{2k+2}(i,i)-L^{-d}\le C_d(k+1)^{-d/2}.
\]
For $k+1\le L^2$, this follows by a Gaussian lattice sum over the Fourier frequencies.
For $k+1>L^2$, the nonzero-mode contribution is at most
$C_dL^{-d}\exp\{-c_d(k+1)/L^2\}$, which obeys the same polynomial bound.
Consequently,
\begin{equation}\label{eq:centeredtail}
 \|D_n-D_n^{(<K)}\|_2^2\le
 C_d\sum_{k\ge K}(k+1)^{-d/2}\le C_d'K^{1-d/2}.
\end{equation}
For fixed $K$, sufficiently large tori have no wraparound for the first $K$ transitions.
These rows agree with the lattice rows, $a_n^k\to1$, and the early consensus subtraction has variance at most $K/m_n$.
Finite-window convergence and \eqref{eq:centeredtail} give $D_n\Rightarrow \mathcal H_d$.
Applying the same truncation to the variance sums separately proves $V_n\to R_d$.

For $0<\gamma<\infty$, the largest coefficient of $C_n$ is at most $1/m_n$.
Its Lindeberg sum at threshold $\epsilon$ is bounded by
$\tau_n\E\{\xi^2\onevec (|\xi|>\epsilon m_n)\}\to0$.
Thus $C_n\Rightarrow\sqrt\gamma Z$. When $\gamma=0$, $C_n\to0$ in $L^2$.
For infinite rows, truncate each row until its omitted variance tends to zero, then apply the finite-array CLT.

To establish joint independence, remove the first $K$ time blocks from $C_n$.
The resulting $C_n^{(\ge K)}$ and $D_n^{(<K)}$ depend on disjoint innovation blocks and are exactly independent.
Their approximation errors are bounded by $\sqrt{K/m_n}$ and $C K^{1/2-d/4}$.
Using $|e^{ix}-e^{iy}|\le|x-y|$, compare the joint characteristic function and the product of marginal characteristic functions with this independent pair.
The error is at most a constant times
$|s|\sqrt{K/m_n}+|t|K^{1/2-d/4}$.
First let $n\to\infty$, then $K\to\infty$. This proves the joint limit.

Finally, if $\tau_n\to\infty$, \eqref{eq:exactgamma} and bounded $V_n$ give $S_{2,n}\sim\tau_n$.
Every unnormalized coefficient of $X_n$ has magnitude at most one.
The largest normalized coefficient is therefore at most $S_{2,n}^{-1/2}\to0$.
Finite-variance Lindeberg, as in Lemma~\ref{thm:boundary}, proves the last assertion.
\end{proof}

\subsection{Proof of Theorem~\ref{thm:transientlaw} and interpretation}
For stationary filters, $\tau_n=\{m_n\eta_nh(2-\eta_nh)\}^{-1}$.
If $m_n\eta_n\to\varrho\in(0,\infty)$, it tends to $\tau_\varrho=(2h\varrho)^{-1}$.
The cases $\varrho=0$ and $\varrho=\infty$ give respectively $\tau_n\to\infty$ and $\tau_n\to0$.
Proposition~\ref{prop:finitephase} proves the three assertions.
For finite horizons the same stationary classification holds if $\eta_nT_n\to\infty$; a normalized initialization tending to zero then leaves the endpoint limit unchanged.
Specifically the required condition is
\begin{equation}\label{eq:phaseinit}
 \frac{|a_n^{T_n}(P_n^{T_n}e_{0,n})_i|}
 {\eta_n\sigma\sqrt{S_{2,n,T_n}}}\longrightarrow0.
\end{equation}
For bounded deterministic initialization, $T_n\ge C\eta_n^{-1}\log(1/\eta_n)$ with $Ch>1$ is sufficient because $S_{2,n,T_n}\ge1/4$.
The condition $\eta_nT_n\to\infty$ alone need not imply \eqref{eq:phaseinit}.

For finite time, \eqref{eq:exactgamma} is the more general classification.
If $m_n\eta_n\to\varrho\in(0,\infty)$ and $\eta_nT_n\to t$, then
$\tau_n\to(1-e^{-2ht})/(2h\varrho)$, including $t=0,\infty$.
For example, if $\eta=m^{-2}$ and $T=m$, then $m\eta\to0$ but $\tau_n\to1$.
If instead $T=\lfloor\sqrt m\rfloor$, $\tau_n\to0$.
These examples have $T\to\infty$ and show why finite running time enters the transition.

Since $p_1(0)=1/2$, $\mathcal H_d$ contains a nonzero multiple of one innovation independent of its $L^2$ remainder.
Cram\'er's theorem implies that $\mathcal H_d$ is non-Gaussian for a fixed non-Gaussian innovation law.
Applying the same theorem to $\mathcal H_d+\sqrt\gamma Z$ gives the finite-$\gamma$ non-Gaussian assertion.
Stationary nonlinear endpoints under the primitive assumptions share these linear limits when normalized by their linear energy, since the coupling in Theorem~\ref{thm:nonlinear} tends to zero.
For finite nonlinear chains the additional contraction term in \eqref{eq:rhot} must vanish.

The lattice field is related to the stationary harness height representation of \citet[Eq.~(2.24), Theorem 2.11]{seppalainen2015}.
Writing $\chi_t(x)=\sum_{k\ge0,z}p_k(x,z)\xi_{t-k}(z)$ gives $\mathcal H_d\overset{\mathcal L}=-(P\chi_t)(0)$.
The extra transition is due to adapt-then-combine.
Harness and harmonic-crystal constructions also connect random-walk representations to invariant fields \citep{ferrari2006,ferrari2007}.
Here the joint finite-torus and discount limit determines the independent consensus component and its exact variance.
For finite third absolute moment, the independent-sum moment inequality \citep{petrov1995} makes the truncated lattice series converge in $L^3$.
Its tail moment is bounded by a constant times the cubic coefficient tail plus the square-energy tail to the power $3/2$.
Expanding the finite-window cube and taking the limit justifies the skewness identity in the pure local regime.

\subsection{Moment-based references}\label{app:momentfloors}
\paragraph{Proof of Proposition~\ref{prop:moments}.}
Let $Y_\infty=(\mathcal H_d+\sqrt{\tau_\varrho}Z)/\sqrt{R_d+\tau_\varrho}$ be the limit under $\xi$ in Theorem~\ref{thm:transientlaw} and $Y'_\infty$ the limit under $\xi'$. Put $w_{k,x}=p_{k+1}(x)/\sqrt{R_d+\tau_\varrho}$. These weights are nonnegative and square summable, and $w_{0,0}=1/\{2(R_d+\tau_\varrho)^{1/2}\}>0$.

Suppose $|\xi|\le A_\xi$. Its moment generating function is entire and equals one at the origin, so $K_\xi(s)=\log\E e^{s\xi}$ is analytic on a disc $|s|<s_0$, with $K_\xi(0)=K_\xi'(0)=0$; hence $|K_\xi(s)|\le C|s|^2$ for $|s|\le s_0/2$. Because $0\le w_{k,x}\le1$, the series $\sum_{k,x}K_\xi(-sw_{k,x})$ converges uniformly on $|s|\le s_0/2$ to an analytic function. For real $s$, Hoeffding's inequality bounds $\E\exp\{2s\sum_{\rm finite}w\xi\}$ uniformly over finite partial sums, so the moment generating function of $Y_\infty$ is the limit of the finite products, and the series plus $\tau_\varrho s^2/\{2(R_d+\tau_\varrho)\}$ is the cumulant generating function of $Y_\infty$. Termwise differentiation at zero gives, for $j\ge3$,
\[
 \kappa_j(Y_\infty)=(-1)^j\kappa_j(\xi)\sum_{k,x}w_{k,x}^j,
 \qquad \sum_{k,x}w_{k,x}^j\ge w_{0,0}^j>0,
\]
since the Gaussian component contributes only to $\kappa_2$. Compactly supported laws are determined by their moments, and the first $n$ moments and the first $n$ cumulants determine each other. As $\xi$ and $\xi'$ differ but share their first $r$ moments, some $j>r$ has $\kappa_j(\xi)\ne\kappa_j(\xi')$, hence $\kappa_j(Y_\infty)\ne\kappa_j(Y'_\infty)$. Both limits have finite moment generating functions, so their laws differ and $\delta_r=\distK\{\mathcal L(Y_\infty),\mathcal L(Y'_\infty)\}>0$.

Both limits have continuous distribution functions. For $\varrho<\infty$ this follows from the independent Gaussian component. For $\varrho=\infty$, $Y_\infty$ is a convergent series of independent terms $-w_{k,x}\xi_{k,x}$, infinitely many with $w_{k,x}>0$; each such term has largest atom at most the largest atom of $\xi$, which is below one, so the series is continuous by L\'evy's theorem on series of independent variables \citep{levy1931}. Theorem~\ref{thm:transientlaw} gives $Y_n\Rightarrow Y_\infty$ and $Y'_n\Rightarrow Y'_\infty$, and P\'olya's theorem makes the convergence uniform in the distribution functions. Hence $\distK\{\mathcal L(Y_n),\mathcal L(Y'_n)\}\to\delta_r$. Finally, $G_n$ is the same function under both laws, because $W$, $\eta$, $h$, $T$, $q$ and the first $r$ moments agree, and the triangle inequality for the supremum norm gives the stated bound. The argument does not require $G_n$ to be monotone.

\paragraph{Two laws with four equal moments.}
A three-point law is determined by its first five moments when one exists. Take mean zero, variance one, third moment two and fourth moment eight, with fifth moment $27/4$ or $153/4$. To six decimals, the first law has atoms $\{-6.126719,-.344843,2.721562\}$ with probabilities $\{.001202,.884073,.114725\}$, and the second has atoms $\{-.514293,1.643048,5.621245\}$ with probabilities $\{.773317,.220333,.006350\}$; the moments are computed from the exact solutions. Inverting the exact characteristic functions of their stationary endpoints with $h=1.4$ gives Kolmogorov distance $.031$ on the $6^3$ torus with $\eta=.05$ and $.016$ on the $12^3$ torus with $\eta=.005$. Any reference using four moments therefore errs by at least $.015$, respectively $.008$, for one of the two laws.

\paragraph{Approximation floors.}
Table~\ref{tab:momentfloors} reports Kolmogorov distances to the exact stationary endpoint law under the skew innovation law, and in its last row under the rare-shock law of Appendix~\ref{app:rareshock}, without Monte Carlo error. The characteristic function is the exact product over all filter coefficients, inverted by the Gil--Pelaez formula on a grid of spacing $.01$ up to frequency 40; the same code reproduces $\Phi$ to $4\times10^{-16}$ from the Gaussian characteristic function. Moment corrections use the exact filter skewness and kurtosis. Oracle hybrid references use the exact law; top-$D$ counts are matched to complete lags $K=1,2,3$ (7, 32 and 92 terms on $6^3$, 7, 32 and 95 on $12^3$), and ties among equal weights are exchangeable. In the three local settings the moment corrections stall at $.013$--$.037$, whereas the hybrid reference approaches the exact law as $D$ grows; at matched counts of 32 or more, top-$D$ is five to thirteen times more accurate than complete-lag retention there. In the consensus-rich setting all corrections are accurate.
\begin{table}[ht]
\caption{Kolmogorov distance between the exact endpoint law and references with exact inputs ($h=1.4$, stationary filters; skew innovations, and rare-shock innovations in the last row). Lag $K$/top: complete-lag retention for $k<K$ and top-$D$ retention with the same count. $\rho$ is the consensus variance share. Where a Cornish--Fisher map is not monotone over the central 99\% range (CF$_4$ for $\eta=.05$, CF$_3$ in the last row), its monotone envelope is used.}\label{tab:momentfloors}
\begin{center}\small\setlength{\tabcolsep}{3.5pt}
\begin{tabular}{lccccccccc}
\toprule
$(L,\eta)$&$\Lambda_i$&$\rho$&Gaussian&CF$_3$&CF$_4$&Edgeworth&Lag 1/top&Lag 2/top&Lag 3/top\\\midrule
$(12,.005)$&.580&.056&.0572&.0169&.0132&.0135&.0210/.0067&.0084/.0016&.0039/.0006\\
$(12,.05)$&.677&.008&.0920&.0366&.0359&.0346&.0353/.0114&.0123/.0014&.0045/.0003\\
$(6,.05)$&.675&.062&.0911&.0361&.0352&.0340&.0347/.0112&.0121/.0014&.0044/.0004\\
$(6,.005)$&.513&.349&.0378&.0098&.0063&.0065&.0121/.0037&.0047/.0011&.0023/.0005\\
$(6,.0005)$&.252&.839&.0043&.0005&.0001&.0001&.0013/.0005&.0006/.0003&.0004/.0003\\
$(12,.005)$, rare&.580&.056&.0969&.1058&.0768&.0572&.0508/.0254&.0272/.0070&.0152/.0027\\
\bottomrule
\end{tabular}
\end{center}
\end{table}
With $D=128$, the oracle hybrid distances are $.0005$, $.0002$, $.0002$, $.0005$, $.0003$ and $.0021$ in the order of the table. Under the rare-shock law the moment corrections remain $.057$--$.106$ from the exact law, and CF$_3$ is farther from it than the Gaussian reference. The same comparison on the lazy complete graph $W=I/2+\onevec\onevec^\top/(2m)$, whose stationary weights are explicit, gives Gaussian, CF$_3$ and oracle top-128 distances of $.188$, $.135$ and $2.0\times10^{-4}$ for $m=1728$ and $\eta=.005$ ($\Lambda_i=.819$), and $.254$, $.244$ and $1.6\times10^{-4}$ for $m=4096$ and $\eta=.01$ ($\Lambda_i=.859$); both Cornish--Fisher maps are non-monotone there.

\subsection{A uniform off-diagonal torus kernel bound}\label{app:offdiagonal}
\begin{lemma}[Lazy-torus heat kernel]\label{lem:toruskernel}
For fixed $d\ge1$, $L\ge3$, $n\ge1$, and Euclidean distance between nearest torus representatives,
\[
 P_L^n(i,j)\le C_d\{n^{-d/2}
  \exp[-c_d\operatorname{dist}(i,j)^2/n]+L^{-d}\}.
\]
The constants depend only on $d$.
\end{lemma}
\begin{proof}
On $\mathbb Z^d$ the Fourier transform is
$\phi(\theta)=1/2+(2d)^{-1}\sum_r\cos\theta_r\in[0,1]$.
Since $1-\phi(\theta)\ge c_d|\theta|^2$ on $[-\pi,\pi]^d$, inversion gives $\sup_xp_n(x)\le C_dn^{-d/2}$.
The sum $S_\ell$ of the bounded centered steps obeys
$\Pp(|S_\ell|\ge R)\le2d\exp\{-R^2/(2d\ell)\}$, by the coordinate Hoeffding bound and a union bound.
For $n\ge2$, split the convolution $p_n=p_\ell*p_{n-\ell}$ with $\ell=\lfloor n/2\rfloor$.
Every summand has $|y|\ge|x|/2$ or $|x-y|\ge|x|/2$.
Combining the sup bound for one factor and the tail bound for the other gives
\[
 p_n(x)\le C_dn^{-d/2}e^{-c_d|x|^2/n}.
\]
For $n=1$, adjust the constant using finite support.

Periodization yields $P_L^n(0,[x])=\sum_{z\in\mathbb Z^d}p_n(x+Lz)$.
Choose $|x_r|\le L/2$. If $n\le L^2$, then for $|z_r|\ge1$,
$|x_r+Lz_r|\ge|x_r|$ and $|x_r+Lz_r|\ge L(|z_r|-1/2)$.
Its square is at least half the sum of these two squared lower bounds.
The coordinate Gaussian series is uniformly summable when $L^2/n\ge1$, giving
$\sum_z e^{-c|x+Lz|^2/n}\le C_de^{-c'|x|^2/n}$.
If $n\ge L^2$, the torus Fourier expansion gives
\[
 |P_L^n(i,j)-L^{-d}|
 \le L^{-d}\sum_{u\ne0}e^{-c_dn|u|^2/L^2}
 \le C_dL^{-d}e^{-c'_dn/L^2}.
\]
Combining the two time ranges proves the lemma.
\end{proof}

\subsection{Level sets and nominal top-\texorpdfstring{$D$}{D} retention}
\begin{proposition}[Nominal retained-count bound]\label{prop:budget}
Consider $d\ge3$ lazy tori with $L\ge3$, $q=0$, $T\ge1$ (finite or infinite), $h,h_0\in[h_-,h_+]$, and $\eta h_+\le1/2$.
Select the largest $D\ge1$ nominal coefficients, with deterministic ties, and assume $\eta D^{2/(d+2)}\le C_*$.
Then
\begin{equation}\label{eq:budgetbound}
 \sum_{\mathcal R_D}|b_{kj}(h)|^3
 \le C\{D^{-2(d-1)/(d+2)}+m^{-2}\min(T,\eta^{-1})\}.
\end{equation}
Constants depend only on $d$, the positive curvature bracket and $C_*$.
If $m\eta\ge c_0>0$, the current retained coefficient supplies uniform anchor density constants for a fixed smooth innovation law.
\end{proposition}
\begin{proof}
Put $u_{kj}=A_d(k+1)^{-d/2}\exp\{-c_d\operatorname{dist}(i,j)^2/(k+1)\}$.
For $0<t\le A_d$, write $K_t=(A_d/t)^{2/d}$.
The level set $u_{kj}>t$ has $n=k+1<K_t$ and lies in a ball of squared radius $dn\log(K_t/n)/(2c_d)$.
Nearest torus representatives form a subset of lattice points, so its size is at most
$C_d\{1+n^{d/2}[\log(K_t/n)]^{d/2}\}$.
Summing dyadic ranges $K_t2^{-\ell-1}<n\le K_t2^{-\ell}$ yields
\[
 N_u(t)\le C_dK_t^{1+d/2}=C_dt^{-(1+2/d)}.
\]
The constant point-count term handles balls of radius below one.
Thus the decreasing rearrangement satisfies $u_{(r)}\le Cr^{-d/(d+2)}$ and
\[
 \sum_{r>D}u_{(r)}^3\le CD^{-2(d-1)/(d+2)}.
\]

Let $a_0=1-\eta h_0$. Lemma~\ref{lem:toruskernel} implies
$c_{kj}(h_0)\le u_{kj}+A_da_0^k/m$.
Compare nominal top-$D$ with the $D$ largest $u_{kj}$, and use $(x+y)^3\le4(x^3+y^3)$:
\[
 \sum_{\mathcal R_D}c_{kj}(h_0)^3
 \le C D^{-2(d-1)/(d+2)}
       +\frac{C}{m^2}\sum_{k<T}a_0^{3k}
 \le C\{D^{-2(d-1)/(d+2)}+m^{-2}\min(T,\eta^{-1})\}.
\]
This comparison retains the discounted consensus remainder.

Take $K=\max\{1,\lceil D^{2/(d+2)}\rceil\}$.
For $k<K$, the mean-value bound on $\log(1-\eta u)$ gives
$c_{kj}(h)/c_{kj}(h_0)\le \exp(2\eta k|h-h_0|)\le C$ on nonzero entries.
For $k\ge K$, the kernel sup and square-sum bounds give
$\sum_jP_L^{k+1}(i,j)^3\le C\{m^{-2}+(k+1)^{-d}\}$.
Discounting and summing this tail gives $C\{m^{-2}\min(T,\eta^{-1})+K^{-(d-1)}\}$.
Separate omitted true coefficients into these early and late parts and use $S_{2,T}(h)\ge P_L(i,i)^2=1/4$ to obtain \eqref{eq:budgetbound}.
The argument also covers $T<K$, when the late part is empty.

The current coefficient equals $1/2$ and is a largest nominal coefficient, so it is retained.
Moreover $S_{2,T}(h)\le C\{1+(m\eta)^{-1}\}$ by the centered-return estimate.
Thus $m\eta\ge c_0$ bounds $b_{0i}=1/(2\sqrt{S_{2,T}(h)})$ away from zero and controls the anchor density constants.
\end{proof}

For a finite filter, $D$ is at most its positive coefficient count.
If all positive coefficients are retained, the cubic remainder is zero; the particular Gaussian-convolution implementation additionally requires a positive fitted residual variance.
For the three-dimensional smooth path in Corollary~\ref{cor:smoothrate}, take
\[
 D\asymp(M/\log M)^{5/16},\qquad B\ge C M^{1/2}\log M.
\]
Then \eqref{eq:budgetbound} is $O((\log M/M)^{1/4})$ and $\eta D^{2/5}\to0$.
The direct support-count comparison $D\ge\sum_{k<K}(2k+3)^d=O(K^{d+1})$ also proves a sufficient bound, giving $D\asymp(M/\log M)^{1/2}$ in dimension three.
Level-set counting improves this sufficient retained count; generating and sorting candidate coefficients remains a separate computational cost.

\paragraph{Finite-precision ranking.}
If every computed nominal score differs from its mathematical value by at most $\epsilon$, pair the omitted exact-top-$D$ entries with the retained replacements.
Each pair obeys $x\le y+2\epsilon$ and $0\le x,y\le1$, hence $x^3-y^3\le6\epsilon$.
At most $D$ exchanges give excess nominal cubic energy at most $6D\epsilon$.
The proof above then adds $CD\epsilon$ to \eqref{eq:budgetbound}.
The code uses fixed 14-decimal nominal ranking and is covered by the deterministic-set theorem.
The exact-top-$D$ rate applies to that implementation with this additional error term whenever the total score error, including kernel arithmetic, is bounded.

\section{Hybrid calibration and sharper smoothing}\label{app:hybrid}
Throughout this section, $\calJ$ is selected at a fixed nominal curvature. Its deterministic indices define the fixed oracle reference, while the selected coefficients are evaluated at the estimated curvature. Extend a finite array with zeros if necessary so that coefficient estimates are indexed on the same set.

\subsection{Smoothing by a retained innovation}
A smooth retained innovation can supply anti-concentration. Suppose one retained component $A=-b_{\ell_0}\xi_{\ell_0}$ has density $f_A$ with $\norm{f_A}_\infty\le L_0$ and $\norm{f_A''}_\infty\le L_2$.
\begin{proposition}[Retained smoothing]\label{prop:anchor}
Under Assumptions~\ref{ass:filter} and \ref{ass:nuisance}, the bound \eqref{eq:coverbound} of Theorem~\ref{thm:hybrid} can be replaced by a constant times
\begin{equation}\label{eq:anchor}
 \begin{aligned}
 &L_2(\beta_3+\E|Z|^3)\sum_{\calR}|b_{kj}|^3\\
 &\quad+\{L_0(\omega_F+\delta_c)\}^{2/3}+L_0u_0\\
 &\quad+t_B+\zeta+\zeta_B.
 \end{aligned}
\end{equation}
If also $\widehat c_{\ell_0}=c_{\ell_0}\ne0$ and $\distK(\widehat F,F_\eps)\le\kappa$ on the nuisance event, replace $\{L_0(\omega_F+\delta_c)\}^{2/3}$ by
\begin{equation}\label{eq:anchorswitch}
 \kappa+C L_0(\omega_F+\delta_c).
\end{equation}
Both versions are independent of inverse powers of $v$.
\end{proposition}
A third-order replacement argument controls the omitted cubic energy \citep{chatterjee2006}. For the linear switch, first replace the fitted anchor by its true law, then let that common anchor smooth the remaining perturbation. The current local coefficient is constant in curvature and supplies the common anchor. The proof below compares the implemented fitted reference through this auxiliary true-law convolution. Discrete innovations, including the logistic models, use the general Gaussian-remainder bound.

\subsection{Oracle tail replacement}
Partition the normalized innovation sum into $U=-\sum_{\calJ}b\xi$ and $V=-\sum_{\calR}b\xi$. The two variables are independent. Apply Berry--Esseen to $V/\sqrt v$ and convolve with the distribution of $U$; convolution cannot increase Kolmogorov distance. Hence
\begin{equation}\label{eq:oracleBE}
 \distK\{\mathcal L(Y),G_0\}\le C\mathcal E_{\calR}.
\end{equation}
The oracle reference has density bounded by $f_0=(2\pi v)^{-1/2}$, since it is a convolution with $\N(0,v)$. All arguments also hold for a countable residual array by truncation and $L^2$ convergence.

\begin{lemma}[Gaussian convolution perturbation]\label{lem:gaussianperturb}
Let $U,\widehat U$ have an integrable coupling, and let $Z$ be an independent standard normal. For $s,\widehat s>0$,
\[
 \distK\{\mathcal L(U+sZ),\mathcal L(\widehat U+\widehat sZ)\}
 \le \min\left\{1,\frac{\E|U-\widehat U|}{s\sqrt{2\pi}}
                      +\frac{|\widehat s-s|}{s}\right\}.
\]
\end{lemma}
\begin{proof}
Insert $\widehat U+sZ$. Conditioning on the coupled locations, the first comparison costs at most $\E|U-\widehat U|/(s\sqrt{2\pi})$.
For the second let $r=\widehat s/s$ and note
$|\partial_{\log r}\Phi(x/r)|\le(2\pi e)^{-1/2}$.
If $r\ge1$, $\log r\le r-1$; if $1/2\le r<1$, $-\log r\le2(1-r)$ and $2/\sqrt{2\pi e}<1$.
If $r<1/2$, the difference between the two centered normal CDFs is at most $1/2\le1-r$ because both CDFs are on the same side of $1/2$ at every $x$.
Thus $\sup_x|\Phi(x)-\Phi(x/r)|\le|1-r|$ in all cases.
Convolving with $\widehat U$ and applying the triangle inequality proves the result.
\end{proof}

\subsection{Centered empirical-law and coefficient coupling}
Condition on all calibration data on the event \eqref{eq:nuisance}. Couple $\eps^*\sim\widehat F$ and $\eps\sim F_\eps$ optimally in $L^2$; an arbitrarily close coupling is enough if needed. Because both laws are centered, the coupling difference has mean zero. Use independent copies of this pair for every retained index. On the true normalized scale, the retained-sum difference is
\[
 \frac{-\sum_{\calJ}\widehat c(\eps^*-\eps)
             -\sum_{\calJ}(\widehat c-c)\eps}{\sigma\sqrt{S_2}}.
\]
Independence and centering, followed by the triangle inequality, bound its $L^2$ norm by $(1+\delta_c)\omega_F+\delta_c$. The reverse triangle inequality in $L^2$ gives $|\widehat\sigma-\sigma|\le\Wass_2(\widehat F,F_\eps)$. Couple the two Gaussian tails with the same standard normal. Their normalized $L^2$ difference is at most
\[
 \left|\frac{\widehat\sigma}{\sigma}
       \frac{\norm{\widehat c_{\calR}}_2}{\sqrt{S_2}}-
       \frac{\norm{c_{\calR}}_2}{\sqrt{S_2}}\right|
 \le \omega_F(1+\delta_c)+\delta_c.
\]
Thus the full conditional coupling error is at most $C(\omega_F+\delta_c)$ for $\omega_F,\delta_c\le1$. Neither the cardinality of $\calJ$ nor the $\ell^1$ norm of the filter appears. Apply Lemma~\ref{lem:gaussianperturb} with $s=\sqrt v$ to the retained-sum coupling and the two positive Gaussian scales. Since $\|\cdot\|_1\le\|\cdot\|_2$, this proves
\begin{equation}\label{eq:conditionalcdf}
 \norm{\widehat H-G_0}_\infty
 \le C\min\{1,(\omega_F+\delta_c)/\sqrt v\},
\end{equation}
where $\widehat H$ is the CDF of \eqref{eq:feasible} divided by $\sigma\sqrt{S_2}$. This common normalization facilitates the proof; computation uses the unnormalized fitted reference. A countable residual filter is obtained by square-summable truncation. The finite retained sums are unchanged, the Gaussian scales converge, and their positive limiting scale gives continuous limiting CDFs. The same uniform bound therefore passes to the limit.

\subsection{Proof of Theorem~\ref{thm:hybrid}}
The initial shift in \eqref{eq:filter} has absolute normalized size at most $u_0$. Even if it is random and correlated with the innovation sum, a pathwise event sandwich and \eqref{eq:oracleBE} give
\[
 \distK\left\{\mathcal L\left(\frac{\widehat\theta_i-\theta^\star}{s_i}\right),G_0\right\}
 \le C\mathcal E_{\calR}+f_0u_0.
\]
On the nuisance and reference-Monte-Carlo good events, \eqref{eq:conditionalcdf} plus $t_B$ uniformly compares the computed CDF to the same fixed $G_0$. Apply Lemma~\ref{lem:quantile} with $p_1=\alpha/2$ and $p_2=1-\alpha/2$. The interval event is precisely that the standardized error lies between these computed reference quantiles. Add the two failure probabilities. This proves \eqref{eq:coverbound}; the single-quantile version of Lemma~\ref{lem:quantile} proves \eqref{eq:onesided}. The constant depends on $\alpha$ only through the trivial extension of Lemma~\ref{lem:quantile} to large errors. If the fitted residual variance vanishes, the procedure reports an undefined interval, which is counted as a failure and charged to $\zeta$.

\subsection{Proof of Proposition~\ref{prop:anchor}}
Write the retained sum as $A+O$, where $A=-b_{\ell_0}\xi_{\ell_0}$ has density $f_A$ and is independent of $O$ and all residual innovations. For fixed $O$, the function of the residual sum relevant to a CDF is $t\mapsto F_A(x-O-t)$. Its third derivative has absolute value at most $L_2$. Replace residual innovations one at a time by independent $b_{kj}Z_{kj}$. The first two Taylor terms cancel because means and variances match. Each remainder is at most
\[
 \frac{L_2}{6}|b_{kj}|^3\{\beta_3+\E|Z|^3\}.
\]
Integrate over $O$ and sum. This is a direct smooth-function Lindeberg argument, with \citet{chatterjee2006} providing broader context. For infinite arrays, apply it to finitely many replacements and use the continuity of the convolution with $A$ and $L^2$ convergence to pass to the limit. The resulting Gaussianized sum is exactly $G_0$.

Both the original sum and $G_0$ have a density bounded by $L_0$, because convolution does not increase that bound. Reuse the centered coupling above and Lemma~\ref{lem:smooth} with $p=2$, now with $D=L_0$. Initialization costs $L_0u_0$. The same deterministic-quantile argument proves \eqref{eq:anchor}. No lower bound on residual variance enters the comparison theorem; a positive residual variance is still used by the particular analytic-Gaussian inversion implementation.

For a normalized retained coefficient bounded away from zero and a fixed sufficiently smooth innovation density, $L_0,L_2$ can stay bounded as the retained set grows. Then the oracle approximation is controlled directly by residual cubic energy. With discrete innovations, such an anchor need not exist, and Theorem~\ref{thm:hybrid}'s Gaussian-tail anti-concentration remains the applicable result.

\paragraph{One-anchor CDF switch.}
For the stronger branch, condition on calibration data on its good event, always dividing references by the same true $\sigma\sqrt{S_2}$.
The equality $\widehat c_{\ell_0}=c_{\ell_0}$ makes the normalized anchor multiplier identical in both references.
Write the fitted reference as $\widehat A+\widehat V$ and the oracle as $A+V$.
Construct all anchor variables independently of a coupling of $V,\widehat V$.
For any nonzero multiplier $\gamma$, scaling both CDFs preserves their Kolmogorov distance.
For $\gamma<0$, this follows from $F_{\gamma X}(x)=1-F_X((x/\gamma)-)$, so the claim remains true for empirical atoms.
Consequently
\[
 d_K(\widehat A+\widehat V,A+\widehat V)\le\kappa,\qquad
 d_K(A+\widehat V,A+V)\le L_0\E|\widehat V-V|.
\]
The centered coefficient coupling from the preceding subsection, omitting the common anchor, bounds the latter expectation by $C(\omega_F+\delta_c)$.
This proves \eqref{eq:anchorswitch}. Initialization, residual replacement, reference simulation and the deterministic-quantile sandwich proceed unchanged.
If $f_\eps$ is the original innovation density and $\gamma=c_{\ell_0}/(\sigma\sqrt{S_2})$, then $L_0=|\gamma|^{-1}\|f_\eps\|_\infty$ and $L_2=|\gamma|^{-3}\|f_\eps''\|_\infty$.
For $q=0$ on the lazy torus, $c_{\ell_0}=c_{0i}=1/2=\widehat c_{0i}$ for every curvature estimate.
The extra CDF condition is part of the nuisance event; Wasserstein control alone is not substituted for it.

\subsection{Retention and Rao--Blackwellization proofs}
For any finite coefficient array, minimizing $\sum_{\calR}|c|^3$ subject to retaining exactly $D$ entries is equivalent to maximizing $\sum_{\calJ}|c|^3$, which is achieved by the $D$ largest magnitudes. For a square-summable countable array the largest $D$ magnitudes exist, except irrelevant ties and zeros, and the same exchange argument applies. Division by fixed $S_2^{3/2}$ preserves the claim. The standardized ratio $\mathcal E_{\calR}$ also depends on the residual variance, so its ordering can differ.

Conditional on calibration data, let $\Xi^*$ have the retained-sum distribution and $Z$ be independent standard normal. For each fixed $x$,
\[
 \E[\onevec\{\Xi^*+\widehat sZ\le x\}\mid \Xi^*]
 =\Phi((x-\Xi^*)/\widehat s).
\]
The law of total variance proves the variance reduction in \eqref{eq:rb}. If $G_B,G$ are the empirical and population CDFs of the retained draws, their convolutions with the fixed Gaussian law satisfy
\[
 \norm{\widetilde F_B-\widehat F_{\rm hybrid}}_\infty
 \le\norm{G_B-G}_\infty.
\]
Conditional DKW--Massart \citep{massart1990} gives the stated $t_B$ and $\zeta_B$. Numerical inversion adds at most the reference density bound times its spatial error, or its CDF residual when it stops on that criterion; both are separate from $t_B$, and Appendix~\ref{app:revisionexperiments} specifies and checks the solvers.

\section{Second-order comparison of scalar reference distributions}\label{app:ordertwo}
This section sharpens the empirical-law term for the scalar reference. It uses the same centered empirical law, its own variance, and the same deterministic retained set as the implemented procedure. Smooth probability metrics are classical \citep[Definition 3.1]{rio2009}; we give the integrated-CDF argument needed here explicitly.

\subsection{An integrated-CDF inequality}
For laws $Q,{\widehat Q}$ with finite second moments and equal means, define
\begin{equation}\label{eq:ordertwodef}
 K_{{\widehat Q},Q}(x)=\int_{-\infty}^x\{F_{\widehat Q}(t)-F_Q(t)\}\,dt,
 \qquad \mathfrak d_2({\widehat Q},Q)=\int_{\mathbb R}|K_{{\widehat Q},Q}(x)|\,dx.
\end{equation}
\begin{lemma}[Centered second-order comparison]\label{lem:ordertwo}
The distance in \eqref{eq:ordertwodef} is finite. For every $\phi\in C^2(\mathbb R)$ with bounded second derivative,
\begin{equation}\label{eq:ordertwotest}
 |\E_{\widehat Q}\phi-\E_Q\phi|\le\|\phi''\|_\infty\mathfrak d_2({\widehat Q},Q).
\end{equation}
Consequently, second moments differ by at most $2\mathfrak d_2({\widehat Q},Q)$, and scaling both laws by $a\in\mathbb R$ multiplies their distance by $a^2$.
\end{lemma}
\begin{proof}
Tonelli gives $K_{{\widehat Q},Q}(x)=\E_{\widehat Q}(x-Y)_+-\E_Q(x-X)_+$. Equal means also express $K$ as the difference of the right-tail expectations $\E(Y-x)_+$ and $\E(X-x)_+$. Use the left-tail expression for $x<0$ and the right-tail expression for $x\ge0$. Integrating their absolute bounds gives
\[
 \int|K(x)|\,dx\le\tfrac12(\E_{\widehat Q} Y^2+\E_Q X^2).
\]
Taylor's integral identity, valid for either sign of $y$, is
\[
 \phi(y)=\phi(0)+\phi'(0)y
 +\int_0^\infty(y-t)_+\phi''(t)\,dt
 +\int_{-\infty}^0(t-y)_+\phi''(t)\,dt.
\]
Finite second moments justify Fubini. Equal mass and means cancel the affine terms; the two representations then give $\E_{\widehat Q}\phi-\E_Q\phi=\int\phi''(t)K(t)\,dt$. This proves \eqref{eq:ordertwotest}. Take $\phi(y)=y^2$ for the moment bound. A change of variable proves the scaling identity for $a>0$; the right-tail representation proves it for $a<0$. Both laws coincide at zero when $a=0$.
\end{proof}
Equal means are essential: for ${\widehat Q}=\delta_1$ and $Q=\delta_0$, $K(x)=-1$ for $x\ge1$, so the integral diverges. Throughout this section empirical laws are centered exactly.

\subsection{Two reference-perturbation bounds}
Condition on the fitted inputs. Use the common true normalization
\[
 b=c/\sqrt{S_2},\quad \widehat b=\widehat c/\sqrt{S_2},\quad
 Q=\mathcal L(\eps/\sigma),\quad
 {\widehat Q}=\mathcal L(\eps^*/\sigma\mid\widehat F),\quad
 z_2=\mathfrak d_2({\widehat Q},Q).
\]
Thus $Q,{\widehat Q}$ have mean zero, $Q$ has variance one, and ${\widehat Q}$ has second moment $s_{\widehat Q}^2=\widehat\sigma^2/\sigma^2$. The common true scale provides a shared normalization for the proof. Let $\delta=\|\widehat b-b\|_2\le1$, $v=\|b_{\calR}\|_2^2>0$, and $\widehat v=\|\widehat b_{\calR}\|_2^2$. Compare
\[
 \widehat H=\mathcal L\left(-\sum_{\calJ}\widehat b_\ell\xi_\ell^*+s_{\widehat Q}\sqrt{\widehat v}Z\right),\qquad
 H_0=\mathcal L\left(-\sum_{\calJ}b_\ell\xi_\ell+\sqrt vZ\right).
\]
The retained set is deterministic and finite. Auxiliary innovations and $Z$ are independent conditional on the fitted inputs. No independence between those inputs and the endpoint is assumed.

\begin{proposition}[Second-order reference perturbation]\label{prop:ordertworef}
For the references above,
\begin{equation}\label{eq:ordertwogeneral}
 \distK(\widehat H,H_0)\le C\min\{1,z_2/v+\delta/\sqrt v\}.
\end{equation}
Suppose also that $\ell_0\in\calJ$, $\widehat b_{\ell_0}=b_{\ell_0}\ne0$, and the true anchor $A=-b_{\ell_0}\xi_{\ell_0}$ has a continuously differentiable density with $L_0=\|f_A\|_\infty$ and $L_1=\|f_A'\|_\infty$ finite. With $\kappa=\distK({\widehat Q},Q)$,
\begin{equation}\label{eq:ordertwoanchor}
 \distK(\widehat H,H_0)\le\kappa+L_1(1+\delta)^2z_2+2L_0\delta.
\end{equation}
The latter bound has no inverse residual-variance factor.
\end{proposition}
\begin{proof}
For \eqref{eq:ordertwogeneral}, the case $\delta\ge\sqrt v/2$ follows from $\distK\le1$ after increasing the universal constant. Otherwise $\sqrt{\widehat v}\ge\sqrt v/2$. First replace $s_{\widehat Q}\sqrt{\widehat v}Z$ by $\sqrt{\widehat v}Z$, leaving the fitted head unchanged. Gaussian scale comparison and convolution contraction bound the cost by
\[
 |s_{\widehat Q}-1|\le|s_{\widehat Q}^2-1|\le2z_2.
\]
The case $s_{\widehat Q}=0$ follows by the trivial bound, since then $2z_2\ge1$. Next replace retained innovation laws individually. Conditional on all other summands, the CDF as a function of the replaced innovation $y$ is $\Phi((x+\widehat b_\ell y-O)/\sqrt{\widehat v})$. Its second derivative is bounded by $\widehat b_\ell^2/(\widehat v\sqrt{2\pi e})$. Lemma~\ref{lem:ordertwo} bounds the total cost by
\[
 \frac{z_2\|\widehat b_{\calJ}\|_2^2}{\widehat v\sqrt{2\pi e}}
 \le \frac{4(1+\delta)^2z_2}{v\sqrt{2\pi e}}.
\]
All retained laws are now true. The residual standard deviations differ by at most $\delta$. Gaussian scale comparison costs $C\delta/\sqrt v$. With the Gaussian scale fixed at $\sqrt v$, couple the true centered innovations at both coefficient arrays. Their head difference has $L^2$ norm at most $\delta$, so Gaussian smoothing costs at most $\delta/\sqrt{2\pi v}$. Sum the terms, use $v\le1$, and cap the result at one.

For \eqref{eq:ordertwoanchor}, first replace only the fitted anchor by its true law. Its identical nonzero multiplier and convolution contraction give cost at most $\kappa$. For a negative multiplier use CDF left limits, whose supremum difference is also bounded by $\kappa$. Construct this common true anchor independently of the other auxiliary variables. Replacing each other retained law costs at most $L_1\widehat b_\ell^2z_2$ by Lemma~\ref{lem:ordertwo}.

For any independent $O$ and $t,s\ge0$, Gaussian convolution satisfies
\begin{equation}\label{eq:heatvariance}
 \|F_{A+O+\sqrt t Z}-F_{A+O+\sqrt s Z}\|_\infty
 \le\tfrac12 L_1|t-s|.
\end{equation}
For positive variance its derivative is $(f_A'*\mathcal L(O)*\phi_t)/2$, bounded by $L_1/2$. Bounded $f_A$ makes the CDF uniformly continuous, giving the same inequality at variance zero by continuity. Thus changing $s_{\widehat Q}^2\widehat v$ to $\widehat v$ costs at most $L_1z_2\widehat v$.

Finally keep the common anchor fixed and couple the remaining true innovations and the same Gaussian at both coefficient arrays. The head and Gaussian-tail differences each have $L^1$ norm at most $\delta$. Convolution with the anchor costs at most $2L_0\delta$. The squared coefficients used in law and variance replacement sum to at most $\|\widehat b\|_2^2\le(1+\delta)^2$, proving the result.
\end{proof}

\subsection{Calibration and centered empirical laws}
\begin{corollary}[Second-order feasible calibration]\label{cor:ordertwocalibration}
Let Assumption~\ref{ass:filter} and parts (i), (iii) and (iv) of Assumption~\ref{ass:nuisance} hold. Suppose a joint event of probability at least $1-\zeta$ gives $\delta\le\delta_c\le1$, $\mathfrak d_2({\widehat Q},Q)\le z_{2,M}$ and positive fitted residual variance. The numerical CDF event has radius $t_B$ and conditional failure probability $\zeta_B$. Then coverage, either one-sided error and the PIT-CDF discrepancy are bounded by a constant times
\begin{equation}\label{eq:ordertwocalibration}
 \mathcal E_{\calR}+\frac{z_{2,M}}v+\frac{\delta_c+u_0}{\sqrt v}+t_B+\zeta+\zeta_B.
\end{equation}
If the same joint event also bounds $\Wass_2({\widehat Q},Q)\le \omega_F\le1$, the law term may be replaced by $\min\{\omega_F/\sqrt v,z_{2,M}/v\}$.

Alternatively, suppose the common anchor conditions of Proposition~\ref{prop:ordertworef} hold, $f_A\in C^2$ with $L_2=\|f_A''\|_\infty<\infty$, and the joint event bounds $\distK({\widehat Q},Q)\le\kappa_M$. Then the error is bounded by a constant times
\begin{equation}\label{eq:ordertwosmoothcalibration}
 L_2(\beta_3+\E|Z|^3)\sum_{\calR}|b_\ell|^3
 +\kappa_M+L_1z_{2,M}+L_0(\delta_c+u_0)+t_B+\zeta+\zeta_B.
\end{equation}
\end{corollary}
\begin{proof}
The oracle replacement bounds are respectively Berry--Esseen and the third-order smooth replacement in Appendix~\ref{app:hybrid}. The fixed oracle densities are bounded by $1/\sqrt{2\pi v}$ and $L_0$, so initialization costs $Cu_0/\sqrt v$ and $L_0u_0$. Apply Proposition~\ref{prop:ordertworef} on the joint event, then add the numerical CDF error. The fixed-reference quantile sandwich of Appendix~\ref{app:tools} proves both one-sided assertions and their sum. The same comparison across quantile levels proves the PIT statement in Corollary~\ref{cor:calibrationcurve}. No conditioning of the endpoint's sampling law on the fitted data is used. Count all failed events in $\zeta+\zeta_B$; if separate metric bounds are available, take their event intersection before taking the smaller bound.
\end{proof}

\begin{lemma}[A bounded centered empirical law]\label{lem:empiricalordertwo}
Let $\xi_1,\ldots,\xi_M$ be iid with mean zero, variance one and $|\xi_t|\le a$. Write $F_M$ for their uncentered empirical law, ${\widehat Q}_M$ for its empirically centered version, $Q=\mathcal L(\xi_t)$, and $d_M=\|F_M-F_Q\|_\infty$. Then
\begin{equation}\label{eq:empiricalordertwo}
 \mathfrak d_2({\widehat Q}_M,Q)\le16a^2d_M.
\end{equation}
If $Q$ has bounded density $f_Q$, then also
\begin{equation}\label{eq:empiricalkappa}
 \distK({\widehat Q}_M,Q)\le(1+2a\|f_Q\|_\infty)d_M.
\end{equation}
With probability at least $1-2M^{-2}$, each applicable bound is $O(\sqrt{\log M/M})$.
\end{lemma}
\begin{proof}
The one-dimensional CDF identity gives $\Wass_1(F_M,Q)\le2ad_M$ and $|\bar\xi|\le\Wass_1(F_M,Q)$. Translation gives $\Wass_1({\widehat Q}_M,Q)\le4ad_M$. Both centered laws lie in $[-2a,2a]$, outside which their integrated CDF difference is zero. Thus $\mathfrak d_2({\widehat Q}_M,Q)\le4a\Wass_1({\widehat Q}_M,Q)$, proving \eqref{eq:empiricalordertwo}. Since $F_{{\widehat Q}_M}(x)=F_M(x+\bar\xi)$, a bounded true density gives \eqref{eq:empiricalkappa}. DKW with $d_M\le\sqrt{\log M/M}$ proves the event statement.
\end{proof}
The second-order bound also covers bounded discrete laws, for which the two metrics can have different orders. For an equal-weight law on $\{-1,1\}$, perturbing the upper mass by $\rho\in[0,1/2]$ and then centering gives $\mathfrak d_2=\rho$ but $\Wass_2^2=4\rho(1-\rho)$.

For reused records, apply the lemma first to the latent iid innovations. If centered residual and latent empirical laws have common bounded support of diameter $A_*$, their additional second-order discrepancy is at most $A_*\Wass_1$ of their paired contamination. The pathwise state-envelope and coefficient terms still enter separately. For nonlinear pilot gradients, empirical centering likewise leaves their state-dependent distributional error. The rate construction below uses fresh additive observations.

\subsection{Explicit scalar rates and their budgets}
\begin{corollary}[Fresh bounded-law rates]\label{cor:ordertworates}
Consider the additive scalar model on three-dimensional lazy tori with $q=0$. Fix $0<h_-\le h,h_0\le h_+$, a nondegenerate bounded innovation law, and uniformly bounded deterministic initialization. For integers $M\to\infty$, set
\[
 \eta=M^{-1},\quad L=M,\quad m=M^3,\quad
 T=\lceil C_0M\log M\rceil,\qquad C_0>4/h_-.
\]
Use $M$ fresh gradients at the chosen pilot for the centered law, and $M$ independent gradients at each of two finite-difference locations separated by two for curvature, projected to $[h_-,h_+]$. All three samples are fresh conditional on the pilot. Let the reference-draw failure probability be $M^{-2}$.

(i) Retain all nonzero coefficients with $k<K\asymp(M/\log M)^{2/7}$ and take $B\ge C M^{5/7}\log M$. Then the coverage, one-sided and PIT errors are $O((\log M/M)^{5/14})$.

(ii) Suppose the fixed innovation density is globally $C^2$ with bounded first two derivatives. Retain the exact nominal top-$D$ coefficients, where $D\asymp(M/\log M)^{5/8}$, and take $B\ge CM$. Then the same errors are $O(\sqrt{\log M/M})$.

Both statements use the general calibration procedure. They are sufficient growing-budget regimes for a non-Gaussian local limit.
\end{corollary}
\begin{proof}
On this path $m\eta=M^2$, $L\sqrt\eta=\sqrt M$, and the finite scalar energy is bounded above and below. Fresh additive gradients equal a common mean plus iid innovations; empirical centering removes the mean exactly, even at a random pilot. Lemma~\ref{lem:empiricalordertwo} gives $z_{2,M}=O(s_M)$ for $s_M=\sqrt{\log M/M}$. Independent bounded curvature samples and Hoeffding's inequality give $|\widehat h-h|=O(s_M)$, and hence $\delta_c=O(s_M)$ by the derivative bound in Appendix~\ref{app:curvature}. These events fail with probability $O(M^{-2})$. Lemma~\ref{lem:ordertwo} gives $s_{\widehat Q}^2\ge1-2z_{2,M}>0$ for sufficiently large $M$. Each retained set leaves positive fitted coefficients in its complement, so the fitted residual is positive on this event.

For (i), the lag bounds in Appendix~\ref{app:curvature} give $v_K\asymp K^{-1/2}$ and $\mathcal E_{\calR}\le CK^{-5/4}$. Their conditions hold: $\eta K\to0$, $K/T\to0$, $L/\sqrt K\to\infty$, $m\eta\gg K^{1/2}$ and $m^2\eta\gg K^2$. The leading terms in \eqref{eq:ordertwocalibration} are
\[
 K^{-5/4}+s_MK^{1/2}+s_MK^{1/4}.
\]
The first two have order $(\log M/M)^{5/14}$ at the chosen $K$, while the curvature term has order $(\log M/M)^{3/7}$. Normalized initialization is $O(M^{1-C_0h_-})$; multiplying by $v_K^{-1/2}$ leaves it smaller than the claimed rate. DKW with the stated $B$ gives $t_B=O(M^{-5/14})$; all failures are smaller. Nearest-neighbor support counting gives $D=O(K^4)$, hence a sufficient retained-sum generation budget $BD=O(M^{13/7}(\log M)^{-1/7})$ when $B\asymp M^{5/7}\log M$.

For (ii), the current coefficient $c_{0i}=1/2$ is retained and independent of curvature. Its standardized multiplier is bounded away from zero, so the fixed density gives uniform $L_0,L_1,L_2$. Lemma~\ref{lem:empiricalordertwo} also gives $\kappa_M=O(s_M)$. Proposition~\ref{prop:budget} yields
\[
 \sum_{\calR_D}|b_{kj}|^3\le C\{D^{-4/5}+M^{-5}\}=O(s_M),
 \qquad \eta D^{2/5}\to0.
\]
With $B\ge CM$, $t_B=O(s_M)$. Initialization and failures are again smaller, so \eqref{eq:ordertwosmoothcalibration} proves the claim. Bounded nondegenerate innovations are non-Gaussian, and the local-limit results apply since $m\eta\to\infty$.
\end{proof}

\paragraph{Summary of sufficient rates.}
On three-dimensional tori with fresh additive observations, $q=0$, $\eta=1/M$, $L=M$ and $T\asymp M\log M$, write $\ell_M=\log M/M$. Appendix~\ref{app:curvature}, Corollary~\ref{cor:smoothrate} and Corollary~\ref{cor:ordertworates} give the following sufficient error rates:
\[
 \begin{array}{c|cc}
 &\Wass_2&\mathfrak d_2\\\hline
 \text{bounded law}&\ell_M^{5/24}&\ell_M^{5/14}\\
 \text{bounded }C^2\text{ density}&\ell_M^{1/4}&\ell_M^{1/2}
 \end{array}
\]
These are sharper sufficient calibration bounds for the same reference construction, each under its own growing retention and reference budget. They do not compare methods at a fixed computational budget, and at fixed $D$ an oracle replacement term remains.

\paragraph{Statistical and computational scope.}
The smooth bound uses larger budgets than Corollary~\ref{cor:smoothrate}: its retained-sum cost is $BD=O(M^{13/8}(\log M)^{-5/8})$ when $B\asymp M$. These sufficient costs exclude network optimization, graph setup and CDF inversion. At fixed $D$, the nuisance comparison improves but the oracle replacement floor can remain. The nominal-curvature reuse rate in Appendix~\ref{app:curvature} is unchanged.

For approximate nominal ranking with score error at most $\epsilon_{\rm rank}$, Proposition~\ref{prop:budget} adds $CD\epsilon_{\rm rank}$. Part (ii) therefore requires
\[
 \epsilon_{\rm rank}=O\{(\log M/M)^{9/8}\}.
\]
This error includes kernel arithmetic as well as final rounding. The fixed numerical ranking in the experiments is covered by the deterministic-set theorem; the rate construction additionally requires the displayed precision.

\section{Curvature, reuse, and a feasible local regime}\label{app:curvature}
\subsection{Curvature and calibration observations}\label{app:calibrationobservations}
On $0<h_-\le h,\widehat h\le h_+$ with $\eta h_+\le1$, the filter satisfies
\begin{equation}\label{eq:curvature}
 \frac{\norm{\partial_h c(h)}_2}{\norm{c(h)}_2}\le\frac{\sqrt2}{h},
 \qquad\delta_c\le C\frac{|\widehat h-h|}{h_-}.
\end{equation}
The sensitivity constant depends on the curvature bracket alone. At a fixed parameter, centering fresh gradients in the additive model removes the unknown mean exactly. For bounded innovations, the elementary empirical-law bound $\omega_F=O_{\Pp}(M^{-1/4})$ suffices; sharper distribution-specific rates can also be used \citep{fournier2015}.

Existing gradient records can replace fresh samples. In the quadratic model record $r_t=g_t-h_0\theta_{i,t-1}$ at deterministic times, and center the empirical distribution. A pathwise coupling gives
\begin{equation}\label{eq:reuse}
 \begin{aligned}
 &\Wass_2(\widehat F_r,F_\eps)\le\Wass_2(\widehat F_\eps,F_\eps)\\
 &\quad+|h-h_0|\left\{M^{-1}\sum(e_{t-1}-\bar e)^2\right\}^{1/2}.
 \end{aligned}
\end{equation}
The right side couples recorded gradients to their latent independent innovations; only the left side is used for calibration.

For $q=0$ and $T=\infty$, under a uniform summable tail condition strengthening \eqref{eq:return}, $m\eta\to\infty$ and $\eta\to0$, a fixed positive $h_0$ can satisfy $\delta_c(h,h_0)\to0$ even without consistent curvature estimation. For three-dimensional lazy tori,
\begin{equation}\label{eq:nominal}
 \delta_c(h,h_0)\le C\{\eta^{1/4}+(m\eta)^{-1/2}\}.
\end{equation}
The derivation and a fixed-network counterexample appear below.

\subsection{Proof of the curvature derivative bound}
For $a=1-\eta h>0$ and $q\ge0$,
\[
 \norm{\partial_hc}_2^2
 =\eta^2\sum_{k=0}^{T-1}k^2a^{2k-2}(W^{2k+2q+2})_{ii}.
\]
In the eigenbasis, divide the contribution of each eigenvalue by its contribution to $S_2$. With $r=a^2\lambda^2$, this is
\[
 \frac{\eta^2}{a^2}
 \frac{\sum_{k=0}^{T-1}k^2r^k}{\sum_{k=0}^{T-1}r^k}
 \le\frac{\eta^2}{a^2}\frac{r(1+r)}{(1-r)^2}
 \le\frac{2\eta^2}{(1-a^2)^2}\le\frac2{h^2}.
\]
The first inequality follows by viewing the ratio as the second moment of a geometric distribution conditioned to be less than $T$; conditioning above by a cutoff cannot increase that second moment. For $T=\infty$ it is the usual geometric-series identity. Weighted summation over eigenmodes proves \eqref{eq:curvature}. At $a=0$, direct differentiation of the $k=1$ term or continuity gives the same bound.

For $h,\widehat h\in[h_-,h_+]$ with $\eta h_+\le1$, integrate the derivative bound. Gronwall's inequality, or integration of $|\partial_h\log\norm{c(h)}_2|\le\sqrt2/h$, first bounds the ratio of the two filter norms by a bracket-dependent constant. It then gives
\[
 \norm{c(\widehat h)-c(h)}_2/\norm{c(h)}_2
 \le C_{h_+/h_-}|\widehat h-h|/h_-.
\]
The same argument covers finite horizons and extra communication rounds, since the additional eigenvalue factors are nonnegative weights in the energy calculation.

\subsection{Proof of same-trajectory residual reuse}
For deterministic record times $t\in\mathcal I$, quadratic residuals satisfy
\[
 r_t=g_t-h_0\theta_{i,t-1}
 =\eps_{i,t}+(h-h_0)e_{i,t-1}-h_0\theta^\star.
\]
After empirical centering, the constant disappears. Pair each centered residual with the centered innovation at the same time. The mean squared coupling error is exactly $(h-h_0)^2M^{-1}\sum(e_{i,t-1}-\bar e)^2$. By the triangle inequality for $\Wass_2$, this proves \eqref{eq:reuse}. Empirical-distribution inequalities apply to the independent latent innovations at these deterministic record times, while the contamination bound handles their associated states pathwise.

If $\eps\in[-A,A]$, the one-dimensional identity $\Wass_1(F_M,F)=\int|F_M-F|$ and the bound $\Wass_2^2\le 2A\Wass_1$ give $\Wass_2(F_M,F)=O_{\Pp}(M^{-1/4})$ from DKW. Centering changes $F_M$ by at most $|\bar\eps|$, which is at most the uncentered $\Wass_2$ discrepancy because the true law has mean zero. Thus the same rate holds for the centered empirical law. This elementary bound is sufficient here; \citet{fournier2015} gives much broader empirical Wasserstein bounds.

\begin{lemma}[CDF control for contaminated residuals]\label{lem:reusecdf}
At $M$ fixed record times suppose $r_t=\eps_t+c+\delta_t$, with $\eps_t$ iid from a centered law $F_\eps$ of density bounded by $L_\eps$.
On the event $\max_t|\delta_t|\le A$, the centered empirical residual law obeys
\[
 d_K(\widehat F_r,F_\eps)
 \le d_K(F_M,F_\eps)+L_\eps(|\bar\eps|+2A),
\]
where $F_M$ is the uncentered empirical law of the latent innovations.
The contamination may depend on the entire trajectory.
\end{lemma}
\begin{proof}
For each paired observation,
$|(r_t-\bar r)-\eps_t|\le|\bar\eps|+2A=:D$.
Thus $F_M(x-D)\le\widehat F_r(x)\le F_M(x+D)$.
The density bound implies $|F_\eps(x\pm D)-F_\eps(x)|\le L_\eps D$, proving the pathwise claim.
DKW and concentration of $\bar\eps$ can now be applied to the latent iid sample, with the contamination-event failure probability added by a union bound.
\end{proof}
In the quadratic model take $A=|h-h_0|\max_t|e_{i,t-1}|$.
Under \eqref{eq:primitive}, the record envelope \eqref{eq:recordenvelope} gives
$A=(L_g+|h_0|)A_{M,\delta}$.
This provides the CDF event for smooth-innovation reuse when these radii are controlled. For discrete innovations the Wasserstein reuse bound and Theorem~\ref{thm:hybrid} remain applicable.

\subsection{When nominal curvature is enough}
Take $q=0$, $T=\infty$, and $h,h_0\in[h_-,h_+]$. Suppose the centered return energies have uniformly summable tails:
\begin{equation}\label{eq:uniformtail}
 \lim_{K\to\infty}\sup_m\sum_{k\ge K}
      \{(W_m^{2k+2})_{ii}-1/m\}=0,
 \qquad \inf_m(W_m)_{ii}>0.
\end{equation}
Assume $\eta\to0$ and $m\eta\to\infty$. Orthogonality of consensus and non-consensus subspaces yields
\[
 \norm{c(h)-c(h_0)}_2^2
 =\sum_{k\ge0}(a_h^k-a_{h_0}^k)^2
       \left\{\frac1m+(W^{2k+2})_{ii}-\frac1m\right\}.
\]
The consensus part is $O((m\eta)^{-1})$. For the non-consensus part, first fix $K$. Its first $K$ terms tend to zero; its remaining terms are bounded by four times the uniformly small tail in \eqref{eq:uniformtail}. Since $S_2(h)\ge W_{ii}^2$, the relative coefficient error tends to zero.

For three-dimensional lazy tori, $|a_h^k-a_{h_0}^k|\le C\min\{\eta k,1\}$ and \eqref{eq:hk} imply
\[
 \sum_{k\ge0}(a_h^k-a_{h_0}^k)^2
       \{(W^{2k+2})_{ii}-1/m\}
 \le C\sum_{k\ge0}\min\{\eta^2k^2,1\}(k+1)^{-3/2}
 \le C\eta^{1/2}.
\]
Add the consensus contribution and take square roots to obtain \eqref{eq:nominal}. In dimensions $3\le d<6$ the non-consensus norm has order at most $\eta^{(d-2)/4}$; at $d=6$ it is at most $\eta\sqrt{\log(1/\eta)}$, and for $d>6$ it is at most $\eta$.

This shortcut fails on a fixed connected graph. As $\eta\to0$, consensus dominates, and replacing $h$ by $h_0$ changes the leading variance by the factor $h/h_0$. The scale error persists with a known innovation law; compensating nuisance errors can improve finite-sample coverage without consistency.

\subsection{An explicit non-Gaussian consistency regime}
The following construction gives simultaneous control of approximation, nuisance and numerical errors. Consider $d\ge3$ lazy tori, $L\sqrt\eta\to\infty$, fixed positive curvature bracket, bounded innovations, and retain all coefficients at lags $k<K$. Standard heat-kernel bounds give $S_2\asymp1$ and, provided $K\to\infty$, $\eta K\to0$, and consensus terms are negligible,
\[
 v_K\asymp K^{-(d-2)/2},\qquad
 \mathcal E_K\lesssim K^{-(d+2)/4}.
\]
For the variance lower bound, use lags $K\le k<2K$ and torus frequencies with $|u|\le cL/\sqrt K$. If $K\to\infty$, $K\le T/2$, $L/\sqrt K\to\infty$, and $\eta K\to0$, this set contains at least $cL^dK^{-d/2}$ frequencies. Their eigenvalues obey $\lambda(u)\ge1-C/K$, so each contributes a fixed positive multiple to $\lambda(u)^{2k+2}a^{2k}$. Dividing by $L^d$ and summing the $K$ lags yields $v_K\gtrsim K^{1-d/2}$ since $S_2\asymp1$. The upper bound uses \eqref{eq:hk} and the discounted consensus energy.
The cubic estimate follows from $\sum_j p_k(j)^3\le\norm{p_k}_\infty\sum_jp_k(j)^2\lesssim k^{-d}$ and summation, followed by division by $v_K^{3/2}$. To make ``consensus terms are negligible'' explicit, one can impose $m\eta\gg K^{(d-2)/2}$ for the variance tail and $m^2\eta\gg K^{d-1}$ for the cubic tail, in addition to $L\sqrt\eta\to\infty$. These conditions are compatible, for example with $L=\eta^{-1}$ rounded to an integer.

For a concrete unconditional statement specialize to $d=3$, $\eta=M^{-1}$, $L=M$, bounded innovations in $[-A,A]$, and bounded deterministic initialization. Use $T=\lceil C_0M\log M\rceil$ with $C_0>4/h_-$ and retain lags $k<K$. DKW for the centered latent empirical law gives
\[
 \omega_{\rm emp}\le C(\log M/M)^{1/4}
\]
with failure probability at most $CM^{-2}$. For fresh curvature, take $M$ independent gradient observations at each of two fixed finite-difference locations separated by two and project onto the stable bracket. Hoeffding's inequality gives $|\widehat h-h|\le C\sqrt{\log M/M}$ with failure at most $CM^{-2}$; \eqref{eq:curvature} transfers this to coefficient error.

For the reuse alternative, choose fixed nominal $h_0\in[h_-,h_+]$. At each of the last $M$ deterministic record times the additive centered state is a sum of bounded independent innovations with squared coefficient sum at most $C\eta^2S_2(h)=O(\eta^2)$. Hoeffding followed by a union bound over records gives a maximum centered displacement $C\eta\sqrt{\log M}$ with failure at most $CM^{-2}$. The deterministic mean is bounded by $C(1-\eta h)^{T-M}$ and is smaller at the chosen horizon. Thus the contamination in \eqref{eq:reuse} is at most $C\eta\sqrt{\log M}$ on that event. The nominal coefficient discrepancy is deterministically $O(M^{-1/4}+M^{-1})$; the finite filter obeys the same bound, since its difference energy is bounded by the stationary difference energy and its variance is bounded below.

In either construction there are deterministic thresholds $\omega_{F,M},\delta_{c,M}\le C(\log M/M)^{1/4}$ and nuisance failure probability $\zeta_M\le CM^{-2}$. The leading coverage terms are therefore
\[
 K^{-5/4}+(\log M/M)^{1/4}K^{1/4}.
\]
Taking $K\asymp(M/\log M)^{1/6}$ gives $O((\log M/M)^{5/24})$. The normalized initialization is $O(M^{1-C_0h_-})$. Choose $B\ge C_1M^{5/12}\log M$ and $\zeta_{B,M}=M^{-2}$, giving $t_B=O(M^{-5/24})$. All remaining failure and initialization terms are smaller. Combining the deterministic radii and failure probabilities proves the unconditional coverage and one-sided calibration rate for this growing lag budget.

\subsection{A faster feasible balance with retained smoothing}\label{app:smoothrate}
\begin{corollary}[Smooth local calibration rate]\label{cor:smoothrate}
Consider the additive scalar model on a three-dimensional lazy torus with $q=0$. Let the calibration sample size be $M$, set $\eta=M^{-1}$, side length $L=M$, and $T=\lceil C_0M\log M\rceil$ with $C_0>4/h_-$. Assume a fixed positive curvature bracket, bounded deterministic initialization, and a fixed nondegenerate innovation law with bounded support and a twice continuously differentiable density whose first two derivatives are bounded. Use $M$ fresh centered innovation observations, a curvature estimate from $M$ independent observations at each fixed finite-difference location separated by two, and retain all lags $k<K=\lceil(M/\log M)^{1/8}\rceil$. With $B\ge C_1M^{1/2}\log M$ reference draws, the interval coverage and one-sided errors are $O((\log M/M)^{1/4})$, while the local endpoint retains a non-Gaussian limiting law.
\end{corollary}
\begin{proof}
Here $m=M^3$, $m\eta=M^2\to\infty$, and $L\sqrt\eta=\sqrt M\to\infty$. The graph calculation gives $S_2\asymp1$ and $b_{0i}=1/(2\sqrt{S_2})$ bounded away from zero. Thus the current local innovation supplies an anchor with uniformly bounded $L_0,L_2$ after standardization. The bounded law is non-Gaussian, so Lemma~\ref{thm:boundary} rules out Gaussian convergence.

Let $p_k(j)=(W^{k+1})_{ij}$. Fourier bounds imply
\[
 \norm{p_k}_\infty\le C\{m^{-1}+(k+1)^{-3/2}\},\qquad
 \sum_jp_k(j)^2\le C\{m^{-1}+(k+1)^{-3/2}\}.
\]
Therefore $\sum_jp_k(j)^3\le C\{m^{-2}+(k+1)^{-3}\}$. Discounting and summing over $k\ge K$ gives
\[
 \sum_{k\ge K,j}|b_{kj}|^3
 \le C\{K^{-2}+(m^2\eta)^{-1}\}=O((\log M/M)^{1/4}).
\]
This upper bound also holds for the finite filter. The empirical centered innovation discrepancy is $O((\log M/M)^{1/4})$ with failure probability $O(M^{-2})$ by DKW. Bounded independent gradient observations at the two finite-difference locations give curvature error $O(\sqrt{\log M/M})$ with the same order of failure probability, after projection onto the stable bracket. The derivative bound in \eqref{eq:curvature} transfers this to coefficient error. The initial normalized shift is $O(M^{1-C_0h_-})$, and the DKW reference error is $O(M^{-1/4})$ for the stated $B$, with failure probability $O(M^{-2})$. For the extra CDF event, DKW and bounded-support mean concentration give
\[
 d_K(\widehat F,F_\eps)
 \le d_K(F_M,F_\eps)+\|f_\eps\|_\infty|\bar\eps|
 \le C\sqrt{\log M/M}
\]
with failure probability $O(M^{-2})$.
The current local anchor has $\widehat c_{\ell_0}=c_{\ell_0}=1/2$, and its density constants are uniformly bounded. Insert these bounds into the switch branch \eqref{eq:anchorswitch}; all failure probabilities and initialization terms are smaller than $(\log M/M)^{1/4}$.
\end{proof}
A compactly supported smooth non-Gaussian law, for example a centered and standardized beta law with parameters $(4,4)$, satisfies the stated density conditions. This corollary gives a simultaneous feasible path with explicit nuisance events. Appendix~\ref{app:newlimits} also supplies a nominal top-$D$ sufficient count. The CDF event for smooth reused residuals can instead be built using Lemma~\ref{lem:reusecdf}, with its contamination radius retained.

\section{Primitive nonlinear and state-dependent extension}\label{app:nonlinear}
\subsection{Scalar transfer and finite-time statements}\label{app:scalarstatements}
The innovation filter also describes state-dependent gradients at leading order. Let $g(x,\mathsf Z)$ be a common scalar stochastic gradient with $\E g(x,\mathsf Z)=f'(x)$ and $f'(\theta^\star)=0$. Suppose, for all $x,z$,
\begin{equation}\label{eq:primitive}
 \begin{gathered}
 \mu\le\partial_xg(x,z)\le L_g,\qquad |f'''(x)|\le M_3,\\
 \sup_{x,z,z'}|g(x,z)-g(x,z')|\le2B_0.
 \end{gathered}
\end{equation}
For $q=0$, define $S_{i,T}(u)=\sum_{k=0}^{T-1}(1-\eta u)^{2k}(W^{2k+2})_{ii}$ and $S_{i,\infty}(u)$ by the infinite sum. These agree with $S_2$ at curvature $u$. The independent $\mathsf Z_{j,t}$ drive the actual nonlinear DSGD recursion. Put $h=f''(\theta^\star)$ and $\eps=g(\theta^\star,\mathsf Z)$, whose variance is $\sigma^2>0$.
\begin{theorem}[Primitive nonlinear transfer]\label{thm:nonlinear}
Let $W$ be vertex-transitive, symmetric, nonnegative and doubly stochastic, and $\eta L_g\le1/2$. Under \eqref{eq:primitive}, the nonlinear chain has a unique stationary law. Couple it to the quadratic chain with curvature $h$ using the same innovations. At any client, for $q=0$,
\begin{equation}\label{eq:nonlinear}
 \frac{\norm{e_{i,\infty}-z_{i,\infty}}_2}
 {\eta\sigma\sqrt{S_2(h)}}
 \le C\eta\sqrt{S_2(h)}\le C\sqrt\eta.
\end{equation}
The constant depends only on $\mu,L_g,M_3,B_0,\sigma$, not on the graph or its size. If a fixed hybrid reference on the unit-linear-standard-deviation scale has density bounded by $f_0$, this adds at most $C\{f_0\eta\sqrt{S_2(h)}\}^{2/3}$ to its Kolmogorov error, in addition to the linear reference error.
\end{theorem}
Theorem~\ref{thm:nonlinear} is the scalar, vertex-transitive case of Proposition~\ref{prop:vectormoments}; Appendix~\ref{app:finitecalibration} gives the reduction and the explicit concentration bounds used for residual reuse.

Conditional on a pilot $x_p$, the centered population law of $g(x_p,\mathsf Z)$ lies within $2L_g|x_p-\theta^\star|$ of $F_\eps$ in $\Wass_2$. Estimating that law adds empirical sampling error to the pilot contribution in $\omega_F$. Bounded-covariate, ridge-regularized logistic gradients satisfy \eqref{eq:primitive}, and provide a test case with both nonlinearity and state-dependent noise.

\begin{corollary}[Finite-time feasible nonlinear inference]\label{cor:finitecalibration}
Under Theorem~\ref{thm:nonlinear}, take a deterministic initial state and $q=0$. Use the finite-$T$ reference of Theorem~\ref{thm:hybrid}, with its nuisance event defined relative to $F_\eps=\mathcal L\{g(\theta^\star,\mathsf Z)\}$. Write $S_T=S_{i,T}(h)$, $S_\infty=S_{i,\infty}(h)$, and
\begin{equation}\label{eq:rhot}
 \rho_T=\frac{\eta S_\infty}{\sigma\sqrt{S_T}}
 +\frac{(1-\eta\mu)^T(\|e_0\|_\infty+2B_0/\mu)}{\eta\sigma\sqrt{S_T}}.
\end{equation}
If its fixed oracle reference has density bounded by $f_0$, the coverage and one-sided bounds acquire at most an additional $C(f_0\rho_T)^{2/3}$. The linear initialization term is retained. The same conclusion holds with Proposition~\ref{prop:anchor}'s bound and $f_0=L_0$.
\end{corollary}
A fresh pilot displacement $r_p$ contributes $2L_gr_p/\sigma$ to $\omega_F$ and $CM_3r_p/h_-$ to coefficient error. For reuse, a local record envelope controls contamination. Appendix~\ref{app:finitecalibration} derives the envelope and failure probabilities from bounded differences of the independent inputs. In transient tori $S_\infty\asymp1$, the stationary coupling contribution is $O(\eta^{2/3})$ in CDF distance for bounded $f_0$.

For smooth innovations, Lemma~\ref{lem:reusecdf} also controls the empirical CDF under state contamination.
The all-order bound of Proposition~\ref{prop:vectormoments}, through Corollary~\ref{cor:vectorfinite} with $p=1$, improves the nonlinear CDF contribution to $O(\eta\log(1/\eta))$ in the transient regime when the reference density is uniformly bounded.

\subsection{Proofs for the scalar primitive model}\label{app:finitecalibration}
Let $G_t(x)$ be the vector with coordinates $g(x_j,\mathsf Z_{j,t})$, so that $\theta_t=W\{\theta_{t-1}-\eta G_t(\theta_{t-1})\}$. Vertex transitivity means that a group of coordinate permutations acts transitively on vertices and preserves $W$. Write $S_i(u)=\sum_{k\ge0}(1-\eta u)^{2k}(W^{2k+2})_{ii}$ and $r=1-\eta\mu$.

\paragraph{Construction, sensitivity and concentration.}
For fixed inputs the maps $x\mapsto x-\eta g(x,z)$ are increasing with derivative in $[1-\eta L_g,r]$, so a nonnegative row-stochastic $W$ makes the update an $r$-contraction in the sup norm. Since $\E g(\theta^\star,\mathsf Z)=0$ and the oscillation of $g(\theta^\star,\cdot)$ is at most $2B_0$, one has $|g(\theta^\star,\mathsf Z)|\le2B_0$, and coupling from the past gives a unique stationary solution with $\|e\|_\infty\le2B_0/\mu$ \citep{diaconis1999}. Replacing one input $\mathsf Z_{j,-k}$, as in the functional-dependence approach of \citet{wu2005}, changes its update by at most $2\eta B_0W_{\cdot j}$ coordinatewise, and monotonicity propagates this to
\begin{equation}\label{eq:nonlinearsensitivity}
 |e_{i,0}-e_{i,0}^{(j,-k)}|\le2\eta B_0r^k(W^{k+1})_{ij}.
\end{equation}
The squared constants sum to $4\eta^2B_0^2S_i(\mu)$. The Efron--Stein inequality \citep{efron1981} gives $\Var(e_i)\le2\eta^2B_0^2S_i(\mu)$, and the bounded-differences inequality, applied to finite-past approximations and passed to the stationary limit by bounded convergence, gives
\begin{equation}\label{eq:localconcentration}
 \Pp\{|e_{i,\infty}-\E e_{i,\infty}|>x\}\le2\exp\{-x^2/(2\eta^2B_0^2S_i(\mu))\}.
\end{equation}
By vertex transitivity all clients share the stationary mean $m_e$. The stationary mean equation $\E f'(\theta^\star+e_i)=0$, a second-order Taylor expansion about $\theta^\star+m_e$ and strong monotonicity give $|m_e|\le M_3\Var(e_i)/(2\mu)\le(M_3/\mu)\eta^2B_0^2S_i(\mu)$.

\paragraph{Proof of Theorem~\ref{thm:nonlinear}.}
Condition \eqref{eq:primitive} implies \eqref{eq:vectorassumptions} with $p=1$ and $B_*=2B_0$, so Proposition~\ref{prop:vectormoments} applies. Under vertex transitivity $\mathcal S_{\max}=\mathcal S_i=S_i(\mu)$, so $\|e_i-z_i\|_2\le C\eta^2S_i(\mu)$. At each eigenvalue $\lambda$ of $W$ the ratio of the resolvent weights $\{1-(1-\eta h)^2\lambda^2\}/\{1-(1-\eta\mu)^2\lambda^2\}$ is largest at $\lambda^2=1$, where it equals $(h/\mu)(2-\eta h)/(2-\eta\mu)\le h/\mu$; hence $S_i(\mu)\le(h/\mu)S_i(h)$. Dividing by $\eta\sigma\sqrt{S_2(h)}$, with $S_2(h)=S_i(h)$, proves the first inequality in \eqref{eq:nonlinear}, and $S_i(h)\le\{1-(1-\eta h)^2\}^{-1}\le(\eta h)^{-1}$ gives the second. Lemma~\ref{lem:smooth} with $p=2$ proves the CDF statement. The constants depend only on $\mu,L_g,M_3,B_0$ and $\sigma$.

\paragraph{Record envelope.}
For $M$ deterministic record times no earlier than $t_0$, contraction towards the stationary chain, the mean bound and \eqref{eq:localconcentration} with a union bound give, with probability at least $1-\delta$,
\begin{equation}\label{eq:recordenvelope}
 \begin{split}
 |e_{i,t}|\le A_{M,\delta}:={}&r^{t_0}(\|e_0\|_\infty+2B_0/\mu)
       +(M_3/\mu)\eta^2B_0^2 S_i(\mu)\\
       &+\eta B_0\sqrt{2S_i(\mu)\log(2M/\delta)}.
 \end{split}
\end{equation}
With $M=1$ this bounds the endpoint pilot.

\paragraph{Proof of Corollary~\ref{cor:finitecalibration}.}
Couple the nonlinear and linear chains, started at the same deterministic $e_0$, with their jointly stationary same-input versions; the stationary supports are bounded by $2B_0/\mu$ and $2B_0/h$. Contraction, $1-\eta h\le r$ and Theorem~\ref{thm:nonlinear} give
\[
 \|e_{i,T}-z_{i,T}\|_2\le C\eta^2 S_{i,\infty}(h)+2r^T(\|e_0\|_\infty+2B_0/\mu).
\]
Dividing by $\eta\sigma\sqrt{S_{i,T}(h)}$ gives at most a constant times $\rho_T$ in \eqref{eq:rhot}. The finite linear statistic is within the oracle-replacement error plus $f_0u_{0,T}$ of $G_0$, and Lemma~\ref{lem:smooth} with $p=2$ adds at most $C(f_0\rho_T)^{2/3}$. The coupling of fitted and oracle references and the quantile sandwich of Appendix~\ref{app:hybrid} apply unchanged once the nuisance event holds relative to $F_\eps=\mathcal L\{g(\theta^\star,\mathsf Z)\}$.

For fresh samples at a pilot, compare centered gradients at $x_p$ with $g(\theta^\star,\mathsf Z)$ using the same latent inputs; this costs at most $2L_g|x_p-\theta^\star|$ in $\Wass_2$. If the centered latent empirical law has error at most $\omega_{\rm emp}\sigma$, $|x_p-\theta^\star|\le r_p$ and $|\widehat h-f''(x_p)|\le r_h$, one may take $\omega_F=\omega_{\rm emp}+2L_gr_p/\sigma$ and $\delta_c=C(r_h+M_3r_p)/h_-$. For residuals $g(\theta_{i,t-1},\mathsf Z_{i,t})-h_0\theta_{i,t-1}$ at deterministic record times, pairing with the latent innovations and \eqref{eq:recordenvelope} give $\omega_F\le\omega_{\rm emp}+2(L_g+|h_0|)A_{M,\delta}/\sigma$, with $\delta$ added to the failure probability. The latent innovations at these times are independent although the recorded states are not. Coefficient error is controlled by a curvature estimator or by the nominal-curvature bounds of Appendix~\ref{app:curvature}.

\subsection{Verification for the logistic experiment}
Let $X$ be uniform on $\{1,2\}$, $\lambda=.1$, $\theta^\star=.5$, and
\[
 p_x=\Pp(Y=1\mid X=x)=\operatorname{sigmoid}(x\theta^\star)
       +\lambda\theta^\star/\E X.
\]
Both probabilities lie strictly between zero and one. Use
\[
 g(\theta;X,Y)=\lambda\theta+X\{\operatorname{sigmoid}(X\theta)-Y\}.
\]
Then $\E g(\theta^\star;X,Y)=0$ exactly. Sample derivatives satisfy $\lambda\le\partial_\theta g\le\lambda+1$. Sample-to-sample gradient oscillation is at most four, uniformly in $\theta$, because the ridge term cancels. The population third derivative is bounded by a constant since $X$ is bounded. Thus \eqref{eq:primitive} holds with $\mu=.1$, $L_g=1.1$ and a finite $B_0,M_3$. The root curvature is $h=f''(\theta^\star)=.6107257225837609$. The innovation $g(\theta^\star;X,Y)$ is discrete and bounded, so the Gaussian-remainder bound in Theorem~\ref{thm:hybrid} applies.

\subsection{Exact random-curvature quadratic transfer}\label{app:randomcurvature}
A complementary state-dependent model admits an exact second-moment analysis. Let
\[
 g(x;A,\eps)=(h+A)(x-\theta^\star)+\eps,
 \qquad \E A=0,\quad \E A^2=\tau_A^2,
\]
with $|A|\le a_0<h$, $\eta(h+a_0)\le1$, and $A$ independent of centered $\eps$ with variance $\sigma^2>0$. Pairs $(A,\eps)$ are independent over clients and time. For a weighted vertex-transitive network, set $\chi=\eta^2\tau_A^2S_i(h)$.
\begin{proposition}[Random-curvature variance and coupling]\label{prop:randomcurvature}
The stationary chain has mean zero error and
\[
 \Var(e_i)=\frac{\eta^2\sigma^2S_i(h)}{1-\chi},\qquad
 \frac{\norm{e_i-z_i}_2}{\eta\sigma\sqrt{S_i(h)}}
       =\sqrt{\frac{\chi}{1-\chi}},
\]
where $z$ is driven by the same additive innovations $\eps$. In particular, $\chi<1$, and if $\eta\sqrt{S_i(h)}\to0$ this state-dependent correction vanishes on the local scale.
\end{proposition}
\begin{proof}
Sample maps contract by at most $1-\eta(h-a_0)$, giving a unique finite-second-moment stationary law for each finite graph. The stationary mean solves $m=(1-\eta h)Wm$, hence is zero. Write the update using the deterministic mean-curvature filter and effective noise $\nu_{j,t}=A_{j,t}e_{j,t-1}+\eps_{j,t}$. This is a martingale difference with zero cross-client covariance and common unconditional variance $\sigma^2+\tau_A^2V$, where $V=\E e_i^2$ is common to all clients by weighted vertex transitivity. Orthogonality of filtered effective innovations yields $V=\eta^2S_i(h)(\sigma^2+\tau_A^2V)$. The bound
\[
 \chi\le\frac{\eta\tau_A^2}{h(2-\eta h)}
 \le\frac{a_0^2}{h(h+2a_0)}<1
\]
justifies solving for $V$. The same-innovation difference solves a deterministic-filter recursion driven by $A_{j,t}e_{j,t-1}$, whose common variance is $\tau_A^2V$. Its variance is therefore $\eta^2\tau_A^2S_i(h)V=\chi V$, which proves the second formula. At a pilot $x_p$, centered residuals have innovation law $\eps+A(x_p-\theta^\star)$; their $\Wass_2$ discrepancy from $F_\eps$ is at most $\tau_A|x_p-\theta^\star|$. The displacement therefore contributes an additional calibration term.
\end{proof}

\section{Fixed-contrast inference for vector objectives}\label{app:vector}
Parameter dimension $p$ is distinct from graph dimension $d$. Throughout this appendix, $p$ is fixed and $u\ne0$ is specified before observing the data. Innovations are independent across clients and times; coordinates within an innovation may be dependent. All matrix norms are operator norms and deterministic vector norms are Euclidean. For a random vector $X$, write $\|X\|_{L^r(\ell_2)}=(\E\|X\|_2^r)^{1/r}$. The notation $A^{\,T}$ denotes a power, whereas $A^\top$ denotes a transpose.

\subsection{Directional filters and feasible calibration}\label{app:vectorfilter}
Let $H\succ0$ be symmetric, $A=I-\eta H$ with $\|A\|<1$, and let the common centered innovation law $F_\eps$ have covariance $\Sigma\succ0$. The homogeneous quadratic recursion is
\[
 e_{i,t+1}=\sum_jW_{ij}\{Ae_{j,t}-\eta\eps_{j,t+1}\}.
\]
For deterministic initial errors, $T$ updates and $q$ communication rounds give
\begin{equation}\label{eq:vectorfilter}
 \frac{u^\top(W^qe_T)_i}{\eta}
 =\frac{\sum_j(W^{T+q})_{ij}u^\top A^{\,T}e_{j,0}}{\eta}
 -\sum_{k<T,j}c_{kj}^\top\eps_{j,T-k},
 \qquad c_{kj}=(W^{k+q+1})_{ij}A^ku.
\end{equation}
This follows by iterating the recursion, since client mixing and the common parameter update act on different tensor factors. Define
\[
 s_u^2=\sum c_{kj}^\top\Sigma c_{kj},\quad
 s_{\calR,u}^2=\sum_{\calR}c_{kj}^\top\Sigma c_{kj},\quad
 v_u=s_{\calR,u}^2/s_u^2>0.
\]
The actual parameter standard deviation is $\eta s_u$. At stationarity $T=\infty$ and the initial term is absent. Geometric stability makes the coefficient sequence square summable. For a deterministic finite retained set $\calJ$, use a centered empirical joint law $\widehat F$ with second-moment matrix $\widehat\Sigma$ and the reference
\begin{equation}\label{eq:vectorreference}
 \widehat S_u=-\sum_{\calJ}\widehat c_{kj}^\top\eps^*_{kj}
 +\left(\sum_{\calR}\widehat c_{kj}^\top\widehat\Sigma\widehat c_{kj}\right)^{1/2}Z,
 \quad \eps^*_{kj}\stackrel{\mathrm{iid}}\sim\widehat F,\quad Z\sim\N(0,1).
\end{equation}
Each retained index receives an independent complete vector. The innovation draws and $Z$ are mutually independent, conditionally on the fitted inputs. Quantiles yield an interval for $u^\top\theta^\star$ by replacing $\widehat\theta_i$ in \eqref{eq:ci} with $u^\top\widehat\theta_i$.

\begin{proposition}[Directional vector calibration]\label{prop:vector}
Suppose $\E\|\eps\|^3<\infty$. Write
\[
 \mathcal E_{\calR,u}=\frac{\sum_{\calR}\E|c_{kj}^\top\eps|^3}{s_{\calR,u}^3}.
\]
Assume that, with probability at least $1-\zeta$, the fitted residual variance is positive and
\begin{equation}\label{eq:vectornuisance}
 \Wass_2\{(\Sigma^{-1/2})_\#\widehat F,(\Sigma^{-1/2})_\#F_\eps\}\le \omega_F,
 \qquad
 \frac{\{\sum\|\Sigma^{1/2}(\widehat c-c)\|^2\}^{1/2}}{s_u}\le\delta_C,
 \quad \omega_F,\delta_C\le1.
\end{equation}
Let $u_0$ bound the absolute initial term in \eqref{eq:vectorfilter} divided by $s_u$. With conditional reference-CDF error $t_B$ except on an event of probability $\zeta_B$, the coverage, each one-sided error, and the repeated-sampling PIT discrepancy are bounded by
\begin{equation}\label{eq:vectorcoverage}
 C\left\{\mathcal E_{\calR,u}+\frac{\omega_F+\delta_C+u_0}{\sqrt{v_u}}+t_B+\zeta+\zeta_B\right\}.
\end{equation}
The nuisance estimates may share the endpoint trajectory. Here $Q_\#F$ is the image of a law under $Q$.
\end{proposition}
\begin{proof}
Normalize coefficients by $s_u$ and whiten the true innovations. Their projections are independent centered scalars, with total variance one. Berry--Esseen applied to the residual sum gives $C\mathcal E_{\calR,u}$; convolution with the retained sum does not increase Kolmogorov distance. One can alternatively use
\[
 \beta_{3,\mathrm{proj}}=\sup_{a\ne0}\frac{\E|a^\top\eps|^3}{(a^\top\Sigma a)^{3/2}},
 \qquad
 \mathcal E_{\calR,u}\le\beta_{3,\mathrm{proj}}
 \frac{\sum_{\calR}(c_{kj}^\top\Sigma c_{kj})^{3/2}}{s_{\calR,u}^3}.
\]
This constant is finite under the stated moment and positive-definiteness conditions, but need not be uniform in $p$.

On the event \eqref{eq:vectornuisance}, optimally couple centered whitened innovations and take independent copies for each index. If $b=\Sigma^{1/2}c/s_u$ and $\widehat b=\Sigma^{1/2}\widehat c/s_u$, a projected difference can be written as $b^\top(\widehat\xi-\xi)+(\widehat b-b)^\top\widehat\xi$. Its $L^2$ norm is at most $\|b\|\omega_F+(1+\omega_F)\|\widehat b-b\|$. The differences are centered and independent across indices, so summing their variances and applying Minkowski in the coefficient sequence bounds the coupled retained sum by $C(\omega_F+\delta_C)$. For the residual standard deviations, realize both as $L^2$ norms of residual innovation sums with their respective laws. The same coupling and the reverse triangle inequality bound their difference by $C(\omega_F+\delta_C)$. This explains why a separate covariance-error term is unnecessary when $\widehat\Sigma$ comes from that same law.

The Gaussian-convolution comparison from Appendix~\ref{app:hybrid}, with oracle Gaussian scale $\sqrt{v_u}$, gives the nuisance term in \eqref{eq:vectorcoverage}. Shifting by the bounded initial term adds $Cu_0/\sqrt{v_u}$. The fixed-reference quantile sandwich and reference-computation event give coverage and one-sided errors; the PIT conclusion follows as in Appendix~\ref{app:calibrationcurve}. For countably many coefficients, first couple finite sums, then use their $L^2$ convergence and the continuous oracle Gaussian convolution. No independence of fitted inputs and endpoint is needed in the sandwich.
\end{proof}

For a smooth retained scalar projection $A_{\ell_0}=-c_{\ell_0}^\top\eps/s_u$, the density and second-derivative bounds of Proposition~\ref{prop:anchor} give its general $2/3$ nuisance branch, with the residual third moments replaced by the directional ones above. The stronger linear switch additionally requires $\widehat c_{\ell_0}=c_{\ell_0}\ne0$ and
\[
 \distK\{(-c_{\ell_0}^\top/s_u)_\#\widehat F,(-c_{\ell_0}^\top/s_u)_\#F_\eps\}\le\kappa
\]
on the same event. Replace the fitted anchor law first; its cost is $\kappa$. The common true scalar anchor then smooths the remaining centered coupling. Density assumptions alone do not imply this projected-CDF event.

Small maximal normalized group influence is sufficient for directional Gaussian approximation under the corresponding moment or Lindeberg condition. Non-Gaussianity of the vector law alone does not give necessity: for independent $G\sim\N(0,1)$ and Rademacher $R$, the vector $(G,R)$ is non-Gaussian, but $H=hI,u=(1,0)^\top$ produce an exactly Gaussian directional filter. The scalar equivalence in Lemma~\ref{thm:boundary} retains its stated scope.

\subsection{Curvature perturbations without commutation}\label{app:matrixperturbation}
\begin{proposition}[Matrix curvature sensitivity]\label{prop:matrixcurvature}
Suppose $h_-I\preceq H,\widehat H\preceq h_+I$, with symmetric Hessians, $h_->0$ and $\eta h_+\le1$. For the same $W,T,q,u$, the relative coefficient error in \eqref{eq:vectornuisance} satisfies
\begin{equation}\label{eq:matrixcurvature}
 \delta_C\le\sqrt{\kappa(\Sigma)}\,
 \frac{\|\widehat H-H\|}{h_-},\qquad
 \kappa(\Sigma)=\lambda_{\max}(\Sigma)/\lambda_{\min}(\Sigma).
\end{equation}
The conclusion holds for finite and stationary filters, without commutation of $H,\widehat H$, or $\Sigma$.
\end{proposition}
\begin{proof}
Set $E=\widehat H-H$. For a network eigenvalue $\lambda\in[-1,1]$, put $x_k=\lambda^kA^ku$, $y_k=\lambda^k\widehat A^ku$ and $d_k=y_k-x_k$. Then $d_0=0$ and
\[
 d_{k+1}=\lambda\widehat A d_k-\eta\lambda E x_k,
 \qquad
 \|d_k\|\le\eta|\lambda|\|E\|\sum_{r<k}
 [|\lambda|(1-\eta h_-)]^{k-1-r}\|x_r\|.
\]
Young's $\ell^1$--$\ell^2$ convolution inequality gives
\[
 \left(\sum_{k<T}\|d_k\|^2\right)^{1/2}
 \le\frac{\eta|\lambda|\|E\|}{1-|\lambda|(1-\eta h_-)}
 \left(\sum_{k<T}\|x_k\|^2\right)^{1/2}
 \le\frac{\|E\|}{h_-}\left(\sum_{k<T}\|x_k\|^2\right)^{1/2}.
\]
For a finite window, extend the input by zero and restrict the output; causality makes the displayed finite terms unchanged. For the infinite window the same inequality applies to square-summable sequences. Diagonalize only $W$: if its eigenvectors are $v_\ell$, then
\[
 \sum_j(W^{k+q+1})_{ij}^2
 =\sum_\ell v_\ell(i)^2\lambda_\ell^{2q+2}\lambda_\ell^{2k}.
\]
Multiply each squared modal inequality by the nonnegative weight $v_\ell(i)^2\lambda_\ell^{2q+2}$ and sum. This bounds $\sum\|\widehat c-c\|^2$ by $\|E\|^2h_-^{-2}\sum\|c\|^2$. Bound the covariance-weighted numerator above by $\lambda_{\max}(\Sigma)$ and its denominator below by $\lambda_{\min}(\Sigma)$ times these respective energies. This proves \eqref{eq:matrixcurvature}.
\end{proof}
There is no inverse network-gap or inverse-stepsize multiplier. The positive-update bracket is substantive: with $W=1,H=1,\widehat H=1.5,\eta=1.3,\Sigma=1$, both updates are stable, but the relative stationary coefficient error is about $2.791$, exceeding $|\widehat H-H|/h_-=.5$. Thus general spectral stability does not suffice for \eqref{eq:matrixcurvature}.

If $H_p=\nabla^2f(x_p)$ lies in the same bracket and a raw symmetric estimator obeys $\|\widetilde H-H_p\|\le r_H$, spectral clipping gives
\begin{equation}\label{eq:matrixpilot}
 \|\widehat H-H\|\le2r_H+M_3r_p,
 \quad \|x_p-\theta^\star\|\le r_p,
 \quad \|\nabla^2 f(x)-\nabla^2 f(y)\|\le M_3\|x-y\|.
\end{equation}
Indeed each out-of-bracket eigenvalue moves by at most $r_H$ by the extremal eigenvalue bounds; hence $\|\widehat H-\widetilde H\|\le r_H$. The triangle inequality then gives $2r_H$ relative to $H_p$. If $r_H$ already bounds the projected estimator, use $r_H+M_3r_p$ instead.

\subsection{Nonlinear vector moments from samplewise contraction}\label{app:vectornonlinear}
Let $g:\R^p\times\mathcal Z\to\R^p$ be a common stochastic gradient, with $\E g(x,\mathsf Z)=\nabla f(x)$ and $\nabla f(\theta^\star)=0$. Assume independent inputs across clients and times, and, for every $x,z$,
\begin{equation}\label{eq:vectorassumptions}
 \begin{gathered}
 D_xg(x,z)=D_xg(x,z)^\top,\quad
 0<\mu\le L_g,\quad \mu I\preceq D_xg(x,z)\preceq L_gI,\quad 0<\eta L_g\le\tfrac12,\\
 \|\nabla^2f(x)-\nabla^2f(y)\|\le M_3\|x-y\|,\qquad
 \|g(\theta^\star,\mathsf Z)\|\le B_* \ \text{a.s.}
 \end{gathered}
\end{equation}
Together with a known matrix $W$ satisfying Assumption~\ref{ass:filter}(i), these are exactly Assumption~\ref{ass:contraction} and the stepsize condition of Theorem~\ref{thm:vectortransfer}. For $q=0$, couple the recursions in \eqref{eq:nonlinearrecursions} using the same inputs. Set $H=\nabla^2f(\theta^\star)$, $A=I-\eta H$, $\eps=g(\theta^\star,\mathsf Z)$, $\Sigma=\Cov(\eps)\succ0$, and $r_0=1-\eta\mu$. Define
\[
 \mathcal S_i(a)=\sum_{k\ge0}(1-\eta a)^{2k}(W^{2k+2})_{ii},\qquad
 \mathcal S_i=\mathcal S_i(\mu),\quad
 \mathcal S_{\max}=\max_j\mathcal S_j,\quad \chi_i=\mathcal S_{\max}/\mathcal S_i.
\]
Each $\mathcal S_i$ is positive and at most $(1-r_0^2)^{-1}$. Weighted vertex transitivity is a sufficient condition for $\chi_i=1$; the result below keeps the energy ratio explicit on general graphs.
\begin{proposition}[Vector stationary moments and coupling]\label{prop:vectormoments}
Under \eqref{eq:vectorassumptions}, the nonlinear chain has a unique stationary law. Couple its error $e_i$ with the stationary linear error $z_i$ using the same inputs $\eps_{j,t}=g(\theta^\star,\mathsf Z_{j,t})$. For every $r\ge2$,
\begin{equation}\label{eq:vectormoments}
 \begin{aligned}
 \|e_i\|_{L^r(\ell_2)}&\le C\sqrt r\,\eta\sqrt{\mathcal S_i}+C\eta^2\mathcal S_{\max},\\
 \max_j\|\E e_j\|&\le C\eta^2\mathcal S_{\max},\qquad
 \|e_i-z_i\|_{L^r(\ell_2)}\le Cr\eta^2\mathcal S_{\max}.
 \end{aligned}
\end{equation}
The constants depend on $\mu,L_g,M_3,B_*$, not on the graph, its size, or $r$. For unit $u$ and the actual stationary directional scale $\eta s_u$, with $s_u^2=\Var(u^\top z_i)/\eta^2$,
\begin{equation}\label{eq:vectornormalized}
 \left\|\frac{u^\top(e_i-z_i)}{\eta s_u}\right\|_r
 \le Cr\frac{\eta\mathcal S_{\max}}{\sqrt{\mathcal S_i}}
 =Cr\chi_i\eta\sqrt{\mathcal S_i}.
\end{equation}
Here the constant may also depend on $\lambda_{\min}(\Sigma)$. Primitive constants may depend on the fixed dimension $p$.
\end{proposition}
\begin{proof}
\emph{Contraction and construction.}
The integral Jacobian of $x\mapsto x-\eta g(x,z)$ is symmetric with spectrum in $[1-\eta L_g,1-\eta\mu]$. With $r_0=1-\eta\mu$, two chains driven by the same inputs satisfy the client-wise norm inequality
\[
 \|e_i^{\rm new}-e_i^{\prime\rm new}\|
 \le r_0\sum_jW_{ij}\|e_j-e_j'\|.
\]
Also $\max_i\|e_{i,t+1}\|\le r_0\max_i\|e_{i,t}\|+\eta B_*$. Chains started at the root in the distant past are uniformly Cauchy under the same two-sided inputs, giving a stationary solution in the product of balls of radius $B_*/\mu$. Contraction gives uniqueness and convergence from every deterministic initial state. On this invariant set,
\[
 \|g(x,z)-g(x,z')\|\le2B_*+2L_g\|x-\theta^\star\|
 \le2\widetilde B,\qquad \widetilde B=B_*(1+L_g/\mu).
\]

\emph{Variance and centered moments.}
Replacing input $(j,-k)$ changes the client-$i$ stationary vector by at most $2\eta\widetilde B r_0^k(W^{k+1})_{ij}$. Efron--Stein applied coordinatewise and summed gives
\begin{equation}\label{eq:vectorvariance}
 V_i:=\E\|e_i-\E e_i\|^2\le2\eta^2\widetilde B^2\mathcal S_i(\mu).
\end{equation}
The scalar function $\|e_i-\E e_i\|$ has the same bounded differences. Its mean is at most $\sqrt{V_i}$, and
\[
 \Pp\{\|e_i-\E e_i\|>\sqrt{V_i}+x\}
 \le\exp\{-x^2/(2\widetilde B^2\eta^2\mathcal S_i)\}.
\]
Integrating this upper tail gives $\|e_i-\E e_i\|_r\le C\sqrt r\,\eta\sqrt{\mathcal S_i}$. For rigor, establish the bounds for finite past constructions; uniform contraction and bounded support pass both the moments and inequalities to the stationary limit. No coordinatewise union bound is used.

\emph{The network mean.}
Let $\bar e_i=\E e_i$ and
\[
 R_i=\E\nabla f(\theta^\star+e_i)-\nabla f(\theta^\star+\bar e_i).
\]
Taylor expansion at $\theta^\star+\bar e_i$ cancels the centered linear term, giving $\|R_i\|\le M_3 V_i/2$. Define the deterministic population update
\[
 (\mathcal T x)_i=\sum_jW_{ij}\{x_j-\eta\nabla f(\theta^\star+x_j)\}.
\]
It fixes zero and contracts the block maximum norm by $r_0$. Stationarity gives the stacked mean equation
\[
 \bar e=\mathcal T(\bar e)-\eta(W\otimes I_p)R.
\]
Consequently
\[
 \max_i\|\bar e_i\|\le r_0\max_i\|\bar e_i\|+\eta\max_i\|R_i\|,
 \qquad
 \max_i\|\bar e_i\|\le\frac{M_3}{2\mu}\max_i V_i
 \le C\eta^2\mathcal S_{\max}.
\]
Combining this with the centered moments proves the state bound in \eqref{eq:vectormoments}. Since $\mathcal S_{\max}\le(1-r_0^2)^{-1}$, the quantity $\eta\sqrt{\mathcal S_{\max}}$ is uniformly bounded by primitive parameters. In particular,
$\max_j\|e_j\|_r\le C\sqrt r\,\eta\sqrt{\mathcal S_{\max}}$, which controls the nonlinear remainder below.

\emph{Population drift and martingale noise.}
Write
\[
 r_f(x)=\nabla f(\theta^\star+x)-Hx,\quad
 \nu(x,\mathsf Z)=g(\theta^\star+x,\mathsf Z)-g(\theta^\star,\mathsf Z)-\nabla f(\theta^\star+x).
\]
Then $\|r_f(x)\|\le M_3\|x\|^2/2$, $\E_{\mathsf Z}\nu(x,\mathsf Z)=0$, and $\|\nu(x,\mathsf Z)\|\le2L_g\|x\|$. Thus $\delta=e-z$ obeys
\[
 \delta_{i,t+1}=\sum_jW_{ij}\{A\delta_{j,t}-\eta r_f(e_{j,t})-\eta\nu(e_{j,t},\mathsf Z_{j,t+1})\}.
\]
Iterate with the deterministic matrix $A$, without interchanging sample Hessians. Minkowski and stochastic row sums give the drift bound
\[
 \left\|\eta\sum_{k,j}(W^{k+1})_{ij}A^kr_f(e_{j,-k-1})\right\|_r
 \le\frac{M_3}{2\mu}\max_j\|e_j\|_{2r}^2\le Cr\eta^2\mathcal S_{\max}.
\]
For $r=2$, conditional cross-client and chronological orthogonality give the noise-filter second moment at most
\[
 4L_g^2\eta^2\mathcal S_i(\mu)\max_j\E\|e_j\|^2\le C\eta^4\mathcal S_i\mathcal S_{\max}.
\]
For $r\ge2$, the Hilbert martingale inequality
\begin{equation}\label{eq:hilbertmartingale}
 \left\|\sum_\ell d_\ell\right\|_r^2\le(r-1)\sum_\ell\|d_\ell\|_r^2
\end{equation}
applies by revealing inputs chronologically and by client within each time block, with pre-window inputs in the initial sigma-field. Each transformed $\nu$ is a martingale difference. Its $L^r$ norm is bounded by $2\eta L_g r_0^k(W^{k+1})_{ij}\|e_j\|_r$. Squaring, summing, and using the first moment bound gives noise-filter norm at most $Cr\eta^2\sqrt{\mathcal S_i\mathcal S_{\max}}\le Cr\eta^2\mathcal S_{\max}$. Finite-window tails tend to zero in $L^r$ by these same geometric energy bounds; the recursion remainder tends to zero by contraction. This proves the coupling part.

\emph{Directional normalization.}
Since the eigenvalues of $H$ lie in $[\mu,L_g]$ and $A$ is positive,
\[
 s_u^2\ge\lambda_{\min}(\Sigma)\|u\|^2\mathcal S_i(L_g).
\]
For each squared network eigenvalue, the ratio of resolvents is maximized at one, and
\[
 \frac{1-(1-\eta L_g)^2}{1-(1-\eta\mu)^2}
 =\frac{L_g}{\mu}\frac{2-\eta L_g}{2-\eta\mu}\le L_g/\mu.
\]
Consequently $\mathcal S_i(\mu)\le(L_g/\mu)\mathcal S_i(L_g)$. Divide the coupling bound by $\eta s_u$ to obtain \eqref{eq:vectornormalized}.
\end{proof}

For completeness, \eqref{eq:hilbertmartingale} follows from the $2$-smoothness argument behind standard martingale moment tools \citep{pinelis1994}. Put $F(t)=(\E\|X+tY\|^r)^{2/r}$. Differentiation and $\langle X+tY,Y\rangle^2\le\|X+tY\|^2\|Y\|^2$ give $F''(t)\le2(r-1)\|Y\|_r^2$ by H\"older; the term multiplied by $2/r-1$ is nonpositive. Taylor's formula therefore implies
\[
 \|X+Y\|_r^2\le\|X\|_r^2+
 2\|X\|_r^{2-r}\E[\|X\|^{r-2}\langle X,Y\rangle]+(r-1)\|Y\|_r^2.
\]
Regularize the norm at zero and pass to the limit, interpreting the middle term as zero when $X=0$ almost surely. For successive martingale differences, conditioning makes that term zero. Iteration proves \eqref{eq:hilbertmartingale}.

\subsection{Comparable local energies and graph perturbations}\label{app:energycomparison}
The local energy ratio in Theorem~\ref{thm:vectortransfer} can remain bounded even when the graph lacks transitivity and its spectral gap tends to zero. A comparison of graph Laplacians supplies a sufficient condition.

\begin{proposition}[Laplacian comparison]\label{prop:energycomparison}
Let $W_0,W$ be symmetric, entrywise nonnegative stochastic matrices with diagonal entries at least $1/2$, and let $W_0$ be weighted vertex-transitive. Set $\mathcal L_0=I-W_0$, $\mathcal L=I-W$ and suppose
\begin{equation}\label{eq:laplaciancomparison}
 c_L\mathcal L_0\preceq\mathcal L\preceq C_L\mathcal L_0,
 \qquad 0<c_L\le C_L<\infty.
\end{equation}
For $r_0=1-\eta\mu\in[0,1)$, define $\mathcal S_i(W)=\sum_{k\ge0}r_0^{2k}(W^{2k+2})_{ii}$ and $w_0=\min_i W_{ii}$. Then
\begin{equation}\label{eq:energycomparison}
 \max_j\frac{\mathcal S_j(W)}{\mathcal S_i(W)}
 \le\frac{2\max(1,C_L)}{w_0^2\min(1,c_L)}.
\end{equation}
The bound is uniform in graph size and stepsize when the comparison constants are uniform.
\end{proposition}
\begin{proof}
Put $G_W=(I-r_0W)^{-1}$ and $B_W=(I-r_0^2W^2)^{-1}$. Nonnegativity gives
$(W^{k+1})_{ij}\ge W_{ii}(W^k)_{ij}\ge w_0(W^k)_{ij}$.
For the $i$th coordinate vector $v_i\in\R^m$, symmetry and $\|W\|\le1$ also give $\|W^{k+1}v_i\|_2\le\|W^kv_i\|_2$. Squaring and summing imply
\[
 w_0^2(B_W)_{ii}\le\mathcal S_i(W)\le(B_W)_{ii}.
\]
Half-laziness makes $W$ positive semidefinite: $2W-I$ is symmetric stochastic with spectrum in $[-1,1]$. For $0\le\lambda\le1$,
\[
 \frac{1}{2(1-r_0\lambda)}\le\frac{1}{1-r_0^2\lambda^2}
 \le\frac{1}{1-r_0\lambda}.
\]
Functional calculus therefore yields
\begin{equation}\label{eq:energyresolvent}
 \frac{w_0^2}{2}(G_W)_{ii}\le\mathcal S_i(W)\le(G_W)_{ii}.
\end{equation}
Write $c_0=\min(1,c_L)$ and $C_0=\max(1,C_L)$. Since $I-r_0W=(1-r_0)I+r_0\mathcal L$, the comparison assumption gives
\[
 c_0(I-r_0W_0)\preceq I-r_0W\preceq C_0(I-r_0W_0).
\]
All three matrices are positive definite, including when the Laplacians are singular. Inverting reverses the order:
\[
 C_0^{-1}G_{W_0}\preceq G_W\preceq c_0^{-1}G_{W_0}.
\]
Transitivity makes the diagonal of $G_{W_0}$ constant. Combining its upper and lower bounds with \eqref{eq:energyresolvent} proves \eqref{eq:energycomparison}.
\end{proof}

\paragraph{Unequal edge weights and missing links.}
For the lazy torus in \eqref{eq:torus}, each undirected edge has weight $1/(4d)$ and the self-weight is $1/2$. Multiply the edge weights by arbitrary fixed factors in $[\kappa,1]$, with $\kappa>0$, and add the removed weight to the corresponding diagonals. The Dirichlet forms satisfy $\kappa\mathcal L_0\preceq\mathcal L\preceq\mathcal L_0$, so \eqref{eq:energycomparison} gives $\chi_i\le8/\kappa$.

Alternatively, delete one undirected edge on a torus with $d\ge2$ and side length at least three, again adding its weight to the two endpoint diagonals. A surviving three-edge path around an adjacent square joins its endpoints. Cauchy--Schwarz gives
\[
 (x_u-x_v)^2\le3\sum_{\{a,b\}\text{ on the path}}(x_a-x_b)^2.
\]
All original edge weights are equal. The removed edge's energy is thus at most three times the retained energy, and
\begin{equation}\label{eq:missingedge}
 \tfrac14\mathcal L_0\preceq\mathcal L\preceq\mathcal L_0,
 \qquad \chi_i\le32.
\end{equation}
For several deletions, suppose every removed edge has a retained detour of length at most $\ell$, and each retained edge belongs to at most $C_{\rm path}$ detours. Summing the same inequality gives
$\mathcal L_0\preceq(1+\ell C_{\rm path})\mathcal L$ and $\chi_i\le8(1+\ell C_{\rm path})$.
These constructions preserve symmetry, stochasticity and half-laziness. Uniform conclusions require uniform comparison constants, or uniformly bounded detour length and congestion.

\paragraph{Non-Gaussian persistence.}
The following corollary is used in Section~\ref{sec:nonlinear}.
\begin{corollary}[Persistence under graph perturbations]\label{cor:graphpersistence}
Consider networks with $W_{ii}\ge1/2$ and a lazy torus matrix $W_0$ on the same vertices, of fixed dimension $d\ge3$. Suppose $c_L(I-W_0)\preceq I-W\preceq C_L(I-W_0)$, where $\preceq$ is the positive-semidefinite order and $0<c_L\le C_L<\infty$ do not depend on network size or stepsize. If $m\eta\ge c>0$, then $\chi_i=O(1)$ and $\mathcal S_{\max}=O(1)$. Under the scalar assumptions of Theorem~\ref{thm:vectortransfer}, with fixed model bounds, a fixed non-Gaussian root noise and $\eta\to0$, the stationary local errors cannot converge to $\N(0,1)$ after centering and scaling by their own standard deviation.
\end{corollary}
This includes a torus with one edge removed and its weight added to the endpoint diagonals. The exact limit on the unmodified torus is given by Theorem~\ref{thm:transientlaw}.
For $d\ge3$, the torus return estimate in Appendix~\ref{app:graphs} gives
$\mathcal S_i(W_0)\le C\{1+(m\eta)^{-1}\}$. If $m\eta$ is bounded below, \eqref{eq:energyresolvent} applied to $W_0$ bounds its resolvent diagonal; Laplacian comparison then gives $\mathcal S_{\max}(W)=O(1)$. Together with $\mathcal S_i(W)\ge W_{ii}^2\ge1/4$, this proves the energy assertions in Corollary~\ref{cor:graphpersistence}.

For its scalar conclusion, let $h=f''(\theta^\star)\in[\mu,L_g]$ and let $S_{2,i}(h)$ denote the linear energy. Along the stated sequence,
\[
 \tfrac14\le W_{ii}^2\le S_{2,i}(h)\le\mathcal S_{\max}=O(1),
 \qquad
 \Lambda_i\ge W_{ii}/\sqrt{S_{2,i}(h)}\ge c>0.
\]
The fixed non-Gaussian root law and Lemma~\ref{thm:boundary} exclude a standard Gaussian limit for the centered linear error $z_i/(\eta\sigma\sqrt{S_{2,i}(h)})$. Proposition~\ref{prop:vectormoments} gives $\|e_i-z_i\|_2=O(\eta^2)$, so its normalized difference tends to zero in $L^2$. Slutsky's theorem excludes the same Gaussian convergence for the nonlinear error on the linear scale. Finally,
\[
 |\E e_i|=O(\eta^2),\qquad
 |\sqrt{\Var(e_i)}-\sqrt{\Var(z_i)}|
 \le\|(e_i-z_i)-\E(e_i-z_i)\|_2\le\|e_i-z_i\|_2.
\]
Since the linear standard deviation is bounded above and below by positive constants times $\eta$, own-mean centering and own-standard-deviation scaling are asymptotically equivalent. This proves Corollary~\ref{cor:graphpersistence}. Primitive model bounds and the root innovation law are fixed along the sequence. This qualitative conclusion needs no density assumption; identifying a particular limiting law remains the content of Theorem~\ref{thm:transientlaw} on the unmodified torus.

\subsection{Finite-time inference and observable nuisance estimates}\label{app:vectorfinite}
Set $Q=\max_j\|e_{j,0}\|+B_*/\mu$, and use the finite linear variance $s_{T,u}^2$ from \eqref{eq:vectorfilter} with $q=0$. The client index is suppressed. For unit $u$, define
\begin{equation}\label{eq:vectortransient}
 J_{T,u}=\frac{\eta\mathcal S_{\max}}{s_{T,u}},\qquad
 I_{T,u}=\frac{(1-\eta\mu)^TQ}{\eta s_{T,u}},\qquad
 \psi(x)=\min\{1,x\log(e+1/x)\},\quad\psi(0)=0.
\end{equation}
\begin{corollary}[Finite-time feasible vector inference]\label{cor:vectorfinite}
Under \eqref{eq:vectorassumptions}, initialize the nonlinear and linear chains at the same deterministic state and fix a unit direction. Suppose the finite fitted reference satisfies Proposition~\ref{prop:vector}'s nuisance, residual-variance and numerical-CDF events. If its fixed oracle reference on the unit-linear-standard-deviation scale has density bounded by $f_0$, the coverage, each one-sided error and PIT discrepancy of the nonlinear endpoint are at most
\begin{equation}\label{eq:vectorfinitecoverage}
 C\left\{\mathcal E_{\calR,u}+\frac{\omega_F+\delta_C+u_0}{\sqrt{v_u}}
 +t_B+\zeta+\zeta_B+\psi(f_0J_{T,u})+f_0I_{T,u}\right\}.
\end{equation}
The nonlinear contribution can instead be bounded by
$C(f_0J_{T,u})^{2/3}+2f_0I_{T,u}$ using only $L^2$ coupling.
\end{corollary}
\begin{proof}
Couple finite nonlinear and linear chains, initialized identically, with the stationary pair using the same two-sided inputs. Both stationary errors are bounded by $B_*/\mu$ in block maximum norm. Contraction bounds the difference of the finite and stationary coupling errors by $2r_0^TQ$. Proposition~\ref{prop:vectormoments} consequently gives, for the finite normalized difference $\Delta_T$,
\[
 \|\Delta_T\|_r\le CrJ_{T,u}+2I_{T,u}.
\]
More precisely the transient component is bounded pathwise, so it does not enter the Markov optimization. For $t\ge2$, apply Markov with $r=t$ to the stationary component to obtain
\[
 \Pp\{|\Delta_T|>eCtJ_{T,u}+2I_{T,u}\}\le e^{-t}.
\]
Comparing the nonlinear CDF to a fixed reference of density at most $f_0$ adds at most $f_0(eCtJ_{T,u}+2I_{T,u})+e^{-t}$ to the finite linear-reference error. For small $f_0J_{T,u}$ choose $t=\max\{2,\log(1/(f_0J_{T,u}))\}$; for larger values use the trivial CDF bound one. This proves the $\psi$ term. The $L^2$ alternative follows by minimizing $f_0\epsilon+CJ_{T,u}^2/\epsilon^2$, with the same separate deterministic shift.

The fixed-reference quantile sandwich then adds fitted-law and numerical-reference errors exactly as in Proposition~\ref{prop:vector}. The linear initial term remains $u_0$: it compares the finite linear chain with its centered reference, whereas $I_{T,u}$ controls transfer from stationary to finite nonlinear coupling. The two terms serve different comparisons. All event failure probabilities are combined by a union bound, with no assertion of their independence.
\end{proof}

Here $f_0\le(2\pi v_u)^{-1/2}$ is always available from the oracle Gaussian remainder; an applicable retained density may give a better bound. This anti-concentration factor enters the graph-uniform calibration bound. On the Laplacian-comparable $d\ge3$ torus families of Appendix~\ref{app:energycomparison}, with $m\eta$ bounded below, $\mathcal S_{\max}=O(1)$ and $\mathcal S_i\ge1/4$. The stationary nonlinear transfer component is then $O(\eta\log(1/\eta))$ when $f_0$ is bounded. The full calibration error still includes residual replacement and estimation.

\begin{corollary}[Fixed final gossip]\label{cor:nonlineargossip}
Under Proposition~\ref{prop:vectormoments}, apply a prescribed $q\ge0$ communication-only rounds to both chains. Put
\[
 \mathcal S_{i,q}(a)=\sum_{k\ge0}(1-\eta a)^{2k}(W^{2k+2q+2})_{ii},\qquad
 s_{u,q}^2=\Var\{u^\top(W^qz)_i\}/\eta^2.
\]
For stationary errors,
\[
 \|u^\top[W^q(e-z)]_i\|_r\le Cr\eta^2\mathcal S_{\max},\qquad
 \frac{\|u^\top[W^q(e-z)]_i\|_r}{\eta s_{u,q}}
 \le Cr\frac{\eta\mathcal S_{\max}}{\sqrt{\mathcal S_{i,q}(\mu)}}.
\]
For finite $T\ge1$, let $s_{T,u,q}^2$ be the variance in \eqref{eq:vectorfilter} divided by $\eta^2$. If $s_{T,u,q}>0$, Corollary~\ref{cor:vectorfinite} holds with the actual post-gossip filter, its $u_0,v_u,\mathcal E_{\calR,u},f_0$, and
\[
 J_{T,u,q}=\eta\mathcal S_{\max}/s_{T,u,q},\qquad
 I_{T,u,q}=r_0^TQ/(\eta s_{T,u,q}).
\]
\end{corollary}
\begin{proof}
The rows of $W^q$ are probability vectors. Minkowski's inequality therefore preserves the unnormalized coupling bound and the pathwise transient allowance $2r_0^TQ$. The normalization argument in Proposition~\ref{prop:vectormoments}, now with modal weights $\lambda^{2q+2}$, gives
$s_{u,q}^2\ge\lambda_{\min}(\Sigma)\mathcal S_{i,q}(L_g)\ge(\mu/L_g)\lambda_{\min}(\Sigma)\mathcal S_{i,q}(\mu)$.
The finite-time CDF comparison then applies with its own variance. The numerator $\mathcal S_{\max}$ describes the optimization chain before gossip; the denominator accounts for the variance reduction from the added communication.
\end{proof}

\paragraph{Fresh observations.}
Conditional on a pilot $x_p$, use fresh inputs independent of the endpoint trajectory and couple each gradient with $g(\theta^\star,\mathsf Z)$. The mean-value bound and empirical centering yield an ordinary $\Wass_2$ discrepancy at most $2L_g\|x_p-\theta^\star\|$ between the centered empirical gradient law and its centered latent-root empirical law. Thus, on a joint event with pilot radius $r_p$,
\begin{equation}\label{eq:vectorfresh}
 \omega_F\le \omega_{\mathrm{emp},p}+\frac{2L_gr_p}{\sqrt{\lambda_{\min}(\Sigma)}},\qquad
 \delta_C\le\sqrt{\kappa(\Sigma)}\frac{2r_H+M_3r_p}{h_-},
\end{equation}
where $\omega_{\mathrm{emp},p}$ measures the centered whitened latent-root empirical law, and $r_H$ is a raw-Hessian radius as in \eqref{eq:matrixpilot}. The fitted Gaussian covariance is computed from the same observed centered law. The same decomposition applies when the pilot is the endpoint.

For analytic sample Hessians at a fresh pilot, each entry is in $[-L_g,L_g]$. Conditional Hoeffding bounds and a union bound over $p^2$ entries give, with failure probability at most $\zeta_H$,
\[
 \|\widetilde H-\nabla^2 f(x_p)\|
 \le pL_g\sqrt{2\log(2p^2/\zeta_H)/M_H}.
\]
The observations are independent of the trajectory conditional on the pilot; Hessian and gradient estimates may use the same observations. For bounded root noise and fixed $p=3$, Theorem~1 of \citet{fournier2015} gives $\E\Wass_2^2=O(M^{-1/2})$ for the latent empirical law. Centering adds a sample-mean term of order $M^{-1/2}$ in $L^2$, so $\omega_{\mathrm{emp},3}=O_{\Pp}(M^{-1/4})$. A deterministic radius at a specified failure probability follows from that moment bound and Markov; the dimension remains fixed in this rate.

\paragraph{Stored observations.}
At deterministic record times, let $r_t=g(\theta_{i,t-1},\mathsf Z_{i,t})-H_{\rm nom}\theta_{i,t-1}$, with fixed $H_{\rm nom}$. Pair with $\eps_t=g(\theta^\star,\mathsf Z_{i,t})$. Empirical centering removes the common $-H_{\rm nom}\theta^\star$ shift and gives the ordinary Wasserstein contamination bound
\[
 2(L_g+\|H_{\rm nom}\|)
 \left(M^{-1}\sum_t\|e_{i,t-1}\|^2\right)^{1/2}.
\]
The latent $\eps_t$ are independent even though the recorded states are not. The concentration proof above, a coupling to stationarity and a union bound over the $M$ specified times give, outside probability $\zeta_p$,
\[
 \|e_{i,t-1}\|\le C\eta\sqrt{\mathcal S_i}\{1+\sqrt{\log(2M/\zeta_p)}\}+C\eta^2\mathcal S_{\max}+r_0^{t-1}Q
\]
at all record times. This supplies an explicit envelope for the contamination bound. Divide it by $\sqrt{\lambda_{\min}(\Sigma)}$ and add the latent empirical-law error. The Hessian estimation event remains separate. Coverage is marginal over trajectory and calibration randomness, not conditional on the observed trajectory.

\subsection{Models and the boundary of the extension}\label{app:vectormodels}
Bounded-feature ridge logistic gradients satisfy \eqref{eq:vectorassumptions}: the ridge supplies $\mu$, the feature bound supplies $L_g$ and $M_3$, and bounded features and binary responses bound root noise. Bounded-feature, bounded-response ridge least squares also satisfies these assumptions, although its sample-to-sample gradient oscillation need not be bounded over all of $\R^p$. The root-noise formulation and invariant ball permit that case.

Heterogeneous client objectives introduce a local bias. With common scalar curvature and client offsets $b_j$ summing to zero, the stationary bias is
\[
 \E e_\infty=-\eta[I-(1-\eta h)W]^{-1}Wb.
\]
On an even three-dimensional lazy torus, $b_x=b_0(-1)^{x_1}$ is a mean-zero mode with eigenvalue $2/3$. Its local bias divided by $\eta$ tends to $-2b_x$, which need not vanish relative to local fluctuations of order $\eta$. Centering innovations does not remove this target shift. Heterogeneous stationary bias is studied by \citet{versini2026}; the present extension concerns homogeneous objectives and predetermined scalar contrasts.

\section{Calibration curves and cancellation of tail errors}\label{app:calibrationcurve}
Let $X$ be the normalized statistic and $H$ the fixed continuous strictly increasing oracle CDF. Suppose $\|F_X-H\|_\infty\le\beta$ and $\|\widehat H-H\|_\infty\le\delta$ except on an event of probability $\zeta$. The computed Gaussian-convolution CDF is continuous and strictly increasing whenever its residual variance is positive. On the good event,
\[
 H(X)-\delta\le U:=\widehat H(X)\le H(X)+\delta.
\]
The CDF of $H(X)$ differs from the uniform CDF by at most $\beta$, including boundary points by clipping probabilities to $[0,1]$. Deterministic bracketing therefore gives
\begin{equation}\label{eq:pitproof}
 \sup_{0\le p\le1}|\Pp\{U\le p\}-p|\le\beta+\delta+\zeta.
\end{equation}
An undefined reference on a bad event may assign any value to $U$ and is charged to $\zeta$. Apply this inequality to the statistic and reference bounds in Theorem~\ref{thm:hybrid}, including the conditional Monte Carlo event, to prove Corollary~\ref{cor:calibrationcurve}. The same proof applies to the smooth and nonlinear versions. The comparison allows dependence between $X$ and $\widehat H$ and yields marginal repeated-sampling calibration.

For $p=\alpha/2$, lower and upper errors are $\Pp(U<p)$ and $\Pp(U>1-p)$. Their sum is exactly one minus closed-interval coverage. Writing their deviations from $p$ as $\Delta_L,\Delta_U$ gives coverage error $-(\Delta_L+\Delta_U)$. Opposite deviations may cancel. This explains why a central-coverage curve alone cannot diagnose distributional calibration. Possible atoms are handled by the left limit at the lower endpoint; their mass is at most twice the uniform-CDF discrepancy in \eqref{eq:pitproof}.

\section{Experimental design and evidence}\label{app:allstudies}
\subsection{Local shape under network averaging}\label{app:experiments}
\subsubsection{Full-network data generation and resource accounting}
For each cell, initialize every client at zero and execute the stated ATC recursion on the full network state. Every client and iteration receives an independent innovation. The common target is $\theta^\star=.35$ for quadratic losses and $\theta^\star=.5$ for logistic losses. The lazy graph is exactly \eqref{eq:torus}. Skew quadratic noise takes $3$ with probability $.1$ and $-1/3$ with probability $.9$; it has zero mean and unit variance. The Gaussian-smoothed skew control is $.8\eps_{\rm skew}+.6Z$ with independent $Z\sim\N(0,1)$, again of unit variance. Gaussian controls use $\N(0,1)$. Rare-shock noise takes $7$ with probability $.02$ and $-1/7$ with probability $.98$ (Appendix~\ref{app:rareshock}).

Let $N_{\rm comm}=\max\{\lceil16/(\eta\mu)\rceil+4,M+10\}$, where $M=400$, $\mu=h$ for quadratics and $\mu=.1$ for logistic loss. The ordinary endpoint uses $T=N_{\rm comm}$ gradient updates. The extra-gossip endpoint uses $T=N_{\rm comm}-4$ updates followed by four communication-only rounds. Both therefore use exactly $N_{\rm comm}$ communication rounds. Their total network gradient counts are $mN_{\rm comm}$ and $m(N_{\rm comm}-4)$. The paired trajectories share innovations up to the gossip branch. All finite filters use the corresponding true number of gradient updates and gossip rounds.

For quadratic fresh calibration, obtain $M=400$ new gradients at that endpoint and center them. Estimate curvature from 100 independent gradients at each of endpoint plus one and endpoint minus one; divide the difference of averages by two and truncate to $[.3,2]$. The two finite-difference samples use independent innovations. There are 600 extra gradient calls at the evaluated client. For logistic fresh calibration, collect 400 new $(X,Y)$ samples at the endpoint. Use the same observations for centered gradients and the analytic sample Hessian, incurring 400 gradient and 400 Hessian evaluations. The nuisance event includes the dependence of these estimates on the endpoint.

The reuse protocol records the last 400 gradients and preceding states. Quadratic reuse uses $h_0=1$ both for the filter and for residual detrending. Logistic reuse uses the known covariate distribution to evaluate $f''$ at its endpoint and detrends by that value. This model-specific control evaluates known population curvature at the endpoint using no extra observations. Recorded gradients are empirically centered after detrending. Their distinct information costs are listed in Appendix~\ref{app:resources}.

The retained set is fixed by ranking the finite filter at nominal curvature one for quadratics and $.6$ for logistic loss. The 128 largest coefficients are retained. For each endpoint, the selected coefficients are recomputed at its estimated or nominal curvature, and residual variance is computed from the exact spectral energy minus their squared sum. Each of the 1499 retained sums uses an independent empirical innovation at every selected coefficient. Hybrid-MC and Hybrid-RB use the same retained draws to isolate Gaussian-tail integration. Quantiles are computed without knowing the true target; targets enter only the evaluation of interval coverage and bias.

Rao--Blackwellized inversion uses 35 bisection steps on \eqref{eq:rb} inside a bracket of nine residual standard deviations beyond the extreme retained draws, which excludes probability at most $2\Phi(-9)$. Spectral, recursion and variance identities and positive residual variance were checked before the experiments; no interval was undefined.

\subsubsection{Replication counts and Monte Carlo uncertainty}
Each main cell uses 1600 independent trajectories with cell seeds $27270901+10^6j$, $j=0,\ldots,7$, in the order of Table~\ref{tab:complete}. Fresh calibration, fresh and reused resampling, and Gaussian-tail simulation add offsets $10^5$, $2\times10^5$, $3\times10^5$ and $4\times10^5$, and replication $r$ adds $7919r$. Settings were fixed before the trajectories were evaluated.

Monte Carlo standard errors are $\sqrt{\widehat p(1-\widehat p)/(N_{\rm rep}-1)}$, figures use Wilson 95\% intervals, and methods evaluated on the same trajectory are compared in pairs. ``Lower'' and ``upper'' refer to errors below or above the reference quantile, which correspond to the opposite interval endpoints; coverage intervals are closed and failures strict.

\begin{longtable}{llrrrrr}
\caption{Complete validation results, 1600 independent trajectories per cell. Cov., Low and Up are percentages; length is on the parameter scale. SE is the Monte Carlo standard error of coverage in percentage points.}\label{tab:complete}\\
\toprule Setting & Method & Cov. & Low & Up & Length & SE\\\midrule
\endfirsthead
\toprule Setting & Method & Cov. & Low & Up & Length & SE\\\midrule
\endhead
\bottomrule\endfoot
Quad. 1D, $\eta=0.05$ & Gaussian-fresh & 94.88 & 4.12 & 1.00 & 0.2347 & 0.55\\
skew & Hybrid-RB & 93.75 & 2.81 & 3.44 & 0.2323 & 0.61\\
 & Hybrid-MC & 93.62 & 2.94 & 3.44 & 0.2318 & 0.61\\
 & Hybrid-reuse & 96.44 & 2.06 & 1.50 & 0.2581 & 0.46\\
 & Gaussian +4 gossip & 94.69 & 3.75 & 1.56 & 0.1919 & 0.56\\
\addlinespace[5pt]
Quad. 2D, $\eta=0.05$ & Gaussian-fresh & 94.44 & 4.75 & 0.81 & 0.1685 & 0.57\\
skew & Hybrid-RB & 94.31 & 2.56 & 3.12 & 0.1686 & 0.58\\
 & Hybrid-MC & 94.19 & 2.69 & 3.12 & 0.1682 & 0.59\\
 & Hybrid-reuse & 95.69 & 2.25 & 2.06 & 0.1779 & 0.51\\
 & Gaussian +4 gossip & 95.19 & 3.56 & 1.25 & 0.1028 & 0.54\\
\addlinespace[5pt]
Quad. 3D, $\eta=0.05$ & Gaussian-fresh & 93.81 & 5.88 & 0.31 & 0.1447 & 0.60\\
skew & Hybrid-RB & 94.38 & 2.56 & 3.06 & 0.1452 & 0.58\\
 & Hybrid-MC & 94.12 & 2.62 & 3.25 & 0.1448 & 0.59\\
 & Hybrid-reuse & 96.19 & 2.19 & 1.62 & 0.1513 & 0.48\\
 & Gaussian +4 gossip & 93.94 & 3.31 & 2.75 & 0.0651 & 0.60\\
\addlinespace[5pt]
Quad. 3D, $\eta=0.05$ & Gaussian-fresh & 95.19 & 2.06 & 2.75 & 0.1450 & 0.54\\
gaussian & Hybrid-RB & 94.75 & 2.12 & 3.12 & 0.1448 & 0.56\\
 & Hybrid-MC & 94.88 & 2.19 & 2.94 & 0.1445 & 0.55\\
 & Hybrid-reuse & 95.38 & 2.19 & 2.44 & 0.1498 & 0.53\\
 & Gaussian +4 gossip & 94.31 & 2.69 & 3.00 & 0.0652 & 0.58\\
\addlinespace[5pt]
Quad. 3D, $\eta=0.05$ & Gaussian-fresh & 94.81 & 3.50 & 1.69 & 0.1448 & 0.55\\
smooth & Hybrid-RB & 94.88 & 1.94 & 3.19 & 0.1470 & 0.55\\
 & Hybrid-MC & 94.69 & 1.94 & 3.38 & 0.1466 & 0.56\\
 & Hybrid-reuse & 95.50 & 1.69 & 2.81 & 0.1518 & 0.52\\
 & Gaussian +4 gossip & 94.69 & 2.94 & 2.38 & 0.0653 & 0.56\\
\addlinespace[5pt]
Logistic 3D, $\eta=0.08$ & Gaussian-fresh & 95.69 & 3.50 & 0.81 & 0.1657 & 0.51\\
 & Hybrid-RB & 95.31 & 2.19 & 2.50 & 0.1633 & 0.53\\
 & Hybrid-MC & 95.12 & 2.12 & 2.75 & 0.1629 & 0.54\\
 & Hybrid-reuse & 95.44 & 1.94 & 2.62 & 0.1632 & 0.52\\
 & Gaussian +4 gossip & 95.44 & 2.69 & 1.88 & 0.0786 & 0.52\\
\addlinespace[5pt]
Logistic 2D, $\eta=0.08$ & Gaussian-fresh & 95.69 & 3.00 & 1.31 & 0.1978 & 0.51\\
 & Hybrid-RB & 95.12 & 2.25 & 2.62 & 0.1964 & 0.54\\
 & Hybrid-MC & 94.88 & 2.31 & 2.81 & 0.1960 & 0.55\\
 & Hybrid-reuse & 95.44 & 2.06 & 2.50 & 0.1966 & 0.52\\
 & Gaussian +4 gossip & 95.38 & 2.19 & 2.44 & 0.1265 & 0.53\\
\addlinespace[5pt]
Logistic 3D, $\eta=0.04$ & Gaussian-fresh & 94.81 & 4.12 & 1.06 & 0.0886 & 0.55\\
 & Hybrid-RB & 94.81 & 2.75 & 2.44 & 0.0878 & 0.55\\
 & Hybrid-MC & 94.75 & 2.94 & 2.31 & 0.0876 & 0.56\\
 & Hybrid-reuse & 94.75 & 2.50 & 2.75 & 0.0878 & 0.56\\
 & Gaussian +4 gossip & 95.06 & 2.62 & 2.31 & 0.0473 & 0.54\\
\addlinespace[5pt]
\end{longtable}

\subsubsection{Exact-moment Gaussian diagnostic}
For each quadratic cell we use the saved endpoint observations, the exact finite mean $-\theta^\star(1-\eta h)^T$ of the estimation error, and variance $\eta^2S_2(h)$ (all quadratic innovations have unit variance). This oracle diagnostic uses exact moments on the same mechanism-study endpoints. Table~\ref{tab:oracle} gives the one-sample empirical Kolmogorov discrepancy against the standard normal CDF. With $N_{\rm rep}=1600$ and five quadratic cells, DKW and a union bound give
\[
 \sup_x|F_{N_{\rm rep},c}(x)-F_c(x)|\le
 \sqrt{\log(2\cdot5/.05)/(2\cdot1600)}=.040691\ldots
\]
simultaneously for all cells with probability at least $.95$. Subtracting this bound from the observed discrepancy in the three-dimensional skew cell gives a population discrepancy of at least $.0536$. The bound describes Monte Carlo uncertainty under the specified model.
\begin{table}[ht]
\caption{Gaussian shape diagnostics using exact finite-horizon mean and variance. Coverage and tails describe the exact-moment Gaussian interval.}\label{tab:oracle}
\begin{center}
\begin{tabular}{llrrrr}\toprule
Graph & Innovations & KS & Coverage (\%) & Lower (\%) & Upper (\%)\\\midrule
1D & skew & 0.0394 & 95.62 & 3.62 & 0.75\\
2D & skew & 0.0780 & 94.94 & 4.56 & 0.50\\
3D & skew & 0.0943 & 94.50 & 5.31 & 0.19\\
3D & gaussian & 0.0167 & 95.12 & 2.19 & 2.69\\
3D & smooth & 0.0643 & 94.88 & 3.56 & 1.56\\
\bottomrule\end{tabular}

\end{center}
\end{table}

\subsubsection{Nonlinear coupling and reference-computation experiments}
The additional nonlinear diagnostic uses the same three-dimensional $6^3$ network and logistic distribution. For each of $\eta=.12,.08,.04,.02$, run 1600 paired trajectories for $\lceil14/(\eta\mu)\rceil$ updates, starting both nonlinear and linearized errors at zero. Every pair uses the same $(X,Y)$ innovations; the linearized chain is driven by the actual score at $\theta^\star$. The normalization uses the stationary linear-filter energy. At these run lengths its difference from the finite-horizon energy is bounded relatively by $10^{-72}$, so the distinction is below floating-point precision. Seeds are 8271901 through 8271904. The normalized difference measures the same-input coupling directly.
\begin{figure}[ht]
 \centering\includegraphics[width=.8\textwidth]{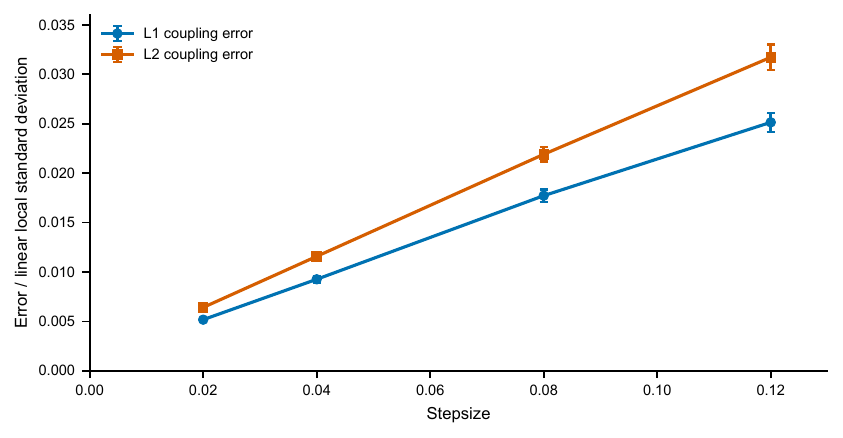}
 \caption{Same-innovation nonlinear--linear coupling for regularized logistic loss. Both $L^1$ and $L^2$ errors are normalized by the linear local standard deviation. Bars are 95\% trajectory-bootstrap intervals from 5000 resamples, with 1600 independent pairs per stepsize.}\label{fig:coupling}
\end{figure}
\begin{table}[ht]
\caption{Actual paired nonlinear--linear coupling errors. The last column is the graph-dependent factor in Theorem~\ref{thm:nonlinear}, not a fitted upper bound.}\label{tab:coupling}
\begin{center}
\begin{tabular}{rrrrr}\toprule
$\eta$ & Normalized $L^1$ & MC SE & Normalized $L^2$ & $\eta\sqrt{S_2}$\\\midrule
0.12 & 0.025139 & 0.000484 & 0.031720 & 0.088468\\
0.08 & 0.017718 & 0.000322 & 0.021902 & 0.061475\\
0.04 & 0.009257 & 0.000174 & 0.011576 & 0.032927\\
0.02 & 0.005172 & 0.000095 & 0.006409 & 0.017855\\
\bottomrule\end{tabular}

\end{center}
\end{table}
\begin{table}[ht]
\caption{Conditional CDF-estimator variance from 1000 reference repetitions with 199 retained draws each. Ratio is RB variance divided by MC variance.}\label{tab:rbvariance}
\begin{center}
\begin{tabular}{rrrrr}\toprule
Retained count & $x$ & RB variance & MC variance & Ratio\\\midrule
8 & -1.96 & 0.000121 & 0.000236 & 0.511\\
8 & 0.00 & 0.000619 & 0.00127 & 0.487\\
8 & 1.96 & 1.2e-07 & 1.97e-05 & 0.006\\
32 & -1.96 & 0.00016 & 0.000227 & 0.706\\
32 & 0.00 & 0.000811 & 0.00123 & 0.661\\
32 & 1.96 & 4.59e-07 & 1.05e-05 & 0.044\\
128 & -1.96 & 0.000187 & 0.000235 & 0.794\\
128 & 0.00 & 0.00101 & 0.00128 & 0.792\\
128 & 1.96 & 2.28e-06 & 9.11e-06 & 0.250\\
\bottomrule\end{tabular}

\end{center}
\end{table}

\begin{figure}[ht]
 \centering\includegraphics[width=.8\textwidth]{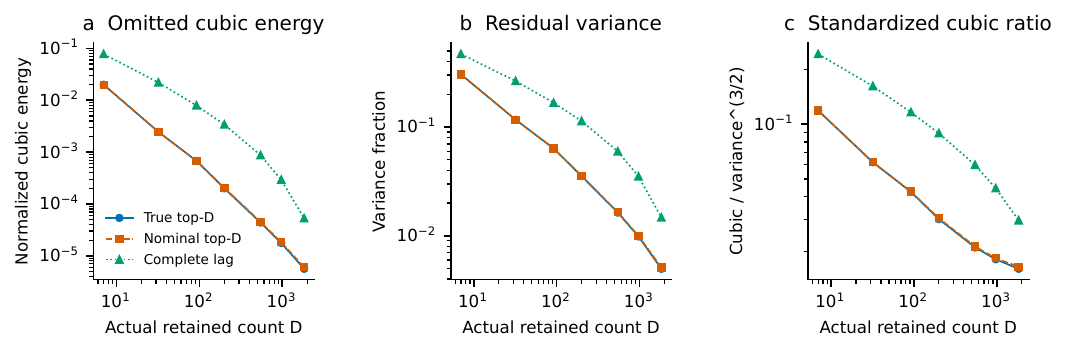}
 \caption{Retention geometry at matched actual counts. Left: omitted cubic energy normalized by total variance. Middle: residual variance fraction. Right: the residual standardized cubic ratio, without the innovation third-moment factor. The panels compare geometric quantities for complete windows and true or nominal magnitude ranking.}\label{fig:retention}
\end{figure}

The retention diagnostic uses the $6^3$ torus, $\eta=.05$, $h=1.4$, nominal curvature 1 and a finite horizon of 500. For each complete lag window $K=1,2,3,4,6,8,12$, we count its nonzero coefficients and compare true top-$D$, nominal top-$D$ and complete-lag retention at exactly that count. Figure~\ref{fig:retention} reports omitted cubic energy, residual variance and their standardized ratio. True magnitude ordering minimizes cubic energy by an exchange argument; neither nominal ranking nor minimizing the cubic numerator alone guarantees the smallest standardized ratio. This is a deterministic filter comparison. The matched lag-3 coverage experiments appear separately below.

For the reference-variance diagnostic, hold a single centered empirical innovation law of size 400 fixed. Repeat the reference calculation 1000 times with $B=199$, retained counts 8, 32 and 128, and evaluation points $-1.96,0,1.96$ on the standardized scale. Within each repetition, Hybrid-RB and Hybrid-MC share retained draws. Table~\ref{tab:rbvariance} reports the conditional variance of the estimated CDF. The theoretical Rao--Blackwell guarantee is conditional on the empirical law and applies to every CDF location; finite estimated variance ratios fluctuate.

\subsection{Scalar calibration and budget sensitivity}\label{app:revisionexperiments}
\subsubsection{Design and independent data}
The calibration study was specified after inspecting the mechanism study, but before examining outcomes from its independent seeds. It comprises skew quadratic ($N_{\rm rep}=4000$, seed 91027001), Gaussian quadratic ($N_{\rm rep}=1600$, 92027001), Gaussian-smoothed skew quadratic ($N_{\rm rep}=1600$, 93027001), logistic ($N_{\rm rep}=4000$, 94027001), and bounded smooth beta quadratic ($N_{\rm rep}=1600$, 95027001). All graphs are the lazy $6^3$ torus. The beta innovation is $(V-4/11)/\sqrt{28/1452}$ for $V\sim\operatorname{Beta}(4,7)$; it has mean zero, variance one, bounded support, and a globally $C^2$ density after zero extension. It represents the smooth bounded class in the rate corollary; the Gaussian-smoothed control separately examines a smooth unbounded law.

Quadratic endpoints use $T=410$, $h=1.4$, $\eta=.05$, target $.35$; logistic endpoints use $T=2004$, $\eta=.08$, target $.5$, and the same ridge/data law as the original study. The whole network starts at zero. Each replication stores its endpoint, last 400 trajectory gradients and states, a fresh stream of 1600 calibration gradients, and curvature inputs. All calibration budgets use the same fixed-horizon endpoints.

All $M$ values use prefixes of the same fresh stream. For quadratic loss, curvature is estimated using two independent 100-gradient samples, independent of that stream. For logistic loss, it uses the first 400 features to evaluate sample Hessians at the endpoint. Thus the innovation-law budget is $M$ gradient evaluations and the curvature budget remains 200 gradients (quadratic) or 400 Hessians (logistic). With $M<400$, the logistic curvature features beyond the first $M$ still incur Hessian evaluations. Oracle comparisons replace the innovation law at $\theta^\star$, the root curvature $f''(\theta^\star)$, or both by their true values. A true Gaussian innovation law is integrated analytically in its oracle reference.

The $D$ sweep is $0,8,32,128,512$ at $M=400$; the $M$ sweep is $100,400,1600$ at $D=128$. The remaining three true/estimated-curvature by true/empirical-law combinations are evaluated at $D=128,M=400$. There are ten unique configurations per cell. Two further controls retain complete lags $k<3$ or $k<4$, containing 92 or 200 nonzero coefficients. Skew quadratic and logistic also compare nominal, true, and fresh-estimated curvature applied to the same last-400-record reuse sample. Fresh-estimated reuse still incurs its curvature sampling cost, and true-curvature reuse is an oracle diagnostic.

Nominal ranking uses $h_0=1$ (quadratic) or $.6$ (logistic), with the same rule in all comparisons. To specify ties reproducibly, ranking scores are rounded to 14 decimal places and ordered by flat $(k,j)$ index within a tie. Evaluated coefficients are not rounded. This deterministic approximate ranking falls under Theorem~\ref{thm:hybrid}. Proposition~\ref{prop:budget} quantifies the additional score-accuracy requirement for its asymptotic budget. The artifact supplies retained indices and tie rules.

The computed reference uses $B=1499$ independent retained sums and a positive Gaussian residual. A safeguarded Newton/bisection solver evaluates the two interval quantiles, terminating at CDF residual $10^{-10}$ with at most 64 iterations inside a nine-residual-standard-deviation bracket. An independent comparison to the original 35-step bisection gave maximum quantile difference below $10^{-7}$ in the kernel check. Reference CDFs at the observed errors are evaluated directly, with no quantile interpolation needed for calibration curves. Undefined references are included as failures in the reported denominator; none occurred.

\begin{table}[ht]
\caption{Five-law calibration study, nominal 95\% intervals. Tails and coverage are percentages. Skew and logistic use 4,000 trajectories; the other controls use 1,600. Both references share 400 observations and a curvature estimate. Smoothed denotes Gaussian-smoothed skew noise.}\label{tab:revisioncore}
\begin{center}\small
\begin{tabular}{lrrrrrr}
\toprule
& \multicolumn{3}{c}{Gaussian} & \multicolumn{3}{c}{Hybrid-RB}\\
Setting & Lower & Upper & Cov. & Lower & Upper & Cov.\\
\midrule
Skew & 5.43 & 0.13 & 94.45 & 2.48 & 2.73 & 94.80\\
Gaussian & 2.31 & 3.00 & 94.69 & 2.31 & 3.06 & 94.63\\
Smoothed & 3.88 & 0.88 & 95.25 & 2.25 & 2.19 & 95.56\\
Logistic & 4.13 & 1.15 & 94.73 & 2.73 & 2.60 & 94.68\\
Beta & 2.69 & 1.50 & 95.81 & 2.38 & 1.69 & 95.94\\
\bottomrule
\end{tabular}

\end{center}
\end{table}

\subsubsection{Probability calibration and statistical uncertainty}
For replication $r$, compute $U_r=\widetilde F_{B,r}((\widehat\theta_{i,r}-\theta^\star)/\eta)$ using that replication's actual fitted reference. The P--P curve is $\widehat C(p)=N_{\rm rep}^{-1}\sum_r\onevec (U_r\le p)$ against $p$. Lower/upper failure curves use $U_r<p$ and $U_r>1-p$. Central coverage at nominal level $c$ is the proportion with $(1-c)/2\le U_r\le(1+c)/2$. All plotted probabilities are fixed in advance. The exact identity in Appendix~\ref{app:calibrationcurve} explains cancellation of opposite tail errors.

Pointwise Wilson 95\% Monte Carlo intervals describe uncertainty across independent full-network replications. For a finite grid of $J$ probabilities, Hoeffding and a union bound also give a uniform deviation bound $\sqrt{\log(2J/\delta)/(2N_{\rm rep})}$ from the repeated-sampling calibration curve. Conditional reference Monte Carlo is part of the implemented procedure and is separately assessed below. The plotted horizontal axis is the nominal probability of each fitted reference; the vertical axis is its repeated-sampling frequency.

For method comparisons, raw outcomes are paired by replication. For binary tail or coverage differences we report the mean paired difference and its standard error, retain both discordance counts, and compute an exact binomial McNemar $p$ value when useful. These scalar comparisons are descriptive, with pointwise uncertainty. For $\Delta_{\rm tail}=|p_L-.025|+|p_U-.025|$, a 20,000-draw paired multinomial bootstrap of the $3\times3$ joint failure-type table supplies exploratory 95\% intervals. Nonsmoothness at the nominal tails motivates this qualification. The sample sizes and reported configurations were fixed before evaluating this study.

For Hybrid-RB minus Gaussian, the paired coverage difference in skew quadratic loss is $.35$ percentage points, interval $[-.38,1.08]$; in logistic loss it is $-.05$, interval $[-.57,.47]$. The tail-loss changes are $-5.05$ points (exploratory interval $[-5.63,-3.83]$) and $-2.65$ ($[-3.08,-1.65]$), respectively. The Gaussian and bounded-beta controls do not show a clear gain.

\begin{figure}[htbp]
\centering\includegraphics[width=.8\textwidth]{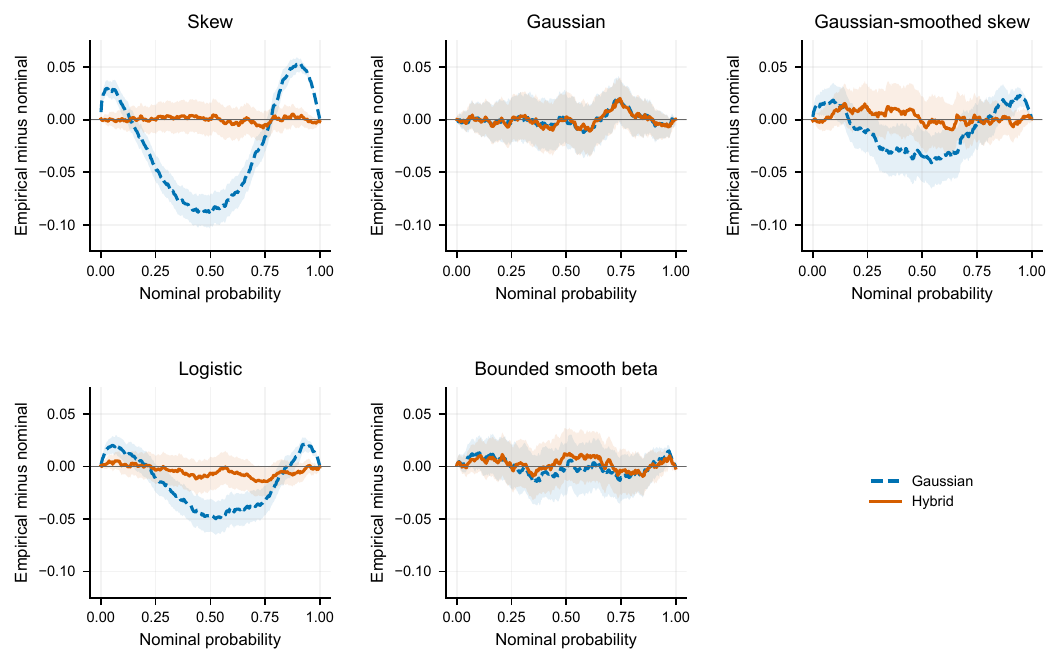}
\caption{All five calibration laws, empirical minus nominal CDF calibration. Shading gives pointwise Wilson 95\% intervals. Neither Gaussian nor bounded-beta controls show the large distortion seen in skew quadratic loss.}\label{fig:allcalibration}
\end{figure}
\begin{figure}[htbp]
\centering\includegraphics[width=.8\textwidth]{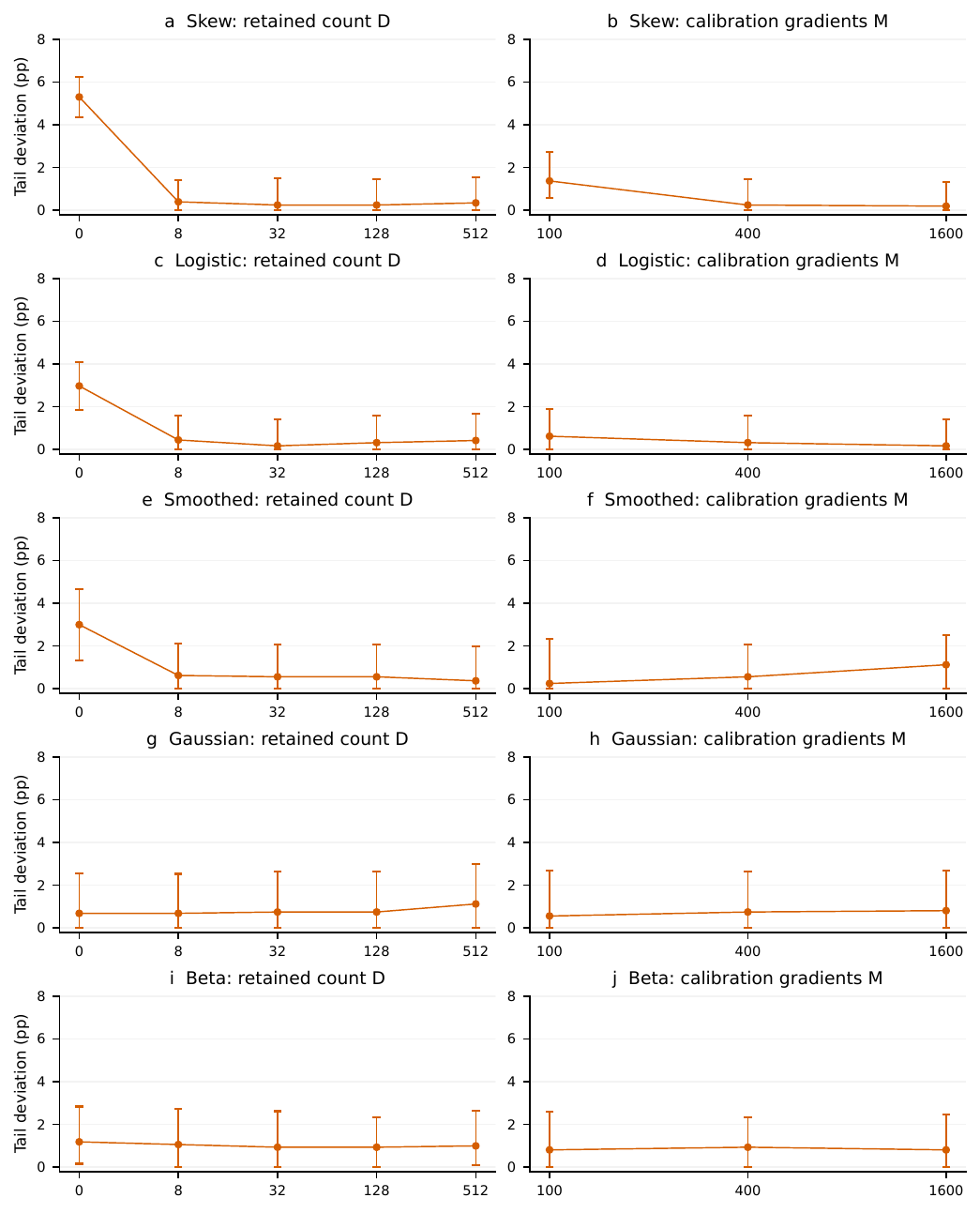}
\caption{Prespecified retention and empirical-law budgets. Total tail deviation is $\Delta_{\rm tail}$, in percentage points. Bars are conservative 95\% intervals obtained from simultaneous exact-binomial intervals for the two tails within each procedure. Lines connect the configured finite-budget settings.}\label{fig:budgets}
\end{figure}

\subsubsection{Reference convergence, near-full filters and cost}
For 64 endpoints per cell, chosen evenly and independently of outcomes, each reference is repeated three times at $B=1499,5999,23999$, with RB and MC sharing retained draws. The root-mean-square across-repeat quantile standard deviation on the parameter scale decreases with $B$: $.00219,.00117,.00060$ for RB and $.00250,.00133,.00070$ for MC in skew quadratic loss, and $.00235,.00122,.00057$ versus $.00283,.00147,.00070$ in logistic loss.

Near-full filters retain all lags below the smallest cutoff whose residual variance fraction is at most $10^{-3}$ for every estimated curvature: $K=34$, with 6,587 coefficients and mean residual fraction $.000477$, and $K=42$, with 8,315 coefficients and $.000151$. On the first 128 skew endpoints their coverage agrees with top-128 retention, at $.141$ versus $.00351$ seconds per endpoint for $B=5999$.

In a timing benchmark matched to equal wall-clock budgets, RB with 1499 draws took $.00619$ seconds per 32-endpoint batch and MC with 1815 draws $.00778$ seconds.

\begin{figure}[htbp]
\centering\includegraphics[width=.8\textwidth]{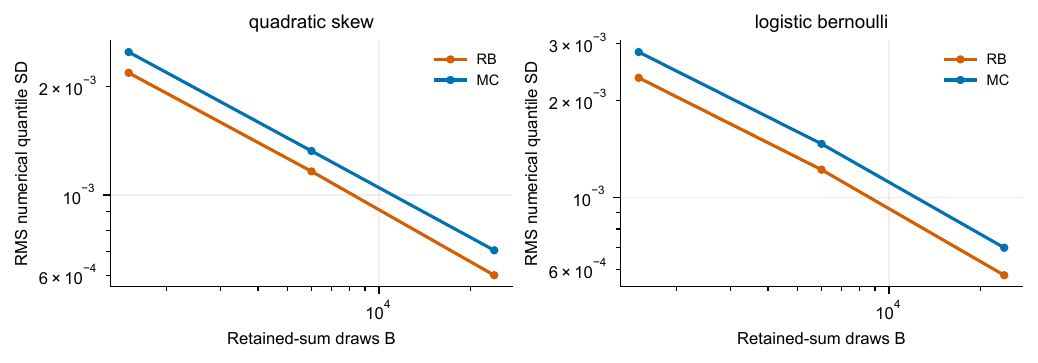}
\caption{Root-mean-square across-repeat quantile SD for independent reference simulations. Each setting uses 64 fixed endpoints and three repetitions at each $B$. Gaussian-tail integration reduces variability at equal retained draws in these tests.}\label{fig:numerical}
\end{figure}

\subsubsection{Deterministic graph-regime diagnostics}
\begin{figure}[t]
\centering\includegraphics[width=.8\textwidth]{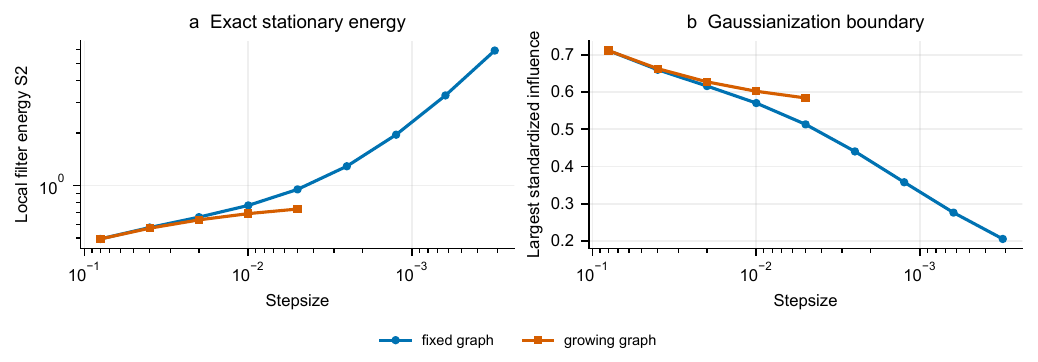}
\caption{Network growth changes the Gaussianization regime. Exact stationary spectral calculations use a fixed $6^3$ torus and a growing sequence $L=6,12,24,48,96$ with $\eta=.08,.04,.02,.01,.005$. As the stepsize decreases from left to right, the fixed graph becomes consensus dominated and its largest influence falls. Growing graphs retain substantial local influence. These finite configurations illustrate the mechanism; they are not fitted limits or coverage simulations.}\label{fig:regimes}
\end{figure}
Figure~\ref{fig:regimes} uses $S_2=m^{-1}\sum_\ell\lambda_\ell^2/(1-a^2\lambda_\ell^2)$ and $\Lambda_i=1/(2\sqrt{S_2})$ on the lazy three-dimensional torus. The consensus fraction is $[m(1-a^2)S_2]^{-1}$. All eigenvalues are evaluated from the explicit cosine formula, without sampling. The fixed graph uses nine halving stepsizes from $.08$ to $.0003125$; the growing sequence uses the five pairs listed in the caption. The study summaries also record the finite-filter skewness \eqref{eq:skewness} for the two skewed calibration cells.

\subsubsection{Reproducibility}
The pipeline stores all trajectory and calibration records, quantiles, PITs, selected coefficients and configurations, and regenerates every table and figure from them. These records total about 370 MB, so the supplementary material contains the code, frozen seeds, design records, study summaries and compact inputs, together with instructions that regenerate the full records; the additional analyses run directly from the compact inputs. All 106 method-setting summaries were checked against the raw outcomes, and 240 numerical quantiles agree with independent Brent inversion to within $7.1\times10^{-8}$.

\begin{longtable}{llrrrrrr}
\caption{Complete calibration study. Tails and coverage are percentages; length is on the parameter scale. True-law and true-curvature rows are oracle diagnostics.}\label{tab:revisionall}\\
\toprule Setting & Method & $M$ & $D$ & Lower & Upper & Cov. & Length\\\midrule\endfirsthead
\toprule Setting & Method & $M$ & $D$ & Lower & Upper & Cov. & Length\\\midrule\endhead
Skew & Gaussian & 400 & 0 & 5.43 & 0.13 & 94.45 & 0.14465\\
Skew & Hybrid & 400 & 8 & 2.45 & 2.15 & 95.40 & 0.14727\\
Skew & Hybrid & 400 & 32 & 2.55 & 2.70 & 94.75 & 0.14552\\
Skew & Hybrid & 400 & 128 & 2.48 & 2.73 & 94.80 & 0.14528\\
Skew & Hybrid & 400 & 512 & 2.48 & 2.83 & 94.70 & 0.14524\\
Skew & Hybrid & 100 & 128 & 2.68 & 3.70 & 93.63 & 0.14418\\
Skew & Hybrid & 1600 & 128 & 2.55 & 2.35 & 95.10 & 0.14559\\
Skew & Hybrid, true h & 400 & 128 & 2.45 & 2.70 & 94.85 & 0.14526\\
Skew & Hybrid, true law & 400 & 128 & 2.58 & 2.23 & 95.20 & 0.14589\\
Skew & Hybrid, true h/law & 400 & 128 & 2.58 & 2.18 & 95.25 & 0.14586\\
Skew & Full lag 3 & 400 & 92 & 2.58 & 2.50 & 94.93 & 0.14617\\
Skew & Full lag 4 & 400 & 200 & 2.50 & 2.65 & 94.85 & 0.14561\\
Skew & Reuse, nominal h & 400 & 128 & 2.38 & 1.73 & 95.90 & 0.15080\\
Skew & Reuse, true h & 400 & 128 & 2.63 & 3.08 & 94.30 & 0.14542\\
Skew & Reuse, fresh h & 400 & 128 & 2.65 & 3.05 & 94.30 & 0.14545\\
Gaussian & Gaussian & 400 & 0 & 2.31 & 3.00 & 94.69 & 0.14503\\
Gaussian & Hybrid & 400 & 8 & 2.25 & 2.94 & 94.81 & 0.14491\\
Gaussian & Hybrid & 400 & 32 & 2.31 & 3.06 & 94.63 & 0.14481\\
Gaussian & Hybrid & 400 & 128 & 2.31 & 3.06 & 94.63 & 0.14498\\
Gaussian & Hybrid & 400 & 512 & 2.13 & 3.25 & 94.63 & 0.14495\\
Gaussian & Hybrid & 100 & 128 & 2.69 & 2.88 & 94.44 & 0.14475\\
Gaussian & Hybrid & 1600 & 128 & 2.25 & 3.06 & 94.69 & 0.14512\\
Gaussian & Hybrid, true h & 400 & 128 & 2.38 & 3.00 & 94.63 & 0.14496\\
Gaussian & Hybrid, true law & 400 & 128 & 2.25 & 2.88 & 94.88 & 0.14522\\
Gaussian & Hybrid, true h/law & 400 & 128 & 2.25 & 2.88 & 94.88 & 0.14520\\
Gaussian & Full lag 3 & 400 & 92 & 2.31 & 3.06 & 94.63 & 0.14489\\
Gaussian & Full lag 4 & 400 & 200 & 2.13 & 2.50 & 95.38 & 0.14491\\
Smoothed & Gaussian & 400 & 0 & 3.88 & 0.88 & 95.25 & 0.14465\\
Smoothed & Hybrid & 400 & 8 & 2.38 & 2.00 & 95.63 & 0.14724\\
Smoothed & Hybrid & 400 & 32 & 2.19 & 2.25 & 95.56 & 0.14680\\
Smoothed & Hybrid & 400 & 128 & 2.25 & 2.19 & 95.56 & 0.14681\\
Smoothed & Hybrid & 400 & 512 & 2.44 & 2.19 & 95.38 & 0.14682\\
Smoothed & Hybrid & 100 & 128 & 2.56 & 2.69 & 94.75 & 0.14526\\
Smoothed & Hybrid & 1600 & 128 & 1.94 & 1.94 & 96.13 & 0.14741\\
Smoothed & Hybrid, true h & 400 & 128 & 2.25 & 2.19 & 95.56 & 0.14680\\
Smoothed & Hybrid, true law & 400 & 128 & 2.25 & 2.06 & 95.69 & 0.14743\\
Smoothed & Hybrid, true h/law & 400 & 128 & 2.25 & 2.19 & 95.56 & 0.14742\\
Smoothed & Full lag 3 & 400 & 92 & 2.13 & 2.06 & 95.81 & 0.14689\\
Smoothed & Full lag 4 & 400 & 200 & 2.44 & 2.31 & 95.25 & 0.14679\\
Logistic & Gaussian & 400 & 0 & 4.13 & 1.15 & 94.73 & 0.16544\\
Logistic & Hybrid & 400 & 8 & 2.85 & 2.40 & 94.75 & 0.16354\\
Logistic & Hybrid & 400 & 32 & 2.68 & 2.50 & 94.83 & 0.16315\\
Logistic & Hybrid & 400 & 128 & 2.73 & 2.60 & 94.68 & 0.16312\\
Logistic & Hybrid & 400 & 512 & 2.73 & 2.70 & 94.58 & 0.16300\\
Logistic & Hybrid & 100 & 128 & 2.73 & 2.90 & 94.38 & 0.16210\\
Logistic & Hybrid & 1600 & 128 & 2.60 & 2.58 & 94.83 & 0.16348\\
Logistic & Hybrid, true h & 400 & 128 & 2.70 & 2.75 & 94.55 & 0.16310\\
Logistic & Hybrid, true law & 400 & 128 & 2.63 & 2.55 & 94.83 & 0.16349\\
Logistic & Hybrid, true h/law & 400 & 128 & 2.58 & 2.70 & 94.73 & 0.16346\\
Logistic & Full lag 3 & 400 & 92 & 2.73 & 2.48 & 94.80 & 0.16330\\
Logistic & Full lag 4 & 400 & 200 & 2.60 & 2.55 & 94.85 & 0.16321\\
Logistic & Reuse, nominal h & 400 & 128 & 2.63 & 2.70 & 94.68 & 0.16338\\
Logistic & Reuse, true h & 400 & 128 & 2.65 & 2.73 & 94.63 & 0.16307\\
Logistic & Reuse, fresh h & 400 & 128 & 2.68 & 2.65 & 94.68 & 0.16310\\
Beta & Gaussian & 400 & 0 & 2.69 & 1.50 & 95.81 & 0.14493\\
Beta & Hybrid & 400 & 8 & 2.69 & 1.63 & 95.69 & 0.14427\\
Beta & Hybrid & 400 & 32 & 2.63 & 1.69 & 95.69 & 0.14420\\
Beta & Hybrid & 400 & 128 & 2.38 & 1.69 & 95.94 & 0.14433\\
Beta & Hybrid & 400 & 512 & 2.56 & 1.56 & 95.88 & 0.14427\\
Beta & Hybrid & 100 & 128 & 2.81 & 2.00 & 95.19 & 0.14362\\
Beta & Hybrid & 1600 & 128 & 2.50 & 1.69 & 95.81 & 0.14452\\
Beta & Hybrid, true h & 400 & 128 & 2.38 & 1.75 & 95.88 & 0.14429\\
Beta & Hybrid, true law & 400 & 128 & 2.44 & 1.63 & 95.94 & 0.14451\\
Beta & Hybrid, true h/law & 400 & 128 & 2.44 & 1.69 & 95.88 & 0.14447\\
Beta & Full lag 3 & 400 & 92 & 2.63 & 1.63 & 95.75 & 0.14434\\
Beta & Full lag 4 & 400 & 200 & 2.63 & 1.63 & 95.75 & 0.14432\\
\bottomrule\end{longtable}
\subsection{Network regimes and alternative estimators}\label{app:v3experiments}
\subsubsection{Independent network experiments}
The graph-regime study contains five cells, each with 4000 independent full-network trajectories. Four use the skew quadratic innovation law from Appendix~\ref{app:experiments}, at $(L,\eta)=(6,.005),(6,.0005),(12,.05),(12,.005)$. The fifth uses symmetric $\pm1$ innovations with equal probability at $(6,.05)$. All have $h=1.4$, target $.35$, zero initial state, $q=0$, $M=400$, two 100-gradient curvature samples, projection onto $[.3,2]$, and $B=1499$ for retained-reference methods. Gaussian references are analytic. The fixed nominal curvature is one. Each method uses the same endpoint and calibration observations. The independent $(6,.05)$ skew cell from the scalar calibration study supplies the baseline in Figure~\ref{fig:empiricalregimes}.

For a common deterministic initial error of magnitude $E_0$, unit-variance innovations and $W_{ii}\ge w_0$, the linear initial shift is bounded by $(1-\eta h)^TE_0/(\eta w_0)$. To make it at most $\epsilon_{\rm init}$, choose
\[
 T\ge\left\lceil\frac{\log(\epsilon_{\rm init}\eta w_0/E_0)}{\log(1-\eta h)}\right\rceil.
\]
With $E_0=.35$, $w_0=.5$ and $\epsilon_{\rm init}=10^{-4}$, the implemented horizons are
\[
 T=\max\left\{410,\left\lceil
 \frac{\log(10^{-4}\eta\cdot .5/.35)}{\log(1-1.4\eta)}
 \right\rceil\right\}.
\]
Since $\sqrt{S_2}\ge.5$, the normalized initialization is at most $10^{-4}$. Root seeds, in the order above, are $1027000101,1027000102,1027000103,1027000104,1027000201$. Endpoint trajectories use a compiled sparse-matrix recursion checked against direct SciPy multiplication. Fresh calibration, retained resampling and reference-sensitivity streams have separate recorded offsets. Runtime pilot observations are excluded from inferential summaries.

The Gaussian, nominal top-128 Hybrid and complete-lag-3 references use the exact finite filter energy. Lag 3 retains 92 coefficients on the $6^3$ torus and 95 on the $12^3$ torus; periodic identifications explain the difference. Exact spectral energy, largest influence, consensus fraction, and standardized third/fourth moments are stored for every cell. With unit innovation variance and fourth cumulant $\kappa_4$,
\[
 \E Y^4=3+\kappa_4\frac{\sum c_{kj}^4}{S_2^2}.
\]
Here $\kappa_4=46/9$ for the skew law and $-2$ for Rademacher noise. This identity complements the third-moment diagnostic without replacing a CDF comparison.

\begin{table}[ht]
\caption{Graph-regime and symmetry study. Tail errors and coverage are percentages. The final column is a conservative 95\% interval for $\Delta_{\rm tail}$ in percentage points, simultaneous over the two tails of that row. All rows use $N_{\rm rep}=4000$ and $M=400$. Retained-reference methods use $B=1499$; Gaussian uses no reference draws.}\label{tab:v3network}
\begin{center}\small\setlength{\tabcolsep}{4pt}
\begin{tabular}{llrrrrr}
\toprule Graph / step & Method & Lower & Upper & Cov. & Length & $\Delta_{\rm tail}$ interval\\\midrule
$12^3$, $.005$ & Gaussian & 4.33 & 0.58 & 95.10 & 0.01682 & [2.73, 4.76]\\
$12^3$, $.005$ & Hybrid & 2.70 & 2.35 & 94.95 & 0.01698 & [0.00, 1.49]\\
$12^3$, $.005$ & Full lag 3 & 2.75 & 2.08 & 95.18 & 0.01706 & [0.00, 1.79]\\
$12^3$, $.05$ & Gaussian & 5.73 & 0.05 & 94.23 & 0.14418 & [4.73, 6.60]\\
$12^3$, $.05$ & Hybrid & 2.55 & 2.93 & 94.53 & 0.14473 & [0.00, 1.75]\\
$12^3$, $.05$ & Full lag 3 & 2.50 & 2.65 & 94.85 & 0.14561 & [0.00, 1.39]\\
$6^3$, $.0005$ & Gaussian & 2.65 & 2.45 & 94.90 & 0.00389 & [0.00, 1.34]\\
$6^3$, $.0005$ & Hybrid & 2.50 & 2.55 & 94.95 & 0.00389 & [0.00, 1.28]\\
$6^3$, $.0005$ & Full lag 3 & 2.45 & 2.55 & 95.00 & 0.00389 & [0.00, 1.23]\\
$6^3$, $.005$ & Gaussian & 4.05 & 1.45 & 94.50 & 0.01908 & [1.45, 3.75]\\
$6^3$, $.005$ & Hybrid & 2.58 & 3.03 & 94.40 & 0.01927 & [0.00, 1.88]\\
$6^3$, $.005$ & Full lag 3 & 2.68 & 2.83 & 94.50 & 0.01932 & [0.00, 1.77]\\
$6^3$, $.05$, symmetric & Gaussian & 1.80 & 2.13 & 96.08 & 0.14504 & [0.17, 1.99]\\
$6^3$, $.05$, symmetric & Hybrid & 2.10 & 2.63 & 95.28 & 0.13978 & [0.00, 1.63]\\
$6^3$, $.05$, symmetric & Full lag 3 & 2.18 & 2.65 & 95.18 & 0.13978 & [0.00, 1.59]\\
\bottomrule\end{tabular}

\end{center}
\end{table}

The first 128 endpoints of each cell also receive references with $B=1499,5999,23999$ to measure numerical sensitivity.

\subsubsection{Uncertainty, pairing and multiplicity}
For tail counts $k_L,k_U$, two Clopper--Pearson intervals at level $.975$ give joint coverage at least $.95$ for $(p_L,p_U)$ by Bonferroni. Mapping their product through $|p_L-.025|+|p_U-.025|$ gives a conservative interval for $\Delta_{\rm tail}$ that remains valid at the nonsmooth nominal point.

For the entire repeated-sampling PIT CDF $C$, DKW gives
\[
 \Pp\left\{\sup_p|\widehat C(p)-C(p)|>
 \sqrt{\frac{\log(2/\delta)}{2N_{\rm rep}}}\right\}\le\delta.
\]
The reverse triangle inequality gives an interval of this radius around $\sup_p|\widehat C(p)-p|$. It includes calibration and reference randomness because each replication runs the complete fitted procedure.

The graph-regime family has 36 exact McNemar tests (two retained references versus Gaussian, three binary outcomes, six cells) with Holm correction. They test equality of outcome probabilities, so a significant change can improve or worsen calibration relative to the nominal level. Tail-loss difference intervals use 20,000 paired multinomial bootstrap draws and are exploratory because the absolute tail-loss functional is not differentiable at nominal calibration.

Hybrid-minus-Gaussian tail-loss intervals are $[-.35,.30]$ points for $(6,.0005)$, $[-.95,.28]$ for the symmetric cell, and $[-5.88,-4.08]$ and $[-3.80,-2.60]$ on $12^3$ at $\eta=.05$ and $.005$.

\subsubsection{Matched residual reuse}
Each Gaussian reuse row uses exactly the same last-400 trajectory records and curvature choice as its Hybrid counterpart. Only the reference distribution changes. Nominal and fresh-estimated curvature are feasible; the true-curvature rows use the root Hessian $h=f''(\theta^\star)$ as an oracle diagnostic. Fresh-estimated reuse still incurs the curvature-data cost. The shape correction persists in these matched comparisons; nominal curvature can also change scale, which is why the fresh and true variants are shown.
\begin{table}[ht]
\caption{Matched residual-reuse comparisons, $N_{\rm rep}=4000$ for each model. Tail errors are percentages; lengths are on the parameter scale. Curvature choices are nominal, root curvature $f''(\theta^\star)$, or estimated from fresh data. The population curvature at the endpoint used in the earlier mechanism study is a separate control.}\label{tab:v3reuse}
\begin{center}\small
\begin{tabular}{llrrrrrr}
\toprule Model & Curvature & \multicolumn{3}{c}{Gaussian reuse} & \multicolumn{3}{c}{Hybrid reuse}\\
& & Lower & Upper & Length & Lower & Upper & Length\\\midrule
Skew & Nominal & 5.15 & 0.13 & 0.14985 & 2.38 & 1.73 & 0.15080\\
Skew & True & 5.43 & 0.18 & 0.14477 & 2.63 & 3.08 & 0.14542\\
Skew & Fresh & 5.38 & 0.20 & 0.14480 & 2.65 & 3.05 & 0.14545\\
Logistic & Nominal & 3.95 & 1.33 & 0.16562 & 2.63 & 2.70 & 0.16338\\
Logistic & True & 4.03 & 1.33 & 0.16533 & 2.65 & 2.73 & 0.16307\\
Logistic & Fresh & 4.05 & 1.30 & 0.16536 & 2.68 & 2.65 & 0.16310\\
\bottomrule\end{tabular}

\end{center}
\end{table}

\subsubsection{A feasible one-step alternative using the same fresh observations}
Given endpoint $x$, let $\bar g$ be the mean of its 400 fresh gradients and $\widehat h$ the existing curvature estimate. Define $\widetilde\theta=x-\bar g/\widehat h$. This changes the estimator; it is evaluated separately from references for the unchanged endpoint. The available information and gradient/Hessian counts are the same as for fresh endpoint calibration.

For logistic loss, write $H_j=.1+X_j^2s_j(1-s_j)$ for the sample Hessian. The Gaussian reference uses the sample variance of
\[
 I_j=\frac{g_j-\bar g}{\widehat h}
       -\frac{\bar g(H_j-\widehat h)}{\widehat h^2},
 \qquad \widehat{\mathrm{se}}^2=\frac{\widehat\Var(I_j)}{400}.
\]
This delta-method estimate includes gradient/Hessian covariance. In quadratic loss, the 400 gradient samples and two 100-gradient curvature samples are independent conditional on the endpoint. Its variance estimate is
\[
 \widehat{\mathrm{se}}^2=
 \frac{\widehat\Var(g)}{400\widehat h^2}
 +\frac{\bar g^2}{4\widehat h^4}
 \left\{\frac{\widehat\Var(g_+)}{100}
       +\frac{\widehat\Var(g_-)}{100}\right\}.
\]
All sample variances use the usual degrees-of-freedom correction. Gaussian intervals are $\widetilde\theta\pm z_{.975}\widehat{\mathrm{se}}$.

The basic bootstrap holds $x$ fixed, jointly draws 400 logistic gradient/Hessian pairs with replacement, and independently draws 400, 100 and 100 observations with replacement from the three quadratic samples. In each of 1499 draws it recomputes the projected curvature and $\widetilde\theta^*=x-\bar g^*/\widehat h^*$. The interval is $[\widetilde\theta-q^*_{.975},\widetilde\theta-q^*_{.025}]$, where $q^*$ are quantiles of $\widetilde\theta^*-\widetilde\theta$. The bootstrap assesses this one-step ratio statistic using the saved observations. Its finite-sample bias and calibration are evaluated empirically alongside the endpoint references.

\begin{table}[ht]
\caption{Estimator-changing comparison on identical saved endpoints and fresh observations, $N_{\rm rep}=4000$ per model. Tail errors and coverage are percentages. Bias and RMSE refer to the estimator in that row; different references for an unchanged estimator have the same bias/RMSE.}\label{tab:v3onestep}
\begin{center}\small\setlength{\tabcolsep}{3.5pt}
\begin{tabular}{llrrrrrr}
\toprule Model & Estimator / reference & Bias & RMSE & Lower & Upper & Cov. & Length\\\midrule
Skew & Endpoint, Gaussian & -0.00003 & 0.03664 & 5.43 & 0.13 & 94.45 & 0.14465\\
Skew & Endpoint, hybrid & -0.00003 & 0.03664 & 2.48 & 2.73 & 94.80 & 0.14528\\
Skew & Endpoint, full lag 3 & -0.00003 & 0.03664 & 2.58 & 2.50 & 94.93 & 0.14617\\
Skew & One-step, Gaussian & 0.00028 & 0.03622 & 1.38 & 3.58 & 95.05 & 0.14041\\
Skew & One-step, bootstrap & 0.00028 & 0.03622 & 1.25 & 4.25 & 94.50 & 0.14039\\
Logistic & Endpoint, Gaussian & 0.00049 & 0.04272 & 4.13 & 1.15 & 94.73 & 0.16544\\
Logistic & Endpoint, hybrid & 0.00049 & 0.04272 & 2.73 & 2.60 & 94.68 & 0.16312\\
Logistic & Endpoint, full lag 3 & 0.00049 & 0.04272 & 2.73 & 2.48 & 94.80 & 0.16330\\
Logistic & One-step, Gaussian & -0.00019 & 0.05759 & 1.95 & 3.93 & 94.13 & 0.22089\\
Logistic & One-step, bootstrap & -0.00019 & 0.05759 & 1.98 & 4.13 & 93.90 & 0.21989\\
\bottomrule\end{tabular}

\end{center}
\end{table}

In skew quadratic loss the one-step estimator has similar RMSE and slightly shorter intervals, but its tail allocation differs from the endpoint Hybrid reference. In logistic loss the one-step RMSE is $.05759$, versus $.04272$ for the endpoint, and its intervals are longer at $M=400$. This illustrates the statistical cost of substituting a finite fresh-sample correction for the network's accumulated information.

\subsection{Moment-correction baselines}\label{app:momentexperiments}
\paragraph{Definitions.}
Moment corrections use the same endpoint, the same $M=400$ fresh gradients and the same curvature estimate as the Gaussian and hybrid references, and require no further observations. Let $\widehat\gamma_\eps$ and $\widehat\kappa_\eps$ be the sample skewness and excess kurtosis of the centered gradients, with denominator $M$, and evaluate the full finite filter at $\widehat h$. By \eqref{eq:skewness} and its fourth-cumulant analogue, the estimated endpoint skewness and excess kurtosis are
\[
 \widehat\gamma_Y=-\widehat\gamma_\eps\frac{\sum\widehat c_{kj}^3}{S_2(\widehat h)^{3/2}},
 \qquad
 \widehat\kappa_Y=\widehat\kappa_\eps\frac{\sum\widehat c_{kj}^4}{S_2(\widehat h)^2}.
\]
With $z_p=\Phi^{-1}(p)$, CF$_3$ uses the standardized quantile $z_p+\widehat\gamma_Y(z_p^2-1)/6$, and CF$_4$ adds $\widehat\kappa_Y(z_p^3-3z_p)/24-\widehat\gamma_Y^2(2z_p^3-5z_p)/36$ \citep{cornish1938,hall1992}. Edgeworth inverts the two-term expansion $\Phi(x)-\phi(x)\{\widehat\gamma_Y\mathrm{He}_2(x)/6+\widehat\kappa_Y\mathrm{He}_3(x)/24+\widehat\gamma_Y^2\mathrm{He}_5(x)/72\}$ after clipping it to $[0,1]$ and taking its monotone envelope on a grid. Standardized quantiles are multiplied by $\eta\widehat\sigma\sqrt{S_2(\widehat h)}$ and inserted in \eqref{eq:ci}. Because $z_p^2$ is the same at both levels, CF$_3$ shifts both Gaussian quantiles equally and keeps the Gaussian length. PIT values invert the monotone envelope of the Cornish--Fisher map. Oracle variants replace $\widehat\gamma_\eps$, $\widehat\kappa_\eps$, $\widehat\sigma$ and $\widehat h$ by their true values. Hybrid and complete lag-3 summaries are recomputed from the saved references.

\paragraph{Results.}
Table~\ref{tab:momentall} reports the ten scalar cells of the primary studies; Appendix~\ref{app:rareshock} treats the rare-shock cell. CF$_4$ is not monotone for Rademacher noise, whose excess kurtosis is negative, so its PIT summaries are omitted there. Exact McNemar tests of Hybrid against CF$_3$ for the lower tail, upper tail and coverage in the ten cells form a Holm family of 30 tests. Only two cells reject after adjustment: logistic loss (upper tail, $p=.0002$; coverage, $p=.015$) and Rademacher noise (upper tail, $p=.0006$; coverage, $p<10^{-4}$). In both, Hybrid has the upper tail closer to nominal. For the difference CF$_3$ minus Hybrid, 2,000 paired bootstrap resamples of trajectories give 95\% intervals for the PIT discrepancy of $[.015,.039]$ ($6^3$, $\eta=.05$), $[.014,.034]$ ($12^3$, $\eta=.05$), $[-.007,.015]$ ($12^3$, $\eta=.005$), $[.001,.010]$ (logistic) and $[-.010,.012]$ (smoothed). The corresponding MAE intervals are $[.31,.71]$, $[.08,.57]$, $[.01,.31]$, $[-.11,.27]$ and $[-.16,.14]$. Both summaries are nonsmooth, so these intervals are exploratory.

With true cumulants, scale and curvature in the $6^3$, $\eta=.05$ skew cell, CF$_3$ and CF$_4$ have tails $2.75/1.90$ and $3.05/3.05$ and PIT discrepancies $.044$ and $.043$, against $2.48/2.73$ and $.008$ for the fitted hybrid reference. In the $12^3$, $\eta=.005$ cell the oracle values are $2.68/2.13$ and $2.85/2.48$, with PIT discrepancies $.019$ and $.017$. With the same 400 fresh gradients and curvature estimate, the one-step estimate $x-\bar g/\widehat h$ has RMSE $.0356$ and $.0361$ in the $12^3$ and $6^3$ cells with $\eta=.005$, against endpoint RMSEs of $.0043$ and $.0049$; for $(6,.0005)$ the values are $.0358$ and $.0010$.

\begingroup\small
\begin{longtable}{llrrrrrr}
\caption{Moment corrections and hybrid references in the scalar cells of the primary studies. Columns as in Table~\ref{tab:maincalibration}; all references share endpoints, fresh gradients and curvature estimates.}\label{tab:momentall}\\
\toprule Setting & Reference & Low & Up & Cov. & Len. & PIT & MAE\\\midrule\endfirsthead
\toprule Setting & Reference & Low & Up & Cov. & Len. & PIT & MAE\\\midrule\endhead
\bottomrule\endfoot
Skew, $12^3$, $\eta=.005$ ($N=4000$) & Gaussian & 4.33 & 0.58 & 95.10 & .0168 & .057 & 1.66\\
 & CF$_3$ & 2.83 & 2.48 & 94.70 & .0168 & .017 & .35\\
 & CF$_4$ & 3.03 & 2.78 & 94.20 & .0166 & .015 & .38\\
 & Edgeworth & 2.90 & 2.50 & 94.60 & .0168 & .014 & .41\\
 & Hybrid & 2.70 & 2.35 & 94.95 & .0170 & .012 & .13\\
 & Lag 3 & 2.75 & 2.08 & 95.18 & .0171 & .012 & .17\\
\addlinespace[3pt]
Skew, $12^3$, $\eta=.05$ ($N=4000$) & Gaussian & 5.73 & 0.05 & 94.23 & .1442 & .100 & 2.44\\
 & CF$_3$ & 2.70 & 2.83 & 94.48 & .1442 & .039 & .65\\
 & CF$_4$ & 3.05 & 3.98 & 92.98 & .1381 & .039 & 1.08\\
 & Edgeworth & 2.70 & 2.33 & 94.98 & .1455 & .039 & .65\\
 & Hybrid & 2.55 & 2.93 & 94.53 & .1447 & .013 & .27\\
 & Lag 3 & 2.50 & 2.65 & 94.85 & .1456 & .015 & .21\\
\addlinespace[3pt]
Skew, $6^3$, $\eta=.005$ ($N=4000$) & Gaussian & 4.05 & 1.45 & 94.50 & .0191 & .038 & 1.19\\
 & CF$_3$ & 2.75 & 3.08 & 94.18 & .0191 & .019 & .40\\
 & CF$_4$ & 2.78 & 3.13 & 94.10 & .0190 & .015 & .40\\
 & Edgeworth & 2.73 & 3.13 & 94.15 & .0191 & .015 & .44\\
 & Hybrid & 2.58 & 3.03 & 94.40 & .0193 & .009 & .25\\
 & Lag 3 & 2.68 & 2.83 & 94.50 & .0193 & .011 & .22\\
\addlinespace[3pt]
Skew, $6^3$, $\eta=.0005$ ($N=4000$) & Gaussian & 2.65 & 2.45 & 94.90 & .0039 & .017 & .15\\
 & CF$_3$ & 2.50 & 2.60 & 94.90 & .0039 & .014 & .15\\
 & CF$_4$ & 2.50 & 2.55 & 94.95 & .0039 & .014 & .14\\
 & Edgeworth & 2.50 & 2.58 & 94.93 & .0039 & .014 & .14\\
 & Hybrid & 2.50 & 2.55 & 94.95 & .0039 & .014 & .15\\
 & Lag 3 & 2.45 & 2.55 & 95.00 & .0039 & .013 & .15\\
\addlinespace[3pt]
Skew, $6^3$, $\eta=.05$ ($N=4000$) & Gaussian & 5.43 & 0.13 & 94.45 & .1447 & .089 & 2.42\\
 & CF$_3$ & 2.63 & 2.53 & 94.85 & .1447 & .044 & .76\\
 & CF$_4$ & 2.98 & 3.73 & 93.30 & .1387 & .041 & 1.12\\
 & Edgeworth & 2.65 & 2.20 & 95.15 & .1459 & .041 & .79\\
 & Hybrid & 2.48 & 2.73 & 94.80 & .1453 & .008 & .12\\
 & Lag 3 & 2.58 & 2.50 & 94.93 & .1462 & .012 & .10\\
\addlinespace[3pt]
Smoothed, $6^3$, $\eta=.05$ ($N=1600$) & Gaussian & 3.88 & 0.88 & 95.25 & .1446 & .042 & 1.28\\
 & CF$_3$ & 2.56 & 2.31 & 95.13 & .1446 & .017 & .21\\
 & CF$_4$ & 2.56 & 2.50 & 94.94 & .1440 & .014 & .19\\
 & Edgeworth & 2.50 & 2.50 & 95.00 & .1446 & .014 & .24\\
 & Hybrid & 2.25 & 2.19 & 95.56 & .1468 & .016 & .22\\
 & Lag 3 & 2.13 & 2.06 & 95.81 & .1469 & .016 & .29\\
\addlinespace[3pt]
Gaussian, $6^3$, $\eta=.05$ ($N=1600$) & Gaussian & 2.31 & 3.00 & 94.69 & .1450 & .022 & .24\\
 & CF$_3$ & 2.19 & 3.00 & 94.81 & .1450 & .022 & .25\\
 & CF$_4$ & 2.25 & 3.00 & 94.75 & .1450 & .022 & .23\\
 & Edgeworth & 2.25 & 3.00 & 94.75 & .1450 & .022 & .23\\
 & Hybrid & 2.31 & 3.06 & 94.63 & .1450 & .021 & .23\\
 & Lag 3 & 2.31 & 3.06 & 94.63 & .1449 & .020 & .19\\
\addlinespace[3pt]
Beta, $6^3$, $\eta=.05$ ($N=1600$) & Gaussian & 2.69 & 1.50 & 95.81 & .1449 & .016 & .63\\
 & CF$_3$ & 2.56 & 1.56 & 95.88 & .1449 & .016 & .38\\
 & CF$_4$ & 2.56 & 1.56 & 95.88 & .1444 & .016 & .39\\
 & Edgeworth & 2.56 & 1.56 & 95.88 & .1444 & .016 & .39\\
 & Hybrid & 2.38 & 1.69 & 95.94 & .1443 & .014 & .40\\
 & Lag 3 & 2.63 & 1.63 & 95.75 & .1443 & .017 & .39\\
\addlinespace[3pt]
Logistic, $6^3$, $\eta=.08$ ($N=4000$) & Gaussian & 4.13 & 1.15 & 94.73 & .1654 & .050 & 1.24\\
 & CF$_3$ & 2.70 & 2.15 & 95.15 & .1654 & .021 & .32\\
 & CF$_4$ & 2.80 & 2.38 & 94.83 & .1630 & .022 & .35\\
 & Edgeworth & 2.85 & 2.15 & 95.00 & .1635 & .021 & .40\\
 & Hybrid & 2.73 & 2.60 & 94.68 & .1631 & .015 & .21\\
 & Lag 3 & 2.73 & 2.48 & 94.80 & .1633 & .017 & .23\\
\addlinespace[3pt]
Rademacher, $6^3$, $\eta=.05$ ($N=4000$) & Gaussian & 1.80 & 2.13 & 96.08 & .1450 & .039 & .63\\
 & CF$_3$ & 1.88 & 2.15 & 95.98 & .1450 & .038 & .62\\
 & CF$_4$ & 2.03 & 2.35 & 95.63 & .1427 & -- & --\\
 & Edgeworth & 2.03 & 2.35 & 95.63 & .1430 & .029 & .37\\
 & Hybrid & 2.10 & 2.63 & 95.28 & .1398 & .023 & .32\\
 & Lag 3 & 2.18 & 2.65 & 95.18 & .1398 & .025 & .32\\
\end{longtable}
\endgroup

\subsubsection{Rare large shocks}\label{app:rareshock}
This cell keeps the $12^3$ torus, $\eta=.005$, $h=1.4$, the horizon $T=2015$ and the calibration protocol of the main local cell, and replaces the skew law by $\xi=7$ with probability $.02$ and $\xi=-1/7$ otherwise, which has skewness $6.86$ and excess kurtosis $45.0$. It uses 4,000 new full-network trajectories with seed root 1027000301 and was specified after the primary studies; the shock probability was chosen with the exact-law calculation of Table~\ref{tab:momentfloors}, because with probability $.01$ about $1.8\%$ of 400-gradient samples would contain no shock. One endpoint had no shock among its 400 calibration gradients, so all its references are undefined. Elsewhere undefined references are counted as failures (Appendices K.2.1 and K.5.4); here we exclude this endpoint from every comparison, leaving 3,999 paired endpoints. Counting it as a non-covering failure would lower every coverage in Table~\ref{tab:rareshock} by at most $.025$ points and change each tail rate by less than $.004$ points. The budgets $M=1600$ and $6400$ and the known-law reference were added after the $M=400$ outcomes had been examined. They use the same endpoints and curvature estimates with independent calibration samples. A second follow-up, specified after these outcomes, calibrates from the client's own trajectory. The trajectories are regenerated with the same seeds, which reproduces every endpoint exactly, and client 0's last $K$ gradients are detrended with $\widehat h$ and centered using the fitted-curvature reuse construction of Appendix~\ref{app:curvature}, for $K=400$ and $1600$. One further endpoint has no shock among its last 400 residuals; it is excluded from the $K=400$ comparison only, and counting both endpoints as failures would lower the $K=400$ coverages by at most $.05$ points. At $M=400$ the calibration resources equal those of the main local cell (Appendix~\ref{app:resources}); the larger fresh budgets change only the number of gradients per endpoint, and reuse stores $K$ gradient--state pairs without further law-sampling gradients, retaining the 200 fresh gradient calls for curvature estimation.

Table~\ref{tab:rareshock} reports every reference. At $M=400$, Hybrid lowers the upper error of CF$_3$ by $6.33$ points, with 253 against 0 discordant endpoints, and raises coverage by $5.90$ points. Relative to Gaussian it raises the upper error by $4.28$ points and lowers the lower error by $1.25$ points; all six tests have Holm-adjusted $p<.003$. For CF$_3$ minus Hybrid, 2,000 paired bootstrap resamples give $[.064,.097]$ for the PIT discrepancy and $[1.77,2.38]$ points for the mean absolute one-sided error. The Hybrid errors at $M=400$ concentrate where the calibration sample underrepresents shocks: the 365 endpoints with at most four shocks among the 400 gradients have tails $6.30\%/21.37\%$, and the 900 endpoints with nine or ten shocks have $2.11\%/1.11\%$. With $B=1499$, $5999$ and $23{,}999$ draws, the first 128 endpoints have the same tail error rates. The moment corrections improve little with $M$, in line with their population distances in Table~\ref{tab:momentfloors}.

With 1,600 reused residuals, Hybrid has tails $2.43\%/2.73\%$, coverage $94.85\%$ and PIT discrepancy $.013$, close to its values with 1,600 fresh gradients. Relative to Hybrid with 400 fresh gradients on the same endpoints, its upper error falls by $2.15$ points, with 120 against 34 discordant endpoints, and its coverage rises by $2.73$ points (both Holm-adjusted $p<10^{-6}$). CF$_3$ computed from the same residuals still misses $8.83\%$ above; Hybrid lowers that upper error by $6.10$ points, with 244 against 0 discordant endpoints, and the paired bootstrap intervals for CF$_3$ minus Hybrid are $[.070,.108]$ for the PIT discrepancy and $[2.25,2.78]$ points for the mean absolute one-sided error. With $K=400$ reused residuals the results resemble those with 400 fresh gradients.
\begin{table}[ht]
\caption{Rare-shock cell on the $12^3$ torus with $\eta=.005$ (3,999 paired endpoints; 3,998 for reuse with $K=400$). Columns as in Table~\ref{tab:maincalibration}. $M$: fresh gradients at the endpoint; reuse: the client's last $K$ trajectory residuals. The known-law reference uses the true innovation law and scale with the estimated curvature.}\label{tab:rareshock}
\begin{center}\small\setlength{\tabcolsep}{3.2pt}
\begin{tabular}{llrrrrrr}
\toprule Budget & Reference & Low & Up & Cov. & Len. & PIT & MAE\\\midrule
$M=400$ & Gaussian & 4.25 & 0.60 & 95.15 & .0167 & .100 & 2.00\\
 & CF$_3$ & 2.58 & 11.20 & 86.22 & .0167 & .107 & 3.15\\
 & CF$_4$ & 2.73 & 14.05 & 83.22 & .0159 & .115 & 5.12\\
 & Edgeworth & 1.90 & 7.43 & 90.67 & .0190 & .064 & 2.58\\
 & Hybrid & 3.00 & 4.88 & 92.12 & .0174 & .025 & 1.08\\
 & Lag 3 & 3.05 & 3.75 & 93.20 & .0176 & .026 & .74\\
\addlinespace[3pt]
$M=1600$ & Gaussian & 4.30 & 0.08 & 95.62 & .0168 & .100 & 2.35\\
 & CF$_3$ & 2.30 & 8.70 & 89.00 & .0168 & .111 & 2.87\\
 & CF$_4$ & 2.50 & 11.23 & 86.27 & .0160 & .087 & 4.08\\
 & Edgeworth & 1.93 & 5.23 & 92.85 & .0191 & .056 & 1.62\\
 & Hybrid & 2.45 & 2.68 & 94.87 & .0177 & .014 & .15\\
\addlinespace[3pt]
$M=6400$ & Gaussian & 3.93 & 0.08 & 96.00 & .0169 & .102 & 2.31\\
 & CF$_3$ & 2.28 & 7.73 & 90.00 & .0169 & .111 & 2.61\\
 & CF$_4$ & 2.50 & 10.30 & 87.20 & .0161 & .079 & 3.58\\
 & Edgeworth & 1.85 & 4.70 & 93.45 & .0193 & .058 & 1.38\\
 & Hybrid & 2.38 & 2.18 & 95.45 & .0179 & .011 & .10\\
\addlinespace[3pt]
Reuse $K=400$ & Gaussian & 4.43 & 0.65 & 94.92 & .0166 & .100 & 1.93\\
 & CF$_3$ & 2.13 & 12.63 & 85.24 & .0166 & .101 & 3.58\\
 & CF$_4$ & 2.25 & 15.63 & 82.12 & .0158 & .130 & 5.77\\
 & Edgeworth & 1.80 & 8.00 & 90.20 & .0189 & .068 & 2.96\\
 & Hybrid & 2.65 & 4.80 & 92.55 & .0173 & .028 & 1.17\\
\addlinespace[3pt]
Reuse $K=1600$ & Gaussian & 4.13 & 0.08 & 95.80 & .0168 & .101 & 2.27\\
 & CF$_3$ & 2.35 & 8.83 & 88.82 & .0168 & .109 & 2.82\\
 & CF$_4$ & 2.50 & 11.35 & 86.15 & .0160 & .090 & 4.09\\
 & Edgeworth & 1.85 & 5.45 & 92.70 & .0192 & .055 & 1.67\\
 & Hybrid & 2.43 & 2.73 & 94.85 & .0177 & .013 & .15\\
\addlinespace[3pt]
Known law & Hybrid & 2.43 & 2.08 & 95.50 & .0180 & .012 & .17\\
\bottomrule
\end{tabular}
\end{center}
\end{table}

\subsection{Nonlinear vector calibration}\label{app:vectorexperiments}
\subsubsection{Model, target and information budget}
The parameter dimension is $p=3$ and the network dimension is $d=3$. Each of the 216 clients on the lazy $6^3$ torus uses the same law. Let $X$ be uniform on the three rows of $I+.4\onevec\onevec^\top$, let $Q=\E XX^\top$, and set
\[
\begin{gathered}
 \theta^\star=(.4,.3,-.2)^\top,\qquad \lambda=.1,\\
 \Pp(Y=1\mid X=x)=s(x^\top\theta^\star)+\lambda x^\top Q^{-1}\theta^\star,\\
 g(\theta;X,Y)=\lambda\theta+X\{s(X^\top\theta)-Y\}.
\end{gathered}
\]
Here $s(t)=(1+e^{-t})^{-1}$. The response probabilities are approximately $.7384,.6852,.4127$, all in $(0,1)$, and
$\E g(\theta^\star;X,Y)=\lambda\theta^\star-\lambda Q Q^{-1}\theta^\star=0$.
The sample Jacobian satisfies $.1I\preceq Dg\preceq.67I$. A global population Hessian-Lipschitz constant is
$M_3\le\E\|X\|^3/(6\sqrt3)=.3313$, and root-noise norm is at most $1.0154$.
The eigenvalues of $H$ are approximately $.1772,.1813,.4840$, those of $\Sigma$ are $.0693,.0770,.3611$, and $\|H\Sigma-\Sigma H\|=.002964$.
The resulting vector gradients have coupled coordinates.

The whole network starts at zero and uses adapt-then-combine updates without final gossip. Step sizes $.08,.04$ use $T=2086,4352$, respectively, with 4,000 independent network trajectories each. The directions $e_1$ and $u_2=(1,-1,1)/\sqrt3$ are fixed before outcomes. For the true finite directional variance, the initialization quantity $2I_{T,u}$ in \eqref{eq:vectortransient} is at most $5.14\times10^{-5}$ in all four cells. This bound controls initialization for the reported finite endpoints.

At each endpoint we collect 400 fresh $(X,Y)$ observations. Their gradients form the centered empirical joint law, and their analytic sample Hessians give a symmetric estimate projected onto $[.1I,.8I]$. There are no extra Hessian observations. Covariance uses the empirical second moment with denominator $M$, consistently with the resampling law. At step $.08$, a prespecified sensitivity study uses a stream of 1600 observations. Its first 400 give the primary law and the Hessian, and the full stream gives the secondary law while the Hessian stays fixed. This comparison adds observations but no optimization trajectories.

All references use the same fitted Hessian and total finite directional variance. The nominal retention matrix is $.3I$; top-128 ranks the scalar nominal weights, rounded to 14 decimal places with flat-index tie breaking. Complete lag-3 retains all 92 positive coefficients at $k=0,1,2$. Actual matrix coefficients are evaluated at the fitted Hessian, retaining off-diagonal covariance terms in its eigenbasis. A complete empirical vector is independently sampled for each retained index. Joint resampling preserves within-vector dependence. Proposition~\ref{prop:vector} supplies calibration for these fixed matrix-coefficient sets.

\subsubsection{All directional results and statistical comparisons}
\begin{table}[ht]
\caption{All vector references and observation budgets, with 4,000 paired trajectories per step. Lower/upper errors and coverage are percentages; length is on the contrast scale. The final column is a conservative 95\% interval for $\Delta_{\rm tail}$ in percentage points, obtained from a joint two-tail Clopper--Pearson rectangle within each row. Rows sharing a step are paired, including the two $M$ values.}\label{tab:vectorall}
\begin{center}\small\setlength{\tabcolsep}{4pt}
\begin{tabular}{llrrrrrr}
\toprule Step / direction & Reference & $M$ & Lower & Upper & Cov. & Length & $\Delta_{\rm tail}$ CI\\\midrule
0.08 / $e_1$ & Gauss. & 400 & 3.35 & 2.00 & 94.65 & 0.10031 & [0.24, 2.51]\\
0.08 / $e_1$ & Hybrid & 400 & 2.43 & 3.45 & 94.13 & 0.10041 & [0.33, 2.24]\\
0.08 / $e_1$ & Lag 3 & 400 & 2.55 & 3.23 & 94.23 & 0.10047 & [0.13, 2.08]\\
0.08 / $u_2$ & Gauss. & 400 & 2.50 & 2.33 & 95.18 & 0.08624 & [0.00, 1.29]\\
0.08 / $u_2$ & Hybrid & 400 & 2.43 & 2.68 & 94.90 & 0.08502 & [0.00, 1.39]\\
0.08 / $u_2$ & Lag 3 & 400 & 2.40 & 2.68 & 94.93 & 0.08504 & [0.00, 1.42]\\
0.08 / $e_1$ & Gauss. & 1600 & 3.35 & 1.98 & 94.68 & 0.10044 & [0.24, 2.53]\\
0.08 / $e_1$ & Hybrid & 1600 & 2.55 & 3.30 & 94.15 & 0.10054 & [0.20, 2.16]\\
0.08 / $e_1$ & Lag 3 & 1600 & 2.63 & 3.03 & 94.35 & 0.10059 & [0.00, 1.94]\\
0.08 / $u_2$ & Gauss. & 1600 & 2.40 & 2.33 & 95.28 & 0.08634 & [0.00, 1.29]\\
0.08 / $u_2$ & Hybrid & 1600 & 2.35 & 2.78 & 94.88 & 0.08511 & [0.00, 1.57]\\
0.08 / $u_2$ & Lag 3 & 1600 & 2.40 & 2.70 & 94.90 & 0.08516 & [0.00, 1.44]\\
0.04 / $e_1$ & Gauss. & 400 & 3.55 & 1.88 & 94.58 & 0.05444 & [0.51, 2.84]\\
0.04 / $e_1$ & Hybrid & 400 & 2.90 & 2.60 & 94.50 & 0.05448 & [0.00, 1.77]\\
0.04 / $e_1$ & Lag 3 & 400 & 3.00 & 2.48 & 94.53 & 0.05451 & [0.00, 1.75]\\
0.04 / $u_2$ & Gauss. & 400 & 2.90 & 2.48 & 94.63 & 0.04731 & [0.00, 1.64]\\
0.04 / $u_2$ & Hybrid & 400 & 2.83 & 2.88 & 94.30 & 0.04689 & [0.00, 2.00]\\
0.04 / $u_2$ & Lag 3 & 400 & 2.93 & 2.85 & 94.23 & 0.04688 & [0.00, 2.08]\\
\bottomrule\end{tabular}

\end{center}
\end{table}

The primary family comprises 24 exact paired McNemar tests: two retained methods versus Gaussian, three binary outcomes, two steps and two directions. Holm adjustment applies across these 24 tests. The $M$ sensitivity uses a separate 12-test family for the two retained methods, directions and three outcomes. These tests compare outcome probabilities; the nominal level supplies the calibration benchmark. Tail-loss difference intervals use a paired 20,000-draw multinomial bootstrap and are labeled exploratory because the absolute-value functional is nonsmooth at the nominal tails. Length comparisons use paired mean-difference intervals. The artifact reports all discordance counts and adjusted probabilities.

At step $.04$, $e_1$ has a Hybrid-minus-Gaussian tail-loss difference of $-1.175$ points, exploratory interval $[-1.575,-.225]$, with a length change of $.0000348$ (paired 95\% interval $[.0000094,.0000602]$). Both tail changes have Holm-adjusted $p<10^{-6}$; the coverage equality test has adjusted $p=1$. At step $.08$, the corresponding tail-loss interval is $[-1.325,1.050]$, spanning zero. Hybrid's remaining upper error is $3.45\%$ and its coverage is $94.125\%$ (exact individual 95\% interval $[93.35,94.83]$). This is a visible finite-sample shortfall, despite a smaller empirical full-CDF discrepancy than Gaussian.

In direction $u_2$, the Gaussian tails are already close to nominal. Hybrid tail-loss difference intervals include zero at both steps. Lag-3 at $.04$ has coverage $94.225\%$, versus $94.625\%$ for Gaussian; the paired equality test has Holm-adjusted $p=.00483$. Raising $M$ to 1600 at $.08$ changes the $e_1$ Hybrid tails to $2.55\%/3.30\%$ and coverage to $94.15\%$; none of the secondary paired tests rejects after Holm adjustment. Increasing $M$ reduces empirical sampling error but leaves the endpoint pilot and fitted Hessian unchanged, leaving both location and Hessian error in \eqref{eq:vectorfresh}.

\begin{figure}[htbp]
\centering\includegraphics[width=.94\textwidth]{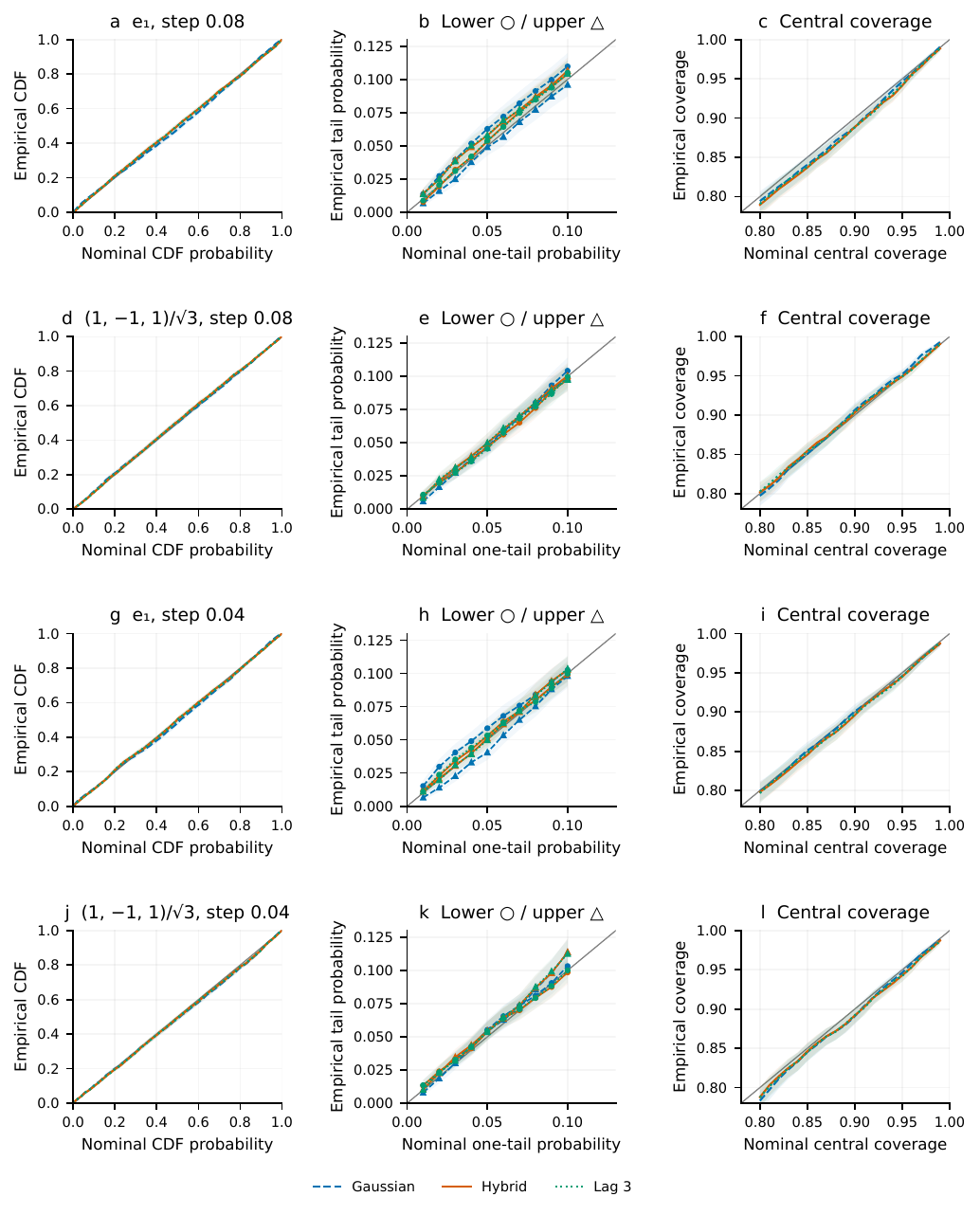}
\caption{All primary nonlinear vector calibration curves. Rows are the two predetermined directions at each step; columns show the full fitted-reference PIT CDF, lower/upper tail probabilities, and central coverage. Each step has 4,000 independent trajectories, shared across directions and methods. Shading gives pointwise 95\% Wilson intervals. Circles and triangles in the middle column distinguish lower and upper tails.}\label{fig:vectorcalibration}
\end{figure}

For a single PIT curve, the 95\% DKW radius is $.02147$. A simultaneous statement over all 12 primary curves instead uses $\sqrt{\log(480)/(8000)}=.02778$; dependence across paired curves does not invalidate this union bound. The narrower pointwise shading in the figures is labeled accordingly.

\subsubsection{Nonlinear coupling}
The first 1,000 trajectories at each step also propagate a linear chain using the same root innovations. They form a 2,000-trajectory subset of the primary study. Normalized $L^2$ coupling errors for $e_1/u_2$ are $.01588/.02045$ at $.08$ and $.00869/.01117$ at $.04$. Their 5,000-draw trajectory-bootstrap 95\% intervals are $[.01505,.01670]/[.01947,.02143]$ and $[.00820,.00919]/[.01061,.01172]$. Exact finite linear skewness is $-.2770/-.0641$ at $.08$ and $-.2202/-.0495$ at $.04$, confirming that directions experience different shape distortions. The small coupling errors and differing directional skewness support the fixed-contrast analysis. Fitted-reference calibration also depends on the nuisance estimates studied below.

\begin{figure}[!t]
\centering\includegraphics[width=.8\textwidth]{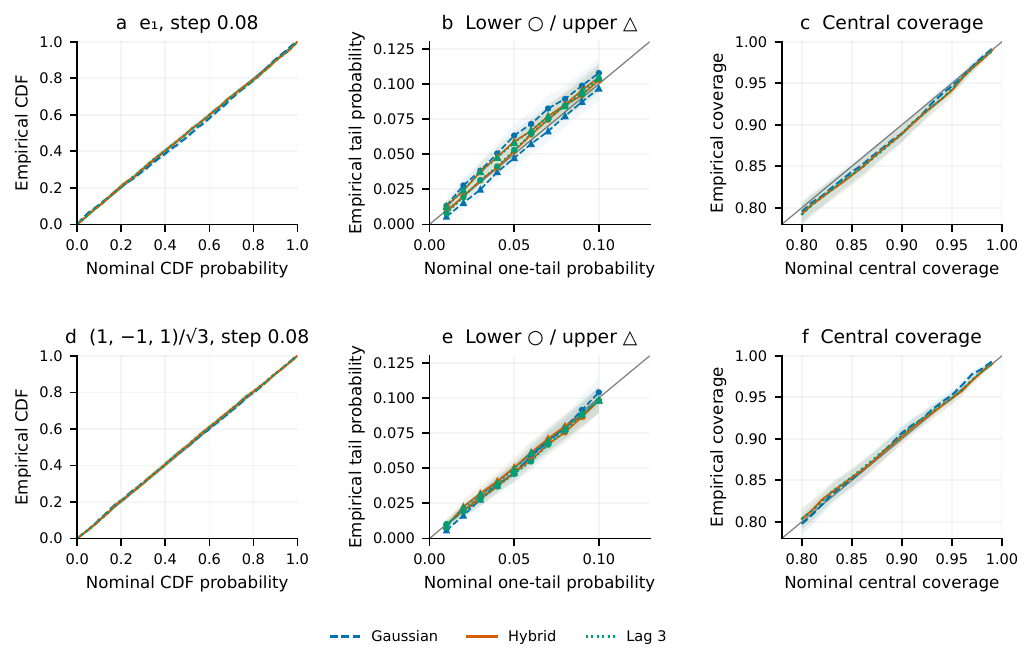}
\caption{Prespecified $M=1600$ sensitivity at step $.08$, with the same 4,000 endpoints and the same first-400-observation Hessian estimates as the primary study. Both predetermined directions and all three references are included. Shading is pointwise 95\% Wilson uncertainty. Both observation budgets use the same trajectories.}\label{fig:vectorM}
\end{figure}

\subsubsection{Numerical accuracy of vector references}\label{app:vectorprecision}
All retained references use $B=1499$ draws with analytic Gaussian-tail integration; Gaussian quantiles are analytic. We assess conditional numerical error separately from uncertainty across network trajectories. For the primary study, the first 128 endpoints at each step also receive independent $B=5999$ and $23999$ references. Relative to the largest-$B$ approximation, normalized RMS quantile discrepancies range from $.0260$ to $.0510$, and median absolute PIT discrepancies from $.00395$ to $.00482$. These satisfy the prespecified tolerances $.08$ and $.015$, respectively. The oracle comparisons in Appendix~\ref{app:vectororacle} have maxima $.0522$ and $.00590$ under the same check, and the pilot-location references in Appendix~\ref{app:pilotlocation} also satisfy both tolerances.

Repeated references quantify variability conditional on the observations. In the primary study, 32 repetitions at eight fixed endpoints give measured-to-estimated PIT variance ratios from $.82$ to $1.02$. For pilot-location comparisons, 16 paired repetitions at the first 32 endpoints per step use common random numbers across locations. Normalized RMS location-induced quantile shifts range from $.00124$ to $.00897$, compared with paired reference standard deviations $.00023$ to $.00132$. These quantify numerical resolution of reference changes; the paired trajectory tests assess changes in tail probabilities.

No reference was undefined; undefined references would have been counted as failures.

\subsubsection{Oracle sensitivity to law and curvature}\label{app:vectororacle}
An exploratory factorial diagnostic uses the same 8,000 endpoints to examine sensitivity to law and curvature estimation. For each reference, step and predetermined direction, it substitutes the true Hessian, the exact six-atom root law, or both. The fitted-Hessian/empirical-law reference completes the four combinations. Each law supplies its own covariance, and all references retain their prescribed deterministic sets. Table~\ref{tab:vectororacle} reports the hybrid references. Across the four combinations the Gaussian tail errors change by at most $.1$ points, and complete lag-3 behaves like top-128; all 48 rows are in the artifact. This diagnostic was specified after inspecting the primary results; its numerical accuracy is assessed in Appendix~\ref{app:vectorprecision}.

At step $.08$ in direction $e_1$, true Hessian alone leaves Hybrid upper error $3.325\%$ and coverage $94.10\%$. The true root law with the fitted Hessian gives $2.775\%$ and $94.75\%$; replacing both gives $3.025\%$ and $94.40\%$. Relative to the fitted-law upper error of $3.45\%$, substituting the true root law lowers the upper error to $2.78\%$; Appendix~\ref{app:vectordecomp} shows that residual replacement contributes little on the same endpoints and that the exact linear law leaves a deviation within Monte Carlo error. The location-only path below separates one aspect of law estimation from access to its exact distribution.

\begingroup\small
\begin{longtable}{lllrrrrr}
\caption{Exploratory vector nuisance sensitivity of the hybrid reference on the same 4,000 endpoints per step. Each tail and coverage is a percentage; length is on the contrast scale. E/T mean estimated/true Hessian or empirical/true root law. Any T row is an oracle diagnostic. Retained references use $B=1499$; Gaussian references are analytic.}\label{tab:vectororacle}\\
\toprule Step / direction & Reference & H / law & Lower & Upper & Cov. & Length & PIT discrepancy\\\midrule\endfirsthead
\toprule Step / direction & Reference & H / law & Lower & Upper & Cov. & Length & PIT discrepancy\\\midrule\endhead
0.08 / $e_1$ & Hybrid & E / E & 2.43 & 3.45 & 94.13 & 0.10041 & 0.0125\\
0.08 / $e_1$ & Hybrid & E / T & 2.48 & 2.78 & 94.75 & 0.10064 & 0.0119\\
0.08 / $e_1$ & Hybrid & T / E & 2.58 & 3.33 & 94.10 & 0.10044 & 0.0117\\
0.08 / $e_1$ & Hybrid & T / T & 2.58 & 3.03 & 94.40 & 0.10061 & 0.0122\\
0.08 / $u_2$ & Hybrid & E / E & 2.43 & 2.68 & 94.90 & 0.08502 & 0.0128\\
0.08 / $u_2$ & Hybrid & E / T & 2.38 & 2.65 & 94.98 & 0.08519 & 0.0126\\
0.08 / $u_2$ & Hybrid & T / E & 2.40 & 2.65 & 94.95 & 0.08506 & 0.0124\\
0.08 / $u_2$ & Hybrid & T / T & 2.40 & 2.60 & 95.00 & 0.08515 & 0.0116\\
0.04 / $e_1$ & Hybrid & E / E & 2.90 & 2.60 & 94.50 & 0.05448 & 0.0156\\
0.04 / $e_1$ & Hybrid & E / T & 2.70 & 2.50 & 94.80 & 0.05462 & 0.0145\\
0.04 / $e_1$ & Hybrid & T / E & 2.83 & 2.60 & 94.58 & 0.05449 & 0.0168\\
0.04 / $e_1$ & Hybrid & T / T & 2.75 & 2.48 & 94.78 & 0.05463 & 0.0151\\
0.04 / $u_2$ & Hybrid & E / E & 2.83 & 2.88 & 94.30 & 0.04689 & 0.0184\\
0.04 / $u_2$ & Hybrid & E / T & 2.85 & 2.75 & 94.40 & 0.04697 & 0.0172\\
0.04 / $u_2$ & Hybrid & T / E & 2.83 & 2.73 & 94.45 & 0.04690 & 0.0176\\
0.04 / $u_2$ & Hybrid & T / T & 2.83 & 2.78 & 94.40 & 0.04696 & 0.0180\\
\bottomrule\end{longtable}
\endgroup
\subsubsection{Exact linear reference and error decomposition}\label{app:vectordecomp}
In the simulation the six root-gradient atoms $x_\ell$, their probabilities $\pi_\ell$ and the Hessian are known. The exact law of the finite linear directional endpoint, normalized by its standard deviation $\eta s_{T,u}$, therefore follows by inverting
\[
 \prod_{k<T,\,j}\ \sum_{\ell=1}^6\pi_\ell\exp\Bigl\{-\mathrm i t\,(W^{k+1})_{ij}(A^ku)^\top x_\ell/s_{T,u}\Bigr\}.
\]
This gives four comparisons on the same 4,000 nonlinear endpoints per step, without new trajectories (Table~\ref{tab:vectordecomp}). First, the exact linear law differs from the oracle top-128 hybrid reference by less than $10^{-4}$ in Kolmogorov distance, so residual Gaussian replacement is negligible; it differs from the variance-matched Gaussian by $.007$--$.021$. Second, the saved same-input linear endpoints, available for the first 1,000 trajectories, agree with the exact law: their PIT discrepancies are $.024$--$.031$, within the 95\% DKW radius $.043$. Third, evaluating the nonlinear endpoints with the exact linear law gives tails within two Monte Carlo standard errors of nominal, and the 95\% bootstrap intervals for the nonlinear-to-linear standard-deviation ratios all contain one. Fourth, the fitted hybrid reference on the same endpoints adds $.53$ points to the upper error for $e_1$ at $\eta=.08$, with 24 against 3 discordant endpoints; in the other cells the largest paired change is a lower-tail increase of $.25$ points for $e_1$ at $\eta=.04$ ($p=.04$). The upper error of $3.45\%$ at $\eta=.08$ therefore splits into $.525$ points from fitting the reference, a significant paired change, and $.425$ points under the exact linear law, which the present Monte Carlo precision does not resolve. The fitted reference combines the empirical root law, the fitted Hessian, the pilot location and finitely many draws; the factorial oracle in Appendix~\ref{app:vectororacle} suggests that the empirical root law contributes most.
\begin{table}[ht]
\caption{Error decomposition for vector contrasts. Distances are Kolmogorov distances from the exact linear law. Tails are percentages for the 4,000 nonlinear endpoints per step under the exact linear law and under the fitted hybrid reference. The paired change is fitted minus exact upper error, with discordant counts and exact McNemar $p$. SD ratio: nonlinear over linear standard deviation.}\label{tab:vectordecomp}
\begin{center}\small\setlength{\tabcolsep}{3.5pt}
\begin{tabular}{lccccrrrr}
\toprule
Step / dir. & Gaussian & Hybrid & Exact: Low/Up & Fitted: Low/Up & Change & Disc. & $p$ & SD ratio\\\midrule
.08 / $e_1$ & .0207 & $<10^{-4}$ & 2.48/2.93 & 2.43/3.45 & +.53 & 24/3 & $<10^{-4}$ & 1.013\\
.08 / $u_2$ & .0095 & $<10^{-4}$ & 2.43/2.70 & 2.43/2.68 & $-$.03 & 4/5 & 1 & 0.991\\
.04 / $e_1$ & .0162 & $<10^{-4}$ & 2.65/2.50 & 2.90/2.60 & +.10 & 10/6 & .45 & 1.008\\
.04 / $u_2$ & .0066 & $<10^{-4}$ & 2.80/2.73 & 2.83/2.88 & +.15 & 9/3 & .15 & 1.020\\
\bottomrule
\end{tabular}
\end{center}
\end{table}

\subsubsection{Fixed-endpoint pilot-location sensitivity}\label{app:pilotlocation}
Fresh gradients estimate the noise law at their evaluation point. To assess sensitivity to this location, we retain all 8,000 vector endpoints and use
\[
 x_p(\alpha)=\theta^\star+\alpha(\widehat\theta-\theta^\star),
 \qquad \alpha\in\{1,1/2,0\}.
\]
The three positions were fixed before evaluating these references. The original endpoint remains the interval center and the statistic whose coverage is measured. Every position uses the same first 400 observation codes, original fitted Hessian and nominal top-128 set. We re-evaluate the six gradient atoms at $x_p(\alpha)$, center their empirical law and use its covariance. Thus the path changes both the centered law and covariance while keeping the Hessian fixed. The positions $\alpha<1$ use the known simulation target and are oracle diagnostics.

\begin{table}[ht]
\caption{Pilot-location sensitivity, with 4,000 paired endpoints per step and $M=400$ throughout. Lower/upper errors and coverage are percentages. $\alpha=1$ uses the endpoint pilot, and $\alpha=0$ uses the true parameter as pilot. All intervals remain centered on the original endpoint. The PIT gap is the empirical uniform-CDF discrepancy.}\label{tab:pilotlocation}
\begin{center}\small\setlength{\tabcolsep}{5pt}
\begin{tabular}{ccclrrrrr}
\toprule
$\eta$&Direction&$\alpha$&Reference&Lower&Upper&Coverage&Length&PIT gap\\
\midrule
0.08&$e_1$&1&Hybrid&2.43&3.45&94.13&0.10041&0.0125\\
0.08&$e_1$&0.5&Hybrid&2.43&3.35&94.23&0.10041&0.0128\\
0.08&$e_1$&0&Hybrid&2.40&3.25&94.35&0.10040&0.0131\\
\midrule
0.08&$u_2$&1&Hybrid&2.43&2.68&94.90&0.08502&0.0128\\
0.08&$u_2$&0.5&Hybrid&2.38&2.68&94.95&0.08500&0.0128\\
0.08&$u_2$&0&Hybrid&2.38&2.68&94.95&0.08500&0.0129\\
\midrule
0.04&$e_1$&1&Hybrid&2.90&2.60&94.50&0.05448&0.0156\\
0.04&$e_1$&0.5&Hybrid&2.90&2.55&94.55&0.05448&0.0157\\
0.04&$e_1$&0&Hybrid&2.90&2.50&94.60&0.05448&0.0159\\
\midrule
0.04&$u_2$&1&Hybrid&2.83&2.88&94.30&0.04689&0.0184\\
0.04&$u_2$&0.5&Hybrid&2.83&2.83&94.35&0.04689&0.0183\\
0.04&$u_2$&0&Hybrid&2.83&2.83&94.35&0.04689&0.0183\\
\bottomrule
\end{tabular}

\end{center}
\end{table}

Table~\ref{tab:pilotlocation} shows modest changes in Hybrid tail allocation. For $e_1$ at $\eta=.08$, moving from the endpoint pilot to the true location lowers upper error from $3.45\%$ to $3.25\%$. Coverage increases from $94.13\%$ to $94.35\%$. At $\eta=.04$, upper error moves from $2.60\%$ to $2.50\%$. Gaussian references, omitted from the table, change by at most $.05$ points, and the other contrast changes little. The resulting Hybrid reference still has appreciable upper-tail error at the larger step; its full-CDF discrepancy also increases slightly along the path. Moving the pilot therefore changes tail allocation only modestly and leaves a gap in overall calibration.

Binary changes are paired by trajectory. We report mean-difference intervals and exact McNemar tests, adjusting the 48 tests by Holm's method: two steps, two directions, two references, two comparisons with $\alpha=1$, and three outcomes. None rejects at level $.05$ after adjustment. Paired tail-loss intervals use 20,000 multinomial bootstrap draws and are exploratory. The artifact includes every comparison, individual tail and coverage intervals, length differences and discordance counts. The finite-sample root-location law here is distinct from the exact-law oracle in Appendix~\ref{app:vectororacle}; the latter also removes empirical-law sampling error.

\subsection{Nonlinear coupling on a graph with a missing edge}\label{app:graphcoupling}
We examine how local graph asymmetry affects the coupling in Theorem~\ref{thm:vectortransfer}. Starting from a lazy $4^3$ torus, we remove the undirected edge $\{0,1\}$ and add its weight to the two endpoint diagonals. Clients are indexed by lexicographic coordinates. The scalar stochastic gradient is
\[
 g(x,\eps)=x+.3\log\cosh(x)+\eps,\qquad
 \Pp(\eps=3)=.1,\quad\Pp(\eps=-1/3)=.9.
\]
The root is $\theta^\star=0$, $h=1$, and the noise variance is one. Since $g'(x)=1+.3\tanh(x)$ and $|g''(x)|\le.3$, the primitive bounds are $\mu=.7$, $L_g=1.3$, $M_3=.3$ and $B_*=3$.

For each $\eta=.1,.05,.025$, we use 2,000 independent full-network trajectories with shared inputs for the nonlinear and linear recursions, initialized at zero. The horizon $T=\lceil14/(\eta\mu)\rceil$ is 201, 401 or 801. Clients 0 and 1 are the deleted edge's endpoints; client 42 is away from the defect. For each client, let $s_{T,i}^{\rm lin}=\sqrt{\Var(z_{i,T})}$, computed from the exact finite linear filter. The reported quantity is
\[
 \widehat C_i=\left\{\frac1N\sum_{b=1}^N
 \left(\frac{e_{i,T}^{(b)}-z_{i,T}^{(b)}}{s_{T,i}^{\rm lin}}\right)^2\right\}^{1/2},\qquad N=2000.
\]
If $Y_b$ denotes the squared normalized difference, its delta-method Monte Carlo standard error is
$\operatorname{sd}(Y_b)/(2\widehat C_i\sqrt N)$, using the sample SD. The intervals in Table~\ref{tab:graphcoupling} are pointwise normal approximations. Rows within a stepsize share trajectories; the total independent trajectory count is 6,000.

\begin{table}[htbp]
\caption{Local energies and nonlinear coupling after a graph perturbation. Top: range of $\max_i\chi_i$ over $\eta=.1,.025,.005$, with $\mu=.7$, and a uniform analytic bound. Unequal edge weights use fixed factors in $[.35,1]$. Bottom: normalized $L^2$ coupling and pointwise 95\% Monte Carlo intervals, with no final gossip. The last column is the largest deterministic initialization allowance $2I_{T,u}$ among the three displayed clients.}\label{tab:graphcoupling}
\begin{center}\small
\begin{tabular}{llrr}
\toprule
Graph & Side length & Energy-ratio range & Uniform bound\\\midrule
Torus & 4 & 1.000--1.000 & 1\\
Torus & 6 & 1.000--1.000 & 1\\
Torus & 8 & 1.000--1.000 & 1\\
One missing edge & 4 & 1.079--1.312 & 32\\
One missing edge & 6 & 1.174--1.333 & 32\\
One missing edge & 8 & 1.237--1.335 & 32\\
Unequal weights & 4 & 1.196--1.604 & $8/.35$\\
Unequal weights & 6 & 1.682--2.096 & $8/.35$\\
Unequal weights & 8 & 1.877--2.144 & $8/.35$\\
\bottomrule\end{tabular}
\par\medskip
{\setlength{\tabcolsep}{4pt}
\begin{tabular}{rrrrr}
\toprule
$\eta$ & Client 0 & Client 1 & Client 42 & Initialization\\\midrule
0.1 & 0.01045 [0.01026, 0.01064] & 0.01052 [0.01034, 0.01071] & 0.01126 [0.01110, 0.01143] & 0.000055\\
0.05 & 0.00568 [0.00561, 0.00576] & 0.00563 [0.00556, 0.00570] & 0.00623 [0.00615, 0.00632] & 0.000134\\
0.025 & 0.00333 [0.00327, 0.00338] & 0.00331 [0.00325, 0.00336] & 0.00366 [0.00360, 0.00373] & 0.000273\\
\bottomrule\end{tabular}}

\end{center}
\end{table}

The coupling decreases with stepsize at both the defect endpoints and the more distant client. The deterministic energy comparison also shows modest local energy variation over the evaluated graphs. Its uniformity over graph families follows from Proposition~\ref{prop:energycomparison}. This scalar additive-noise model isolates the population nonlinear remainder: its conditionally centered state-dependent noise term is zero. The fixed 64-client study measures finite-time coupling; the growing-graph non-Gaussian conclusion and fitted-reference coverage concern separate quantities.

The same saved endpoints also permit a scale comparison after $q=4$ and 16 extra communication rounds. At $\eta=.05$, client 0's linear standard deviation is $.04559,.02427,.02007$ for $q=0,4,16$, while normalized coupling is $.00568,.01038,.01243$. Absolute coupling decreases slightly, but the standard deviation decreases more. This illustrates the denominator in Corollary~\ref{cor:nonlineargossip}. These rounds are added after the same optimization horizon; the main gossip comparison instead holds total communication fixed. The artifact reports the complete client/gossip comparisons and their finite variances.

\section{Position relative to closely related work}\label{app:related}
\begin{table}[htbp]
\caption{Nearest inference and stationary-analysis settings. Rows distinguish target statistics, model scope and distributional guarantees.}\label{tab:positioning}
\begin{center}\small
\begin{tabular}{P{.22\textwidth}P{.27\textwidth}P{.43\textwidth}}
\toprule
Work & Statistic / setting & Relevant conclusion and distinction\\
\midrule
\citet{li2022localsgd}; \citet{gu2024} & Local-SGD time averages; decentralized online one-step estimators & Inference for estimators with additional averaging or correction. These operations change the local endpoint studied here.\\
\citet{bonnerjee2025} & Client-averaged outputs and associated client processes; decreasing stepsizes & Final-output and functional inference, including client-process limits. Their decreasing-step analysis differs from the fixed-step local stationary shape studied here.\\
\citet{levin2026} & Aggregated FedLSA endpoint; heterogeneous clients & Gaussian approximation and multiplier bootstrap for constant or decreasing stepsizes. The communicated aggregate is the statistic of interest.\\
\citet{versini2026} & Multidimensional constant-step local DSGD; heterogeneous objectives & Bias/covariance expansions with network-dependent remainders. Our homogeneous analysis complements these moment results by characterizing and calibrating the surviving distributional shape.\\
\citet{bianchi2013}; \citet{pu2020asymptotic,pu2022transient} & Distributed SA and DSGD on fixed networks; decreasing steps & Network-level central limit theorem and network-independent mean-square error. The local regime $m\eta\to\infty$ and single-client shape are not addressed.\\
This work & One client's constant-step endpoint; homogeneous objectives & Scalar Gaussian boundary and transient law; feasible shape calibration with transfer to fixed contrasts of nonlinear vector objectives.\\
\bottomrule
\end{tabular}

\end{center}
\end{table}
Table~\ref{tab:positioning} compares target statistics, model scope and guarantees. The comparisons below explain how these settings relate to local distributional calibration.

\paragraph{Averaging, correction and endpoint inference.}
The constant-step CLT of \citet{mou2020} concerns averaged linear stochastic approximation and includes a stepsize correction to its asymptotic covariance. Online covariance estimation \citep{zhu2023cov} and the hierarchical trajectory construction of HiGrad \citep{su2023higrad} likewise exploit averaging. A local endpoint retains a different set of spatial and temporal coefficients. In the homogeneous quadratic model, spatial averaging eliminates disagreement eigenmodes exactly, whereas a client endpoint generally retains them. Appendix~\ref{app:v3experiments} evaluates a local one-step comparator with the same fresh gradients. Its single-client construction complements the distributed online procedure of \citet{gu2024}.

\paragraph{Residual and dependent-data bootstrap.}
Resampling innovations through an estimated linear filter has a substantial precedent in the sieve bootstrap \citep{buhlmann1997}. Its validity depends on the statistic and the relation between the original process and its autoregressive companion \citep{kreiss2011}. Our filter coefficients are determined by the known network and the curvature, or common Hessian, at the target. The target can converge to a non-Gaussian weighted innovation sum, and only a prescribed subset is resampled. The proof accounts for the residual Gaussian replacement and the fitted innovation law separately. With reused records, the observed iterates remain dependent, but the underlying innovations are independent under our model. A pathwise residual-contamination bound recovers their empirical law without treating the iterates as independent. This differs from directly resampling blocks of stationary observations \citep{kunsch1989}. Dependent innovations require additional control, as illustrated by Markovian stochastic approximation \citep{huo2023,samsonov2025}.

\paragraph{Stationary laws and non-Gaussian mechanisms.}
\citet[Lemma 3.3 and Theorems 3.4--3.5]{wei2025constant} give linearization and Gaussian approximation from arbitrary initialization, with concentration in Section 4. Our graph-uniform normalization $\eta\sigma\sqrt{S_i(h)}$ resolves a client's uncertainty in growing networks. Uniform sample contraction and bounded noise at the minimizer give local moment control, which supports empirical-law calibration when a Gaussian limit fails. Power-law tails under multiplicative random maps \citep{gurbuzbalaban2025heavy} and non-Gaussian small-step limits at flat minima \citep{zhang2026flat} concern different mechanisms. In the latter, vanishing curvature changes the scaling and the limiting drift. Here curvature stays positive and an explicit non-Gaussian innovation component survives the network limit.

\paragraph{Harness representations and proof tools.}
Harness processes construct stationary random fields by propagating independent noise through random-walk kernels \citep{ferrari2006,seppalainen2015}; massive and non-homogeneous crystals provide related spatial constructions \citep{ferrari2007}. We use this existing representation for the local component and prove its joint transition with finite-torus consensus energy. Disjoint time windows establish independence of the limiting components. The empirical calibration, finite-horizon conditions and nominal coefficient budget are additional statistical conclusions. Gaussian remainder smoothing, smooth replacement \citep{chatterjee2006}, empirical Wasserstein control \citep{fournier2015}, and martingale $L^p$ bounds \citep{pinelis1994} supply the analytical tools.

\paragraph{Network-level limits and adaptive networks.}
For decreasing stepsizes on fixed networks, distributed stochastic approximation satisfies a central limit theorem whose covariance, under doubly stochastic weights, coincides with that of centralized SGD \citep{bianchi2013}, and \citet{morral2017} characterize when adaptation--diffusion schemes keep this property. \citet{pu2020asymptotic,pu2022transient} show that the mean-square error of DSGD becomes network independent after a transient time governed by the spectral gap. For constant steps, diffusion strategies have per-agent steady-state mean-square expansions \citep{lopes2008diffusion,chen2015learning,sayed2014adaptation}, slow-adaptation analyses describe detection performance \citep{matta2016}, and consensus+innovations schemes give related asymptotics for distributed estimation \citep{kar2013consensus}. Gradient tracking and push-sum methods improve optimization rates \citep{yuan2016convergence,pu2021distributed,assran2019stochastic}, and communication-efficient distributed inference aggregates statistics across machines \citep{jordan2019communication}. These works describe network-level or variance-level behavior; the local regime $m\eta\to\infty$ and its distributional shape are the subject of the present analysis.

\paragraph{Moment corrections and bootstrap.}
Edgeworth and Cornish--Fisher expansions refine normal approximations for sums of many comparable terms \citep{cornish1938,bhattacharya1976,hall1992}, and the bootstrap captures the leading skewness correction in such problems \citep{singh1981}. Both rest on a small parameter, here $\sum|b_{kj}|^3\le\Lambda_i$, which does not vanish in the local regime. Proposition~\ref{prop:moments} turns this into a lower bound for every finite-moment reference, and Appendix~\ref{app:momentexperiments} quantifies the resulting errors. Calibration is assessed through probability integral transforms, as in probabilistic forecasting \citep{gneiting2007}.

\section{Fixed-filter empirical-law diagnostics}\label{app:metricexperiments}
This experiment isolates the empirical-law comparison in Proposition~\ref{prop:ordertworef}. It holds the true scalar filter, curvature and retained set fixed, and compares the fitted hybrid CDF with its oracle counterpart for that same set. The common oracle replacement error is held fixed, isolating reference sensitivity to law estimation.

\paragraph{Design.}
We use the lazy $6^3$ torus, $\eta=.05$, $h=1.4$, nominal curvature 1 and $T=410$. The innovation laws are the skew two-point law and the centered, variance-one $\operatorname{Beta}(4,7)$ law. For the skew law, the retained set contains the eight largest nominal coefficients. For the beta law, it contains only the current local coefficient, supplying the common smooth anchor. Each law therefore has its own fixed-filter comparison. For each law, 500 independent sample sequences of length 6400 supply paired prefixes at $M=100,400,1600,6400$. Every prefix is centered and its own empirical variance is used. The 1,000 independent sequences yield 4,000 paired-prefix references, all defined; their design and seeds are supplied in the artifact.

\paragraph{Numerical evaluation and uncertainty.}
The skew head is summed over its $2^8$ configurations. The empirical beta reference is a finite Gaussian mixture; its oracle uses 128-point Gauss--Jacobi integration, checked against 256 points and adaptive integration. No retained-sum Monte Carlo is used in this diagnostic. We compute $\mathfrak d_2$ from the integrated empirical CDF, $W_2$ from quantile integration, and both outer reference quantiles. For the beta law, integrated-CDF sign changes are located before integration. Independent moment identities and adaptive quadrature check this calculation.

\begin{figure}[ht]
\centering\includegraphics[width=.8\textwidth]{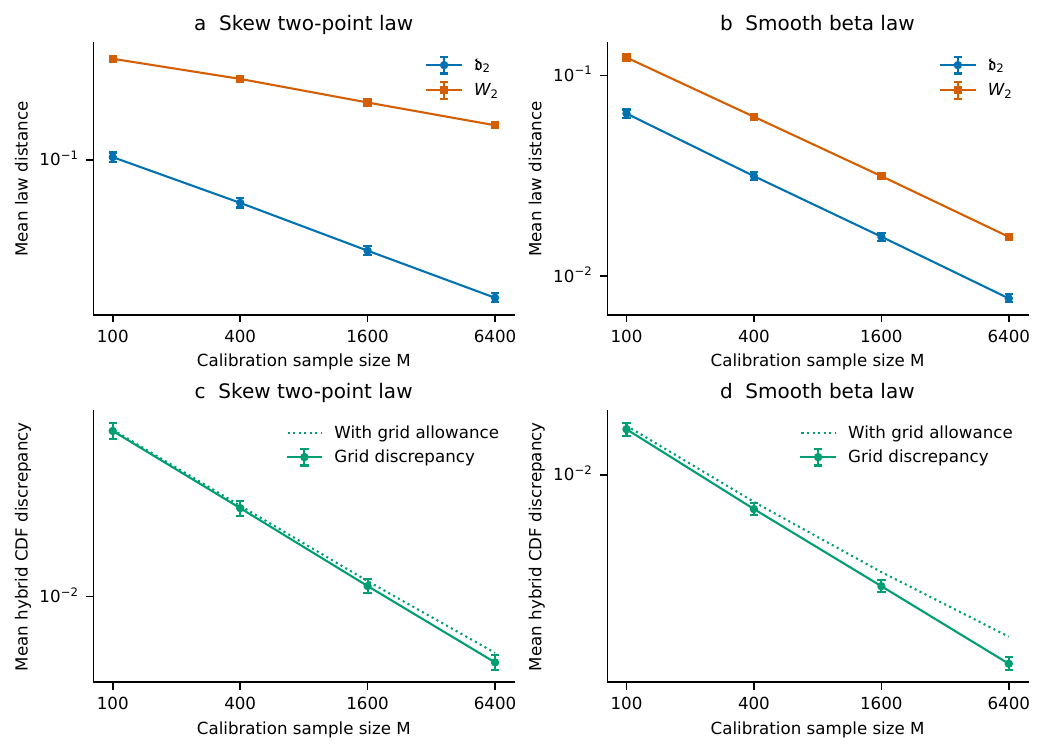}
\caption{Fixed-filter law estimation. Top: mean centered second-order distance and Wasserstein distance. Bottom: fitted-versus-oracle hybrid CDF discrepancy, with a dotted numerical upper bracket. Bars are pointwise 95\% bootstrap intervals over independent calibration sequences. Axes are logarithmic. The filter and retained set are fixed within each law.}\label{fig:metricdiagnostic}
\end{figure}

CDF discrepancies are evaluated on a grid whose second-derivative interpolation allowance is at most $5\times10^{-4}$. Gaussian tail truncation and a $2\times10^{-8}$ evaluation allowance are included in the reported numerical upper bracket. Direct mixture evaluation, doubled quadrature and adaptive integration agree within $1.1\times10^{-12}$ in the specified checks. The upper bracket combines the analytic interpolation bound with a conservative floating-point evaluation allowance. Figure~\ref{fig:metricdiagnostic} shows both grid estimates and estimates with the allowance. Pointwise 95\% percentile intervals use 4000 bootstrap resamples of the 500 independent sequence IDs, preserving the pairing across $M$. They describe Monte Carlo uncertainty in the mean reference discrepancies.

\begin{table}[ht]
\caption{All fixed-filter settings, with 500 independent sequences per law. CDF gap is the mean grid estimate; the last column is its bootstrap interval. A numerical allowance of at most $.000501$ applies in addition to Monte Carlo uncertainty. Full metric intervals and quantile discrepancies are released.}\label{tab:metricdiagnostic}
\begin{center}\small\begin{tabular}{llrrrr}
\toprule Law & $M$ & $\mathfrak d_2$ & $W_2$ & CDF gap & 95\% interval\\\midrule
Skew & 100 & 0.10398 & 0.43681 & 0.04440 & $[0.04114,0.04763]$\\
Skew & 400 & 0.05334 & 0.32565 & 0.02216 & $[0.02067,0.02372]$\\
Skew & 1600 & 0.02654 & 0.23072 & 0.01099 & $[0.01029,0.01174]$\\
Skew & 6400 & 0.01335 & 0.16549 & 0.00552 & $[0.00516,0.00590]$\\
\midrule
Beta & 100 & 0.06459 & 0.12277 & 0.01522 & $[0.01435,0.01609]$\\
Beta & 400 & 0.03150 & 0.06225 & 0.00731 & $[0.00691,0.00775]$\\
Beta & 1600 & 0.01573 & 0.03150 & 0.00360 & $[0.00341,0.00381]$\\
Beta & 6400 & 0.00775 & 0.01570 & 0.00177 & $[0.00167,0.00187]$\\
\bottomrule
\end{tabular}

\end{center}
\end{table}

\paragraph{Interpretation.}
Increasing $M$ from 100 to 6400 reduces the skew-law mean $\mathfrak d_2$ from $.10398$ to $.01335$, while mean $W_2$ decreases from $.43681$ to $.16549$. The corresponding reference CDF discrepancy decreases from $.04440$ to $.00552$. This illustrates why the second-order comparison can better reflect reference sensitivity for discrete laws. For the smooth beta law, both distances decrease comparably over this range, and the reference discrepancy falls from $.01522$ to $.00177$. This experiment measures law estimation at fixed geometry; Corollary~\ref{cor:ordertworates} additionally varies graph size, stepsize and retention budgets.
\section{Information requirements and computational accounting}\label{app:resources}
The calibration studies use 49,600 independent scalar trajectories and 8,000 vector trajectories. Separate scalar coupling studies use 6,400 torus trajectories and 6,000 missing-edge trajectories. Vector coupling uses a 2,000-trajectory subset of the calibration study; the oracle and pilot-location diagnostics reuse its 8,000 endpoints. These reused observations do not increase the independent replication count.

Calibration uses known filter geometry, stepsize, horizon, retained indices, fitted curvature and centered innovations. Vector models use a Hessian, joint innovation sample and fixed contrast.

\paragraph{Observations and derivatives.}
Table~\ref{tab:resources} separates independent observations from derivative evaluations. The counts are per fitted reference; methods on the same trajectory share their observations. Scalar quadratic curvature uses two independent samples of 100 gradients each. Logistic Hessians use 400 records, shared with the gradient sample whenever those indices overlap. In particular, $M=100$ does not reduce the logistic observation requirement below 400. The two vector contrasts share the same vector observations, Hessian and simulated head sums. A vector gradient or matrix Hessian counts as one evaluation here.
\begin{table}[ht]
\caption{Calibration resources in the numerical studies, excluding optimization. Reuse additionally stores $M$ gradient/state pairs. Nominal-curvature controls and model-specific known curvature are distinguished in the released ledger. Oracle laws and Hessians are diagnostic information and have no assigned deployable acquisition cost.}\label{tab:resources}
\begin{center}\small
\begin{tabular}{lrrr}
\toprule Calibration input & Fresh records & Gradient calls & Hessian calls\\\midrule
Scalar quadratic, fresh & $M+200$ & $M+200$ & 0\\
Scalar logistic, fresh & $\max(M,400)$ & $M$ & 400\\
Vector logistic, fresh & $\max(M,400)$ & $M$ & 400\\
Scalar quadratic, reused + fitted curvature & 200 & 200 & 0\\
Scalar logistic, reused + fitted curvature & 400 & 0 & 400\\
Reused + fixed nominal curvature & 0 & 0 & 0\\\bottomrule
\end{tabular}

\end{center}
\end{table}

\paragraph{Actual reference computation.}
The analytic Gaussian has actual $D=B=0$; the one-step bootstrap uses 1499 resampled estimators. Complete lag-3 retains 92 terms on the $6^3$ torus and 95 on the $12^3$ torus. The resource records distinguish configured budgets from executed work.

Scalar retained-sum generation costs $O(BD)$ arithmetic and $O(M+BD)$ memory in the array implementation. Quantile inversion costs $O(BI)$ for $I$ CDF evaluations. Vector projections, matrix coefficients and covariance operations add dimension and contrast factors. For one client, storing $T$ graph-power rows uses $O(Tm)$ entries; propagation costs $O((T+q)\,\mathrm{nnz}(W))$ and full sorting costs $O(Tm\log(Tm))$. This geometry is shared across replications. The torus spectral identities for residual energy are graph-specific.

\paragraph{Timing and communication.}
The resource records separate observation, computation and communication costs, with the scope of each timer specified. Vector timings cover both contrasts; separate Gaussian timing is unavailable. The main gossip comparator exchanges four gradient updates for communication-only rounds. Appendix~\ref{app:graphcoupling} instead adds gossip after a common optimization horizon. Pilot-location costs include gradient re-evaluation and reference computation using shared observations.
\end{document}